\PassOptionsToPackage{numbers, compress}{natbib}
\documentclass{article}

\usepackage[main, final]{neurips_2026}

\usepackage{tcolorbox}
\usepackage[utf8]{inputenc} 
\usepackage[T1]{fontenc}    
\usepackage{hyperref}       
\usepackage{url}            
\usepackage{booktabs}       
\usepackage{amsfonts}       
\usepackage{nicefrac}       
\usepackage{microtype}      
\usepackage{xcolor}         
\usepackage{pdflscape}
\usepackage{svg}
\newcommand{\tadpoleicon}{%
    \raisebox{-0.15em}{%
        \includegraphics[height=1em]{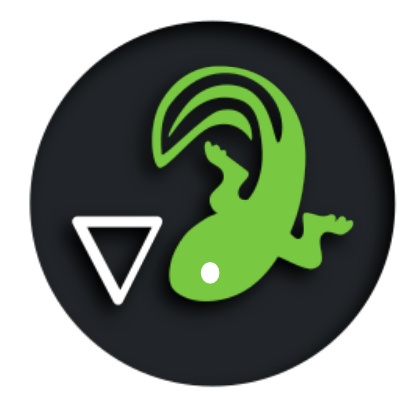}%
    }%
}
\usepackage{graphicx}
\usepackage{subcaption}
\usepackage{placeins}
\usepackage{multirow}
\usepackage{amsmath}
\usepackage{amssymb}
\usepackage{amsthm}
\usepackage[capitalise]{cleveref}
\crefname{appendix}{Appendix}{Apps.}
\Crefname{appendix}{Appendix}{Apps.}
\usepackage{rotating}
\usepackage{algorithm}
\usepackage{algorithmic}
\usepackage{wrapfig}
\hypersetup{hidelinks}

\newcommand{\current}{$\mathbf{u}_t$}
\newcommand{\future}{$\mathbf{u}_{t+\Delta t}$}

\newcommand{\latentcurrent}{$\mathbf{z}_t$}
\newcommand{\encoder}{$\mathcal{E}$}
\newcommand{\decoder}{$\mathcal{D}$}

\newcommand{\discriminator}{$\mathcal{A}$}
\newcommand{\subnet}{$\mathcal{S}$}

\newcommand{\formatedtable}[3]{
\begin{table}[tb]
\caption{#1}
\label{#2}
\centering
#3
\end{table}
}

\newcommand{\singlecolumnformatedtable}[3]{
\begin{table*}[tb]
\caption{#1}
\label{#2}
\centering
#3
\end{table*}
} 

\newcommand{\floatingformatedtable}[3]{
\begin{table}[htbp]
\caption{#1}
\label{#2}
\centering
#3
\end{table}
} 

\newcommand{\iso}{\texttt{Iso}}
\newcommand{\tcf}{\texttt{TCF}}
\newcommand{\mhd}{\texttt{MHD}}
\newcommand{\tbl}{\texttt{TBL}}
\title{\tadpoleicon{} Tadpole: Autoencoders as Foundation Models \\ for 3D PDEs with Online Learning}

\definecolor{ntcol}{rgb}{0.0, 0.62, 0.24}

\definecolor{colfelix}{HTML}{BF40BF}

\author{%
  Qiang Liu$^{1}$, Felix Koehler$^{1}$, Benjamin Holzschuh$^{1}$, Nils Thuerey$^{1,2}$ \\
  TUM School of Computation, Information and Technology$^{1}$\\
  Technical University of Munich\\
  Garching, Germany 85748 \\
  MCML, Munich Center for Machine Learning$^{2}$\\
  Munich, Germany 80804 \\
  \texttt{\{qiang7.liu,nils.thuerey\}@tum.de} \\
}

\begin{document}

\maketitle

\begin{abstract}
We introduce Tadpole, a novel foundation model for three-dimensional partial differential equations (PDEs) that addresses key challenges in transferability, scalability to high dimensionality, and multi-functionality. Tadpole is pre-trained as an autoencoder on synthetic 3D PDE data generated by an efficient online data-generation framework. This enables large-scale, diverse training without storage or I/O overhead, demonstrated by scaling to an equivalent of hundreds of terabytes of training data. By autoencoding single-channel spatial crops, Tadpole learns rich and transferable representations across heterogeneous physical systems with varying numbers of state variables and spatial resolutions. Although pre-trained solely as an autoencoder, Tadpole can be efficiently applied for multiple downstream tasks beyond reconstruction, including dynamics learning and generative modeling. For dynamics learning, we propose a novel parameter-efficient fine-tuning strategy that integrates low-rank adaptation, latent-space transformations, and reintroduced skip connections, achieving accurate temporal modeling with a minimal number of trainable parameters. Tadpole demonstrates strong fine-tuning performance across various downstream tasks, highlighting its versatility and effectiveness as a foundation model for 3D PDE learning. Source code and pre-trained weights of Tadpole are available at \url{https://tum-pbs.github.io/tadpole/}.
\end{abstract}

\section{Introduction\label{sec:introduction}}
The foundation model paradigm has achieved transformative success in Natural Language Processing (NLP) and Computer Vision (CV) \cite{nlmfoundation2024, visionfoundation2025}. Recently, it has been adapted to scientific machine learning to solve Partial Differential Equations (PDEs) \cite{Towards2023,p3d2025,yang2026toward,fluidintelligence2025}. Unlike specialized solvers, these foundation models aim to learn transferable representations across diverse physical systems to allow efficient
fine-tuning for new dynamics.

The prevailing strategy for building PDE foundation models is to learn the PDE dynamics by pre-training on large-scale trajectory datasets that capture a rich diversity of physical phenomena. These datasets are composed of numerous simulations, each representing a unique system defined by its governing equations, boundary conditions, and parameters (for instance, fluid viscosities or material stiffnesses). The model's task is to approximate the mapping from a system's past states to its future states. Formally, for a given phenomenon $\mathbb{P}^i$, the model learns to predict the state $\mathbf{u}_{t+\Delta t}^i$ from prior states $\mathbf{u}_{\leq t}^i$. The ambition is that by learning from many such examples, the model will distill universal physical principles, allowing it to \textit{zero-shot} generalize to new phenomena and adapt to specific tasks with minimal additional \textit{fine-tuning}.

Despite the attractiveness and potential of PDE foundation models, three fundamental challenges remain.  First, there is a notable shortage of PDE foundation models for \textit{three-dimensional} data. Most existing PDE foundation models focus on 1D or 2D problems, and a few notable exceptions that support 3D often rely on datasets that combine 3D with 2D/1D data \cite{MORPH2025,Walrus2025}, or purely rely on 2D data \cite{dpot2024}. Besides the heavily increased computational cost, a key reason for this lack of 3D models is the difficulty of collecting diverse and large-scale 3D PDE datasets for pre-training. Generating, storing, reading, and processing 3D data is significantly more expensive than 2D data, which fundamentally limits the diversity and scale of precomputed 3D PDE datasets. Since many real-world applications (e.g., weather forecasting, fluid dynamics, and material science) inherently involve 3D spatial domains, developing effective 3D PDE foundation models is crucial for advancing scientific machine learning.

In addition, transferability and generalization remain inconsistent. Ideally, most parameters of a foundation model can be reused without retraining, since the network should have learned general, transferable representations. For example,  zero-shot evaluation and Parameter-Efficient Fine-Tuning (PEFT) have become standard benchmarks for the quality of  NLP and CV foundation models \cite{peft2023,reviewPEFT2024,peftvision2024,peftfoundation2025,zeroshotvision2022}. However, most PDE foundation models still rely on full-parameter fine-tuning (FPFT), and preliminary zero-shot/PEFT experiments have shown limited success \cite{mpp2024,pdetransformer2025,MORPH2025}. The reliance on FPFT casts doubt on the paradigm of training PDE foundation models: whether a model can really learn a generalizable representation through pre-training on PDE dynamics with extreme variability.

Finally, current PDE foundation models focus solely on dynamics learning, neglecting the ability to extend to other functionality. For example, generative modeling has emerged as a powerful paradigm in scientific machine learning~\cite{uncertaintyaware2024,ruhling2024dyffusion, Jacobsen2025CoCoGen}. Achieving \textit{multi-functionality} across diverse downstream tasks, such as generative modeling, is a new challenge for PDE foundation models.

Therefore, developing a foundation model for 3D PDEs that efficiently and reliably generalizes across different tasks remains an open problem. Our work makes important steps to address these challenges with 
\textbf{Tadpole}, \textbf{t}hree-dimensional \textbf{a}utoenco\textbf{d}ers for \textbf{P}DEs with \textbf{o}nline \textbf{le}arning. It challenges the widespread notion that PDE foundation models require pretraining on the PDE dynamics of massive precomputed local data. We instead establish that foundation models can be trained for representation learning using simple, synthetic data generated on-the-fly during training. In contrast to foundation models in NLP, where representations emerge implicitly from next-token prediction, we achieve representation learning via autoencoding, explicitly optimizing a continuous latent space to capture the underlying data manifold. Our key innovations are: 

\begin{itemize}
    \item \textbf{A synthetic online learning framework:} We propose an efficient online learning framework with highly accurate, yet efficient GPU-based pseudo-spectral solvers and a novel buffer strategy, effectively bypassing I/O bottlenecks and storage limits at training time.
    \item \textbf{Transferable representations:} By pre-training Tadpole as an autoencoder on cropped individual fields, our models learn rich, transferable representations, enabling them to process varying PDE systems across different resolutions.
    \item \textbf{Efficient dynamics fine-tuning:} We propose a novel PEFT method for dynamics learning that integrates latent transformations, re-introduced skip connections, and LoRA~\cite{lora2022} fine-tuning, which better utilizes the pre-trained representations and achieves high accuracy.
    \item \textbf{Multi-task versatility:} We demonstrate that Tadpole excels across different downstream tasks, including autoencoding, dynamics learning, and generative modeling, and at resolutions up to $1024^3$ (i.e., on more than one billion degrees of freedom).
\end{itemize}

\section{Related work\label{sec:related_works}}

The potential of pre-trained neural networks to generalize across diverse physical systems was 
first characterized by \citet{Towards2023}. Subsequent research has prioritized architectural scalability, from traditional U-Net structures \cite{SPUS2025} to modern vision-transformer (ViT) designs \cite{Poseidon, dpot2024, pdetransformer2025}. Poseidon \cite{Poseidon} utilizes a multiscale transformer with time-conditioned layer norms 
to achieve continuous-in-time evaluations, while DPOT \cite{dpot2024} scales to 1 billion parameters using a Fourier-attention-based architecture. 
Other similar works include MPP \cite{mpp2024} and Walrus \cite{Walrus2025}, the latter introducing compute-adaptive tokenization to maintain stability. 

A central line of research focuses on the representation and embedding of heterogeneous PDE systems. Researchers have explored encoding PDEs as computational graphs to capture symbolic and numerical information simultaneously \cite{Pdeformer1d2024, Pdeformer2D2025}, introducing point-wise deep conditions to guide the global attention of transformers \cite{unisolver2025}, and utilizing SymPy-based libraries for automated symbolic tokenization \cite{timeseries2024}. To overcome the limitations of single-modality inputs, multimodal frameworks such as PROSE-PDE \cite{SunTowards2025} and UPS \cite{UPS2024} integrate numerical states with symbolic or textual descriptions \cite{Towards2025, multimodal2025}. In addition, UPS \cite{UPS2024} warm-starts from pre-trained Large Language Models (LLMs) to explicitly align data and improve computational efficiency.

Drawing inspiration from LLMs, recent studies
investigated In-Context Learning (ICL) for PDE foundation models \cite{Incontext2023, vicon2025, HumanDesigned2024}. Zebra \cite{zebra2025} and VICON \cite{vicon2025} leverage prompt-based trajectories to solve parametric PDEs, while \citet{Bcat2025} utilize a block causal transformer to treat historical frames as contextual priors for next-frame prediction. Parallel to these ICL methods, PhysiX \cite{physix2025} utilizes discrete tokenization and autoregressive next-token prediction to model physical processes.

Beyond these themes, the field is advancing on several adjacent topics. PreLowD~\cite{hemmasian2024pretraining}, MORPH~\cite{MORPH2025} and OmniArch~\cite{omniarch2025} have proposed lower dimensional pre-training. Frequency-adaptive fine-tuning was proposed~\cite{zhang2025f}, while constraint-aware pre-training \cite{pavingway2025} and physics-informed temporal alignment \cite{physicsinformed2025} incorporate PDE residuals to ensure physical consistency. A recent work~\cite{masked_ae} also pre-trains autoencoders for 2D PDEs, where the decoder is removed for dynamics fine-tuning, similar to previous latent-space learners~\cite{latent2019,latentdynamics2024}. Finally, frontiers such as operator discovery \cite{rahman2024pretraining,disco2025} and reward-model-driven reasoning \cite{reasoningPDE2025} represent the latest efforts for scientific foundation models. 

\section{Self-supervised pre-training\label{sec:pre-training}}

\begin{figure*}[t]
    \centering
    \includegraphics[width=\linewidth]{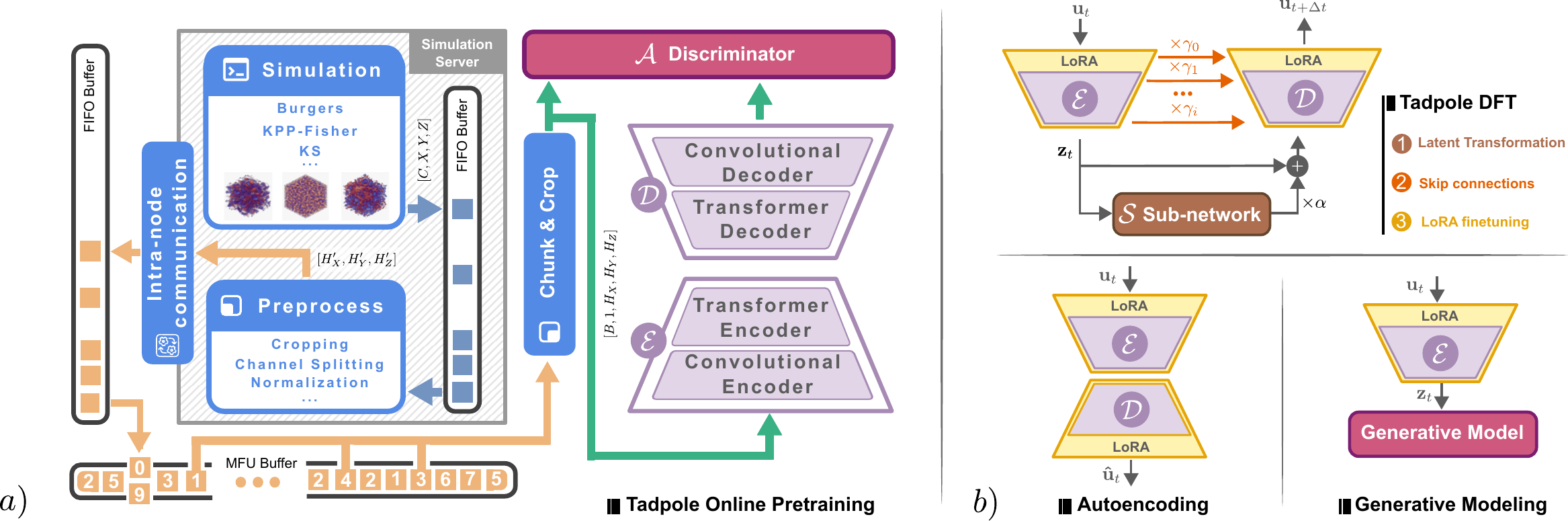}
    \caption{
        Overview of Tadpole:  a) Tadpole is pre-trained as an autoencoder on single-channel crops of 3D PDE data generated on-the-fly by a GPU-based solver with an efficient buffer strategy. b) The pre-trained Tadpole can be used for various downstream tasks, including autoencoding, dynamics learning with the novel Tadpole-DFT method, and generative modeling via latent flow matching. 
    }
    \vspace{-10pt}
    \label{fig:graphical_abstract}
\end{figure*}

\subsection{Training objective \label{sec:pre-training:training_objective}}

In traditional pre-training for PDE foundation models, the models learn the dynamics mapping a previous state \current{} to a future state \future{}. In contrast, we pre-train Tadpole as an autoencoder that reconstructs \current{} to learn rich, transferable spatial features of \current{} itself. Specifically, Tadpole is pre-trained as a Variational Autoencoder (VAE) with an adversarial loss to encourage sharper reconstructions, following the success of representation learning paradigms in CV \cite{taming2020,latentdiffusion2021}. Tadpole consists of an encoder $\mathcal{E}$ and a decoder $\mathcal{D}$. The encoder transforms the input $\mathbf{u}_t$ into a latent distribution $p_{\mathcal{E}}(\mathbf{z}_t|\mathbf{u}_t)$, while the decoder reconstructs the input from a sampled latent representation $\mathbf{z}_t$. A discriminator network $\mathcal{A}$ is optimized simultaneously to distinguish between real and reconstructed inputs and send feedback to the backbone training. Details of the pre-training target are provided in \cref{sec:training_objective}. We choose reconstruction as the pre-training target  over the dynamics target for the following reasons: 

\begin{itemize}
    \item In dynamics pre-training, a single \current{} may evolve into significantly different future states depending on PDE type, boundary conditions, and physical parameters. This necessitates high architectural complexity, as the network must distinguish between different physical systems by embedding a diverse set of parameters. 
    \item The dynamics pre-training target can usually only be applied to dynamics downstream tasks. Instead, reconstruction pre-training will provide a meaningful latent space of the solution domain, enabling more diverse applications in different types of downstream tasks.
    \item Reconstruction only requires learning the low-dimensional manifold of admissible PDE solutions, which is often smooth and highly structured due to spatial correlations induced by differential operators. In contrast, predicting the mapping from \current{} to \future{} entails learning the nonlinear flow on this manifold. This is inherently more difficult to learn than reconstruction, as it must accurately capture both the geometry of the solution space and the vector field governing its evolution. A more detailed discussion can be found at \cref{sec:discussion_mainfold}.

\end{itemize}

We show below that Tadpole, with its reconstruction-based pretraining, can be effectively fine-tuned for various downstream tasks (including the dynamics prediction) thanks to the generalization learned during pre-training. Details of fine-tuning methods will be discussed in \cref{sec:downstream_tasks}. 

\subsection{Online learning framework\label{sec:pre-training:online_learning}}

Another significant difference between Tadpole and other PDE foundation models is the efficient online learning~\cite{online,Melissa_online,Melissa2017} pipeline for pre-training, which
	(i) exposes the model to a diverse training distribution; 
	(ii) and eliminates I/O and storage challenges induced by large-scale 3D PDE datasets while sustaining high-throughput training without stalling. An overview is shown in \cref{fig:graphical_abstract} a).

\textbf{Data Generation:}
All pre-training data are generated on-the-fly using a PyTorch-based GPU solver, TorchFSM \cite{liu_2025_torchfsm}. Spatial derivatives are computed via pseudo-spectral methods based on Fast Fourier Transforms (FFTs), and time integration is performed using Exponential Time Differencing Runge--Kutta (ETDRK) schemes \cite{ETDRK2002,koehler2024apebench}, providing a highly efficient simulation backbone. Details of the solver can be found in \cref{sec:solver}. Although Fourier-spectral solvers impose periodic boundary conditions at the global domain level, Tadpole is trained exclusively on randomly sampled crops as will be discussed in \cref{sec:pre-training:dataset}. These crops correspond to local regions with non-periodic boundaries, thereby preventing the model from learning spurious periodic structures. Meanwhile, we also impose various initial conditions for different PDEs,  further increasing data diversity (cf. \cref{sec:initial_condition}).

\textbf{Buffering:}
To avoid the simulation speed affecting the training speed, we employ a three-stage buffering strategy. Simulation outputs are first written to a small First-In-First-Out (FIFO) buffer and asynchronously forwarded to the training processes. Each training process maintains a second FIFO buffer and a larger cache governed by a Most-Frequently-Used (MFU) replacement policy, from which batches are drawn directly during training. Background threads continuously replenish the MFU cache from newly arriving samples, effectively hiding simulation and communication latency behind training computation. The designed communication and buffer strategy can be effectively extended to multi-node HPC setups,
and more details are provided in \cref{sec:online_training_detail}.

\subsection{Dataset structure\label{sec:pre-training:dataset}}
Another significant challenge in pre-training PDE foundation models is the variability in spatial resolutions and state variable counts across different systems, which complicates batching for diverse datasets. We address this in Tadpole by pre-training on single-channel crops of 3D PDE data. Specifically, given training data of shape $[B, C, X, Y, Z]$ where $B$ is the batch size, $C$ is the channel dimension representing the number of state variables, and $X, Y, Z$ are spatial dimensions, we collapse the channel dimension into the batch dimension and randomly sample contiguous crops of shape $[B \times C, 1, H_X, H_Y, H_Z]$ for training. During inference, large domains are processed by encoding crops for each state variable in mini-batches, thereby avoiding the memory overhead of jointly processing all variables and spatial locations (cf. \cref{sec:computation_scaling}). In this work, we set the crop sizes to $H_{X,Y,Z}=64$ for pre-training. Meanwhile, to reduce data transfer volume and increase sample diversity, we apply an intermediate pre-cropping step before sending simulation data to the training process. Each simulation output is first cropped to an intermediate spatial size $H_{X,Y,Z}' = 96$, smaller than the full simulation resolution but larger than the final training crop size $H_{X,Y,Z}$, enabling multiple distinct random crops to be drawn from the same transmitted sample. Notably, in conventional dynamics pre-training, learning from single-channel crops is challenging due to dynamics dependencies across variables (e.g., velocity and pressure) and error accumulation at crop boundaries during long rollouts. In contrast, Tadpole pre-training avoids these issues due to its reconstruction learning target.

\subsection{Network architecture\label{sec:pre-training:network_architecture}}
We employ P3D \cite{p3d2025}, a state-of-the-art transformer for 3D PDEs, as the backbone for Tadpole. We adapt it to the reconstruction objective while preserving its core design principles. Specifically, we remove all embeddings related to PDE parameters, as the autoencoder focuses on input reconstruction rather than modeling complex dynamics. Furthermore, we eliminate the skip connections between the encoder and decoder to ensure that all information necessary for reconstruction is contained in the bottleneck latent space. An additional projection layer is appended to the encoder to map its output to the latent distribution parameters (mean and log-variance). Detailed architectural specifications are provided in~\cref{sec:training_details:network_details}.

A primary motivation for choosing P3D, beyond its scalability for 3D data, is its hybrid architecture, which combines convolutional layers with a transformer-based bottleneck. This configuration leverages the translation equivariance of convolutions. Consequently, during inference, crops with different sizes than those used in training can be directly processed thanks to the convolutional layers. This design not only provides greater flexibility for processing data with different spatial resolutions but also lays a firm foundation for downstream dynamics fine-tuning, as discussed in the following sections. It is worth noting that the proposed training and fine-tuning strategy is not limited to the P3D architecture but also applies to other architectures with translation equivariance.

\section{Flexible fine-tuning on downstream tasks\label{sec:downstream_tasks}}

Below, we outline our fine-tuning methodology for dynamics, autoencoding, and generative modeling. Importantly, we will introduce Tadpole dynamics fine-tuning (\textit{Tadpole-DFT}) methods for the dynamics task. An overview of the fine-tuning pipeline is shown in \cref{fig:graphical_abstract} b).

\textbf{Dynamics Learning:} 
With a pre-trained autoencoder like Tadpole, a natural approach for downstream dynamics tasks is to learn the PDE dynamics in the latent space rather than the high-dimensional physical space \cite{latent2019,latentdynamics2024}. However, since the latent representation \latentcurrent{} is significantly more compact than the original input \current{}, capturing precise dynamics purely in the latent space is often challenging. To address this, we propose the novel 
Tadpole-DFT approach, which encompasses:

\begin{itemize}
    \item \textbf{Latent transformation:} Tadpole-DFT introduces a lightweight sub-network \subnet{} between the pre-trained Tadpole encoder and decoder with a residual connection. As discussed in ~\cref{sec:pre-training:dataset}, capturing cross-variable interactions is crucial for dynamics learning. To solve this issue, we aggregate the latent spaces of all state variables after the encoder, enabling \subnet{} to learn correlations between state variables.
    \item \textbf{Re-introduced skip connections:} During fine-tuning, the skip connections between the encoder and decoder are re-established, each governed by a trainable scale factor $\gamma$. This allows the model to leverage both the latent dynamics from the sub-network and high-resolution spatial information from the skip connections to predict the future state.
    \item \textbf{LoRA fine-tuning:} The pre-trained encoder \encoder{} and decoder \decoder{} are fine-tuned using LoRA~\cite{lora2022} while their core weights remain frozen. For a pre-trained weight matrix $W_0 \in \mathbb{R}^{d \times k}$, LoRA approximates the update $\Delta W$ via a low-rank decomposition: $\Delta W = AB$, where $A \in \mathbb{R}^{d \times r}$, $B \in \mathbb{R}^{r \times k}$, and the rank $r \ll \min(d, k)$. During fine-tuning, the effective weight matrix is computed as $W = W_0 + AB$, where only the low-rank matrices $A$ and $B$ are trainable while $W_0$ remains frozen. The updated weights will be merged back into the original weights during inference to avoid additional computational overhead. LoRA enables the backbone to adapt to the new information flow from the skip connections and the sub-network \subnet{} while preserving the robust representations acquired during pre-training.
\end{itemize}

In addition, to prevent error accumulation at crop boundaries during rollouts, we exploit the translation equivariance of the Tadpole encoder to construct latent representations from uncropped data. Despite removing cropping, Tadpole-DFT remains easier to scale to larger datasets than conventional setups, as inputs are still encoded in mini-batches along the channel axis. Meanwhile, we zero-initialize the weights of the sub-network and backbone LoRA layers, as well as the scale factors for skip connections, which ensures that the model starts as a standard pre-trained autoencoder (outputting the current state \current{}) and gradually learns to transform \current{} into the future state \future{} during training, which further maintains the prior knowledge learned during pre-training.

\textbf{Autoencoding:} 
The pre-trained Tadpole model can be applied directly to unseen PDE systems for zero-shot autoencoding tasks. Alternatively, it can be fine-tuned on specific downstream systems to further enhance reconstruction quality, either by updating all model parameters or using LoRA to reduce the number of trainable parameters. 

\textbf{Generative Modeling: }
With a fine-tuned Tadpole, we can efficiently build a latent generative model~\cite{latentdiffusion2021} for 3D PDEs. For this, a new generative component is trained in the \latentcurrent{} space using a standard flow matching~\cite{flowmtaching2023} objective. At inference time, new samples are drawn from the latent flow matching model and transformed into high-fidelity 3D PDE data by the Tadpole decoder \decoder{}. Operating in the compact latent space rather than the high-dimensional pixel space enables high-quality generative modeling of complex 3D physical systems while minimizing memory and processing requirements.

\section{Experiments}
In this section, we first summarize the pre-training statistics and then evaluate Tadpole across various downstream tasks that involve challenging 3D phenomena. We assess its zero-shot capabilities and fine-tuning performance, comparing it against state-of-the-art PDE foundation models. 

\subsection{Pre-training statistics}

With the proposed online-learning framework, we pre-train Tadpole on 7 distinct PDE systems at four spatial resolutions ($64^3$, $128^3$, $256^3$, and $384^3$). Details of the PDEs and corresponding configurations can be found in \cref{sec:pretraining_equation}. We perform a spectral distribution analysis on the pre-training dataset, illustrating its large spectral diversity, cf.~\cref{sec:spectral_distribution}.  Meanwhile, we also scale Tadpole models of varying sizes. Specifically, we investigate S, B, and L sizes of 8.8, 38.1, and 152.1 million parameters, with corresponding compression ratios of 16, 8, and 4. The pre-trained Tadpole achieves average reconstruction RMSEs of $5.48\times10^{-3}$, $2.83\times10^{-3}$, and $2.06\times10^{-3}$, for the S-, B- and L-size models across all pre-training PDEs. The online data generation pipeline yields approximately 202 TB of training data, with no local storage required. While naturally dependent on hardware specifics, the online training pipeline achieves an overall $1.8\times$ speedup over an offline setup with pre-generated data in our high-speed SSD environment. In HPC environments, which are typically used for FM training and have lower I/O bandwidths for data loading, we have measured differences that are an order of magnitude larger. Thus, the online training is especially effective in such cases.

\subsection{Autoencoding\label{sec:downsream_autoencoding}}
\begin{figure*}[t!]
    \begin{subfigure}[t]{0.28\textwidth}
    \vspace{0pt}
    \includegraphics[scale=0.48]{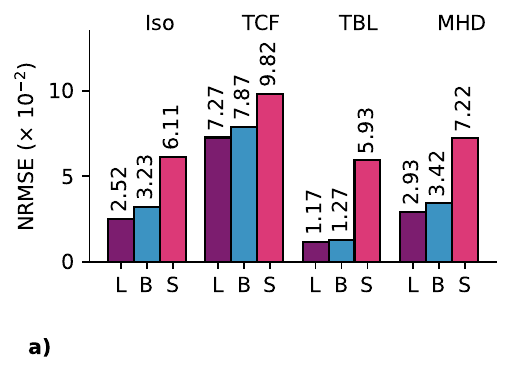}
    \end{subfigure}
    \begin{subfigure}[t]{0.38\textwidth}
    \vspace{0pt}
    \includegraphics[scale=0.48]{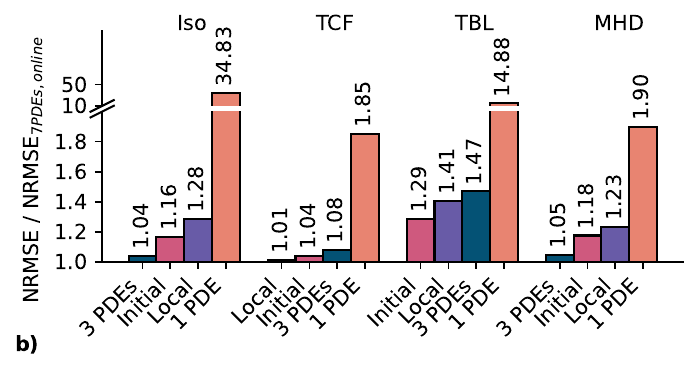}
    \end{subfigure}
    \begin{subfigure}[t]{0.28\textwidth}
    \vspace{0pt}
    \includegraphics[scale=0.48]{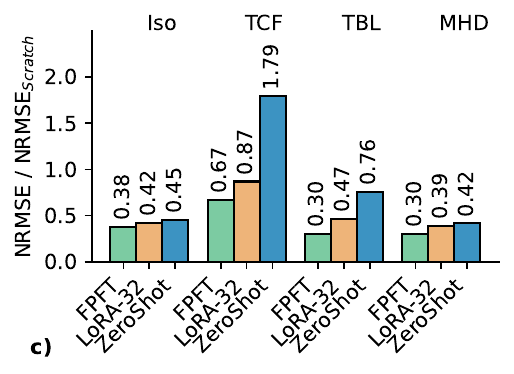}
    \end{subfigure}
    \caption{Performance of Tadpole on the downstream autoencoding task (exact NRMSE values in \cref{tab:downstream_autoencoder}). a) Zero-shot reconstruction NRMSE of Tadpole with different model sizes. Tadpole shows consistent
    scaling with respect to model size. b) Zero-shot relative NRMSE of Tadpole-B models pre-trained on datasets with varying diversity compared to the full-size online setup (at 1.0). All variants perform worse, with relative NRMSE values larger than 1.0, demonstrating that greater PDE diversity improves reconstruction generalization. 
    c)  Relative NRMSE of Tadpole-B models with different fine-tuning methods compared to models trained from scratch. Pre-training consistently improves performance.
    E.g., LoRA-32 fine-tuning reduces errors by more than 60\% for MHD compared to training from scratch.
    }
    \label{fig:reconstruction_bar_plot}
\end{figure*}
\begin{figure*}[t!]
    \centering
    \includegraphics[width=0.98\linewidth]{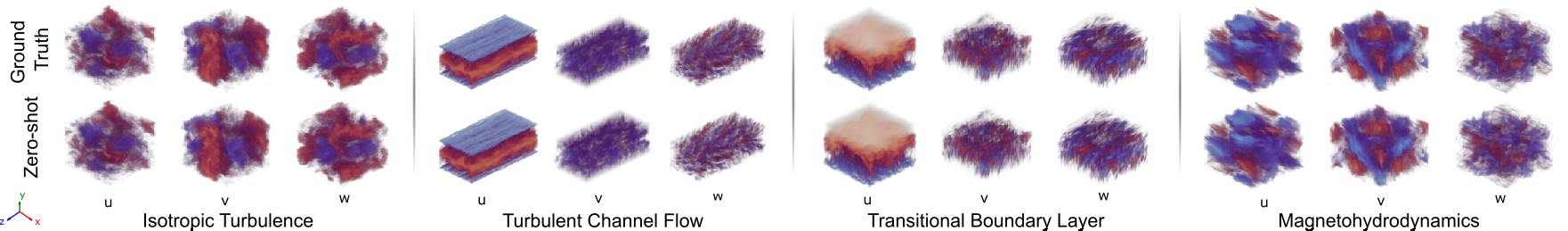}
    \caption{
    Visualizations of Tadpole-B zero-shot reconstruction on different datasets. Only velocity channels are shown here; additional ones are provided in \cref{sec:additional_results_visualizations:autoencoding}. The datasets feature high resolutions, ranging from 
    $96^2\times192$ for \tcf{} to 
    $1024^3$ for \iso{}.
    }
    \label{fig:reconstruction_vis_example}
\end{figure*}

Let $D=C\times X \times Y \times Z$ denote the dimension of a 3D PDE state. To evaluate Tadpole’s autoencoding performance for unseen data, we consider four representative 3D PDE systems: isotropic turbulence (\iso{}, $D=4\times1024^3$), turbulent channel flow (\tcf{}, $D=3\times96^2\times192$), magnetohydrodynamics (\mhd{}, $D=10\times512^3$), and transitional boundary layer flows (\tbl{}, $D=4\times224^3$). These systems exhibit diverse physical characteristics and significantly higher resolutions that Tadpole has not encountered during pre-training. Details of these datasets are provided in \cref{sec:downstream_dataset}.

\textbf{Zero-Shot Performance w.r.t Model and Dataset Sizes.} We first evaluate Tadpole in zero-shot settings. The results indicate a clear scaling trend on model size: larger models consistently outperform smaller ones, as shown in \cref{fig:reconstruction_bar_plot} a). Besides, a comparison of models pre-trained on datasets with varying diversity is presented in \cref{fig:reconstruction_bar_plot} b). Three successively simpler online training setups are introduced: pre-training with only three PDEs (KS, Burgers, and KPP-Fisher), with one PDE (KS), or only with PDE initial conditions. The findings show that incorporating additional PDEs during pre-training continuously improves zero-shot performance. In contrast, the reduced performance of the pre-training solely with initial conditions suggests that the PDE dynamics generate novel features that enhance reconstruction generalization. Additionally, we also introduce a model pre-trained on a 500GB local dataset generated with the same PDEs and parameter distributions (cf. \cref{tab:pde-trajectory-configs}). The online pre-trained model outperforms the local variant, indicating that the online learning strategy further increases data diversity while eliminating the training I/O and data storage bottleneck. Visualizations of zero-shot reconstructions are shown in \cref{fig:reconstruction_vis_example}.

\textbf{Fine-Tuning with Pre-trained Model:}
To verify the effect of pre-training, we fine-tune Tadpole under different settings and compare the performance with models trained from scratch. The results are summarized in \cref{fig:reconstruction_bar_plot} c). Pre-trained Tadpole shows substantial advantages, where both FPFT and LoRA-based PEFT consistently outperform their from-scratch counterparts. Notably, the zero-shot Tadpole B clearly surpasses from-scratch models on the \iso{}, \mhd{}, and \tbl{} tasks, which further highlights the effectiveness of pre-training.

\begin{wrapfigure}{o}{0.45\textwidth}
	\centering
	\includegraphics[scale=0.5]{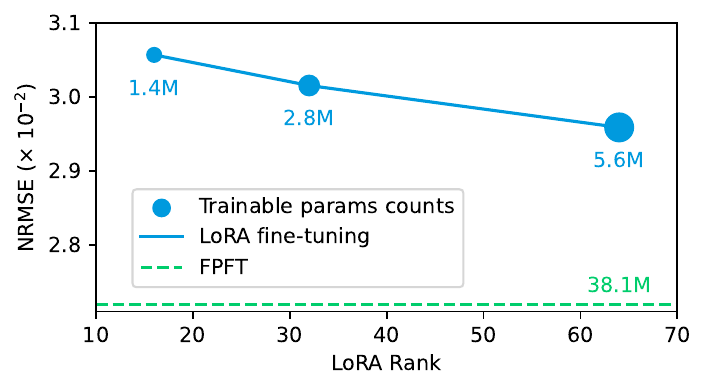}
	\caption{
		Reconstruction NRMSE of Tadpole-B with different LoRA ranks on \iso{} task. Larger rank shows better performance.
	}
    \vspace{-10pt}
	\label{fig:autoencoder_lora}
\end{wrapfigure}

\textbf{Effect of LoRA Rank:} \cref{fig:autoencoder_lora} further illustrates the impact of the LoRA rank on fine-tuning performance. Even with a small LoRA rank, Tadpole achieves lower reconstruction error than models trained from scratch (NRMSE$=7.17\times10^{-2}$), highlighting the efficacy of pre-trained representations. As the LoRA rank increases, performance continues to improve and approaches that of FPFT, while maintaining substantially fewer trainable parameters. 

\textbf{Flexibility in Domain Size and State Variable Count:}
The proposed crop-based strategy enables Tadpole to seamlessly handle varying domain sizes and state variable counts. In particular, for \iso{}, which features extremely high spatial resolutions, and \mhd{}, which has a high state variable count, Tadpole effectively accommodates these variations without any architectural modifications or retraining. Meanwhile, although Tadpole is trained on crops with $H_{X,Y,Z}=64$, we adopt different inference crop sizes for \tcf{} ($H_{X,Y,Z}=48$) and \tbl{} ($H_{X,Y,Z}=32$) to fully cover the spatial domains in these two cases, and Tadpole can still obtain impressive zero-shot performance by leveraging the translation equivariance of convolutional layers. Furthermore, for spatial resolutions such as \tcf{} with $96^2\times192$, the entire spatial domain can be processed in a single forward pass with similar performance (cf. \cref{sec:additional_results_visualizations:whole_domain_inference}). 

\subsection{Dynamics learning}
In this section, we evaluate Tadpole on a challenging dynamics learning task involving 3D cropped turbulent flows, following the setup of prior work~\cite{p3d2025}. The test contains an isotropic turbulence (\iso{}) and a turbulence boundary layer (\tbl{}) simulation, both cropped from a larger domain, with $128^3$ points. The cropping removes periodicity and introduces complex boundaries, thereby substantially increasing the difficulty of the dynamics learning task. We compare Tadpole against state-of-the-art 3D-PDE foundation models MORPH~\cite{MORPH2025} and DPOT~\cite{dpot2024}. We select variants with total parameter counts comparable to those of the corresponding Tadpole models to ensure fair comparisons. Meanwhile, we also include a concurrent foundation model, Walrus~\cite{Walrus2025}, with a significantly larger parameter count than all the other models. All PDE foundation model baselines are performed via FPFT from their released pre-trained weights. For Tadpole variants, we employ Tadpole-DFT with LoRA rank $r=32$ and a sub-network \subnet{} with a standard encoder-only transformer architecture. Details of the \subnet{} can be found in \cref{sec:training_details:network_details}. We also additionally compare Tadpole against several state-of-the-art network architectures trained from scratch in \cref{sec:exact_metric_values}. We utilize an enstrophy-based spectrum metric NRMSE$^{ES}$ to accurately evaluate the rollout performance (cf. ~\cite{probabilistic_forecasting,p3d2025}, and \cref{sec:enstrophy}). Corresponding evaluation in pixel space is also provided in \cref{sec:exact_metric_values}.

\begin{figure*}[t!]
    \begin{subfigure}[t]{0.69\textwidth}\vspace{0pt}\centering\includegraphics[scale=0.51]{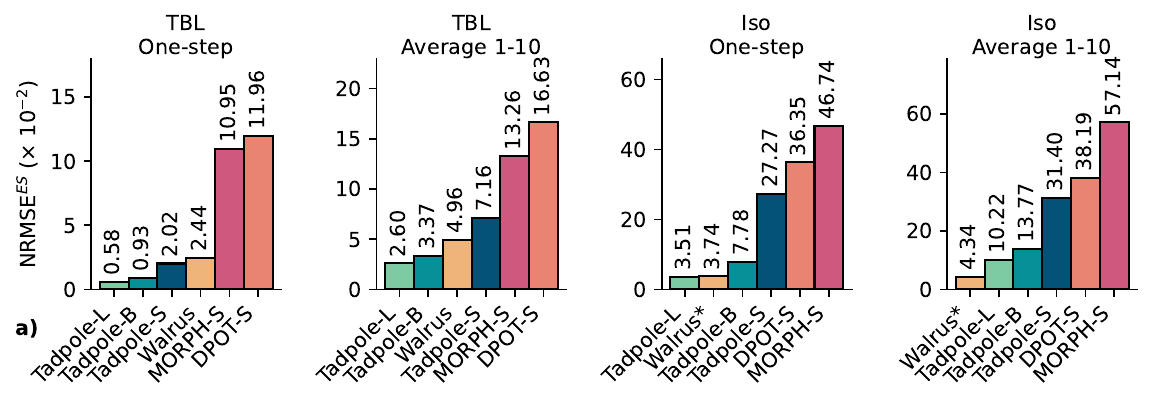}
    \end{subfigure}
    \begin{subfigure}[t]{0.29\textwidth}\vspace{0pt}\centering\includegraphics[scale=0.51]{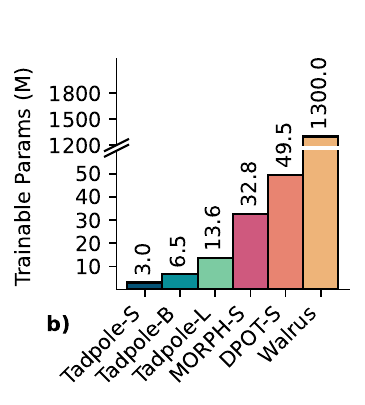}
    \end{subfigure}
    \caption{Performance of Tadpole on two distinct downstream dynamics tasks \tbl{} and \iso{}. a) Prediction NRMSE$^{ES}$ of different foundation models. Tadpole performs best on \tbl{}, and ranks among the top-performing methods on \iso{}.  b) The trainable parameters of different foundation models. Thanks to the LoRA introduced in the Tadpole-DFT method, only a very few parameters are fine-tuned in Tadpole, which makes it significantly smaller than the best-performing competitor, Walrus.}\label{fig:dynamics_bar_plot}
\end{figure*}

\begin{figure*}[t!]
    \begin{subfigure}[t]{0.69\textwidth}\vspace{0pt}\centering\includegraphics[scale=0.51]{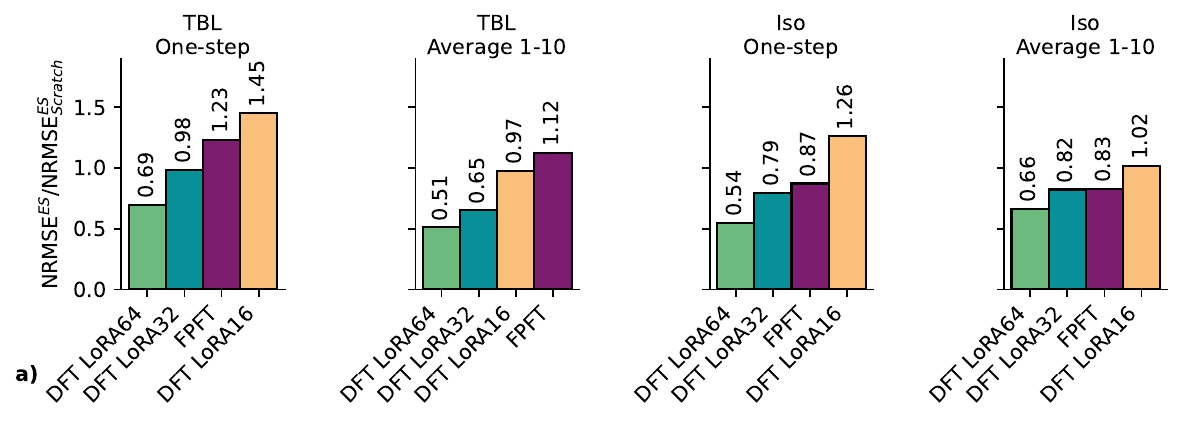}
    \end{subfigure}
    \begin{subfigure}[t]{0.29\textwidth}\vspace{0pt}\centering\includegraphics[scale=0.51]{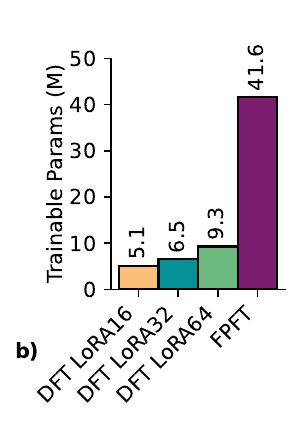}
    \end{subfigure}
    \caption{Performance improvements on the dynamics test from pre-training of Tadpole.
a) Relative NRMSE$^{ES}$ of Tadpole-B fine-tuned using various methods compared to the from-scratch variant. Increasing the LoRA rank in Tadpole-DFT consistently improves performance. b) Trainable parameters for different fine-tuning methods. The largest Tadpole-DFT variant utilizes only 22.3\% of the trainable parameters required by the FPFT/from-scratch variant.}\label{fig:dynamics_improvement_bar_plot}
\end{figure*}

\cref{fig:dynamics_bar_plot} summarizes model performance for two dynamics tasks. Tadpole models exhibit meaningful scaling: larger models outperform smaller ones. The Tadpole models always perform better than the DPOT and MORPH models, even the smallest S model with 10x fewer trainable parameters. Importantly, in the \tbl{} test, both the L and B size models outperform the Walrus model, which has over two orders of magnitude more trainable parameters: Tadpole-B features a 10-step enstrophy error of 3.37, while the 200x larger Walrus model yields 4.97. It is worth noting that Walrus performs better on the \iso{} test case for longer rollout, which, however, is included in Walrus's training data (marked with subscript * in \cref{fig:dynamics_bar_plot}a) ). This provides an advantage that the other models in this comparison do not have.

\textbf{Fine-Tuning Methodologies:} 
\cref{fig:dynamics_improvement_bar_plot} presents a comparison between the Tadpole-DFT fine-tuning strategy, FPFT, and training from scratch. As the LoRA rank increases, Tadpole-DFT consistently demonstrates improved performance. Notably, the LoRA 32 configuration outperforms the from-scratch model across all evaluated metrics. Furthermore, the LoRA 64 variant achieves a 49\% reduction in the 10-step average prediction error on the \tbl{} dataset compared to the from-scratch approach, while utilizing 77.7\% less trainable parameters. This result highlights the effectiveness of Tadpole-DFT in adapting the pre-trained Tadpole model to dynamics-learning tasks. In contrast, the FPFT variant of Tadpole-B does not consistently surpass the Tadpole-DFT variants, despite requiring substantially more trainable parameters. This behavior can be attributed to the fact that Tadpole is pre-trained as an autoencoder without explicit exposure to temporal dynamics; consequently, directly fine-tuning all parameters may lead to suboptimal local minima and unstable training. In contrast, Tadpole-DFT preserves the pre-trained representation by leveraging frozen weights in LoRA and incrementally incorporates dynamics learning through latent transformations and skip connections, thereby facilitating more effective learning of new dynamics features.  \cref{fig:val_rmse_lora32_pretrained} in the appendix presents the validation loss curves for Tadpole-B fine-tuned with Tadpole-DFT LoRA-32 and with FPFT. Tadpole-DFT not only achieves a lower final error but also converges faster and exhibits more stable training behavior than FPFT. 

\begin{figure}[b]
    \centering
    \begin{minipage}[t]{0.48\textwidth}
        \centering
        \includegraphics[width=\linewidth]{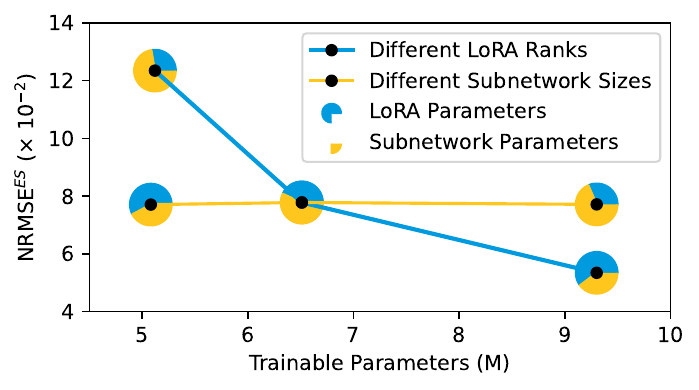}
        \caption{
            One-step NRMSE$^{ES}$ of Tadpole-B model with different
            sub-network sizes and LoRA ranks. LoRA ranks positively
            affect performance.
        }
        \label{fig:dynamics_size_ablation}
    \end{minipage}
    \hfill
    \begin{minipage}[t]{0.48\textwidth}
        \centering
        \includegraphics[width=\linewidth]{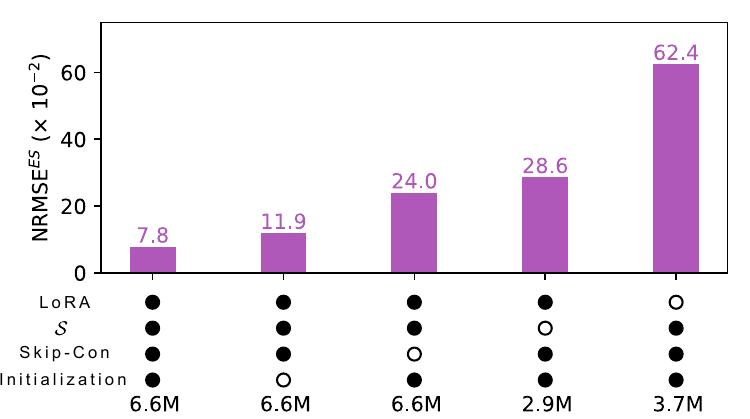}
        \caption{
            NRMSE$^{ES}$ of Tadpole-B fine-tuned with different
            Tadpole-DFT components. All components contribute
            meaningfully to overall performance.
        }
        \label{fig:dynamics_ablation}
    \end{minipage}
\end{figure}

\textbf{LoRA Rank vs Capacity of \subnet{}:} \cref{fig:dynamics_size_ablation} presents an ablation study on fine-tuning capacity for the dynamics learning task. We evaluate the Tadpole-DFT strategy under different configurations by varying the LoRA rank (LoRA 16 and LoRA 64) while keeping the sub-network size fixed, and by varying the sub-network size while fixing the LoRA rank. While increasing LoRA rank consistently improves performance, the size of the sub-network has little effect on performance. In \cref{sec:exact_metric_values}, we provide NRMSE values in physics space, where increasing the sub-network size slightly improves the prediction accuracy,  still showing less effect than the LoRA rank.
Thus, model performance benefits more strongly from  LoRA rank than from the sub-network's capacity.

\textbf{DFT Components:}
Our proposed Tadpole-DFT strategy consists of three major ingredients. To evaluate their individual efficacy, we conducted an ablation study with results presented in \cref{fig:dynamics_ablation}. We evaluate the performance of Tadpole-B fine-tuned with different Tadpole-DFT variants, including 
removing the latent transformation sub-network, removing the reintroduced skip connections, and freezing the backbone without LoRA fine-tuning. The results indicate that each component contributes meaningfully to overall performance, and removing any single component results in a noticeable increase in NRMSE$^{ES}$. Although previous analysis in \cref{fig:dynamics_size_ablation} suggests that varying the sub-network size has a relatively limited impact on final performance, completely removing the sub-network results in a 4x increase in error. In addition, reintroducing skip connections incurs almost no increase in trainable parameters while improving performance by 68\%, highlighting their effectiveness in enhancing information flow across scales during dynamics learning. Meanwhile, we also evaluated a pure latent dynamics variant, in which a network with the same architecture as \subnet{} was trained to predict directly in the latent space encoded by the best-performing FPFT Tadpole encoder. The results are summarized in \cref{tab:dynamics_bl_spectrum,tab:dynamics_bl_nrmse,tab:dynamics_iso_spectrum,tab:dynamics_iso_nrmse} in the Appendix. This latent dynamics variant performed significantly worse than all Tadpole-DFT variants, despite having more trainable parameters. These results highlight the limitations of relying exclusively on the latent space for dynamic prediction and emphasize the necessity of utilizing all components of Tadpole-DFT.

\subsection{Generative modeling}
\singlecolumnformatedtable{Statistical evaluation of generated samples. \textbf{Bold} and \underline{underlined } text shows the best and second-best values, respectively. Tadpole with FPFT fine-tuning performed best across all metrics.}{tab:tcf_generative}{
\begin{tabular}{cccccccc}
\toprule
\multirow{2}{*}{Model} & \multirow{2}{*}{$\chi^2_\mathrm{PQM} \downarrow$} & $\mathcal{W}_1 \downarrow$ & $\text{MMD}_{\text{RBF}} \downarrow $ & \multicolumn{2}{c}{NRMSE ($\times 10^{-2}$) $\downarrow$} & Rel.       & Trainable                       \\ \cline{5-6}
                       &                                                   & $(\times 10^{-2})$         & $(\times 10^{-2})$                    & Mean                        & Std.                       &    Time                        & Params                          \\ \midrule
UNet$_\mathrm{GenCFD}$ & 395.5                                             & \underline{0.63}           & 0.74                                  & 11.49                        & \underline{26.81}           & 183.72                     & 100.0M                          \\AFNO                   & 373.9                                             & 3.37                       & 18.60                                 & 5.35                        & 274.19                      & \underline{1.81}           & 64.1M                           \\
AViT                   & 557.78                                            & 36.53                      & 19.45                                 & 12.57                        & 1013.51                      & 10.39                      & \underline{60.0M}               \\ \hline
Scratch                & 1775.7                                            & 7.08                       & 25.32                                 & 40.11                       & 37.30                       & \multirow{3}{*}{\bf{1.00}} & \multirow{3}{*}{\textbf{12.3M}} \\
FPFT                   & \bf{181.7}                                        & \bf0.47                    & \bf0.13                               & \bf{0.76}                   & \bf{20.16}                  &                            &                                 \\
LoRA 32                & \underline{256.5}                                 & 0.81                       & \underline{0.47}                      & \underline{2.28}            & 32.06                       &                            &                                 \\ \bottomrule
\end{tabular}

}

In this section, we evaluate Tadpole as a backbone for generative modeling of 3D turbulent flows. We focus on the \tcf{} dataset, which exhibits complex, anisotropic flow structures. We implement a latent generative model based on flow matching, in which a 12.3M network with the same architecture as \subnet{} is trained in the latent space defined by the Tadpole encoder to generate realistic 3D \tcf{} fields. During inference, the latent flow matching model generates latent samples, which are subsequently decoded into high-fidelity 3D \tcf{} fields using the Tadpole decoder. We consider three Tadpole-B variants: trained from scratch, fine-tuned with FPFT, and fine-tuned with LoRA-32 on the \tcf{} autoencoding task, as described in \cref{sec:downsream_autoencoding}. These Tadpole-based latent generative models are compared with baselines trained to generate samples directly in physical space, without leveraging pre-trained backbones. We introduce several metrics to evaluate the model performance, including the $\chi^2_\mathrm{PQM}$~\cite{DBLP:conf/iclr/LemosSMSSLH25}, Wasserstein-1 distance $\mathcal{W}_1$, Maximum Mean Discrepancy with a Radial Basis Function kernel $\text{MMD}_\text{RBF}$ \cite{DBLP:journals/jmlr/GrettonBRSS12}, and the NRMSE of the mean and standard deviation of the distribution. Details of these metrics can be found in \cref{section:stat_tcf}. Meanwhile, we also evaluate the relative sample generation time (Rel. Time) normalized to the best-performing model. Fine-tuned Tadpole substantially improves generative modeling performance, as shown in \cref{tab:tcf_generative}. The latent generative model built upon the FPFT Tadpole achieves the best performance across all metrics. The LoRA 32 variant is the second-best model except for the $\mathcal{W}_1$ and the Std. metrics, where UNet$_\mathrm{GenCFD}$ is better but 183 times slower. These results highlight the advantages and effectiveness of the latent representation learned with Tadpole's pre-training.

\section{Conclusions, limitations and outlook\label{sec:conclustion_outlook}}
We have introduced Tadpole, a foundation model for 3D PDEs that leverages an efficient crop-based training strategy and a novel online pre-training framework using synthetic data generators. Tadpole is pre-trained as a variational autoencoder on a diverse set of 3D PDEs and can be effectively fine-tuned for various downstream tasks, including autoencoding, dynamics learning, and generative modeling. Extensive experiments demonstrate Tadpole's strong zero-shot reconstruction capabilities and its ability to achieve impressive performance across multiple tasks.  

At the same time, several limitations exist: as the approach focuses on regular grids, unstructured grids are a natural extension. While our work, like other approaches, focuses on short-term rollouts, long-term predictions represent an important challenge for all scientific foundation models. Likewise, despite the high accuracy, even larger Tadpole models should be pre-trained and evaluated. In the future, it will be highly interesting to evaluate Tadpole's capacity to predict a broader range of physical systems~\cite{ohana2024well}, to couple with differentiable solvers~\cite{um2020sol,thuerey2021pbdl,pbfm2025}, and to combine the framework with active learning techniques~\cite{active_learning,pestourie2020active}.

\section*{Acknowledgements and disclosures}
Qiang Liu acknowledges the support from the China Scholarship Council (No.202206120036) for his Ph.D. research at the Technical University of Munich. The authors gratefully acknowledge the scientific support and HPC resources provided by the Erlangen National High Performance Computing Center (NHR@FAU) of the Friedrich-Alexander-Universität Erlangen-Nürnberg (FAU) under the NHR project b278bb. NHR funding is provided by federal and Bavarian state authorities. The authors also gratefully acknowledge the computational and data resources as well as the support provided by the Leibniz Supercomputing Centre. The authors also acknowledge the EuroHPC Joint Undertaking for providing access to the EuroHPC supercomputer LEONARDO, hosted by CINECA (Italy) and the LEONARDO consortium. The authors also acknowledge support from the NVIDIA Academic Grant Program for using four RTX PRO 6000 GPUs.

\bibliographystyle{unsrtnat}
\bibliography{main}

\newpage
\appendix

\section*{Technical appendices and supplementary material}
This appendix provides details and background information for Tadpole on the following topics:

\begin{itemize}
    \setlength\itemsep{-1pt} 
    \item \Cref{sec:additional_results_visualizations}:  Additional Analysis, Results, and Visualizations.
    \item \Cref{sec:dataset_details}: Dataset and online training setups
    \item \Cref{sec:training_details}: Training details and network architectures
    \item \Cref{sec:evaluation_metrics}: Evaluation metrics
    \item \Cref{sec:nomenclature_abbreviations}: Nomenclature and abbreviations
    \item 
    \Cref{sec:changelog}: Changelog
\end{itemize}

\section{Additional analysis, results, and visualizations \label{sec:additional_results_visualizations}}
\subsection{On the geometric complexity of reconstruction and dynamics learning\label{sec:discussion_mainfold}}
Below, we explain why the reconstruction objective is easier to learn compared to the dynamics objective from a manifold point of view.

\begin{figure}[htbp]
    \centering
    \includegraphics[scale=0.56]{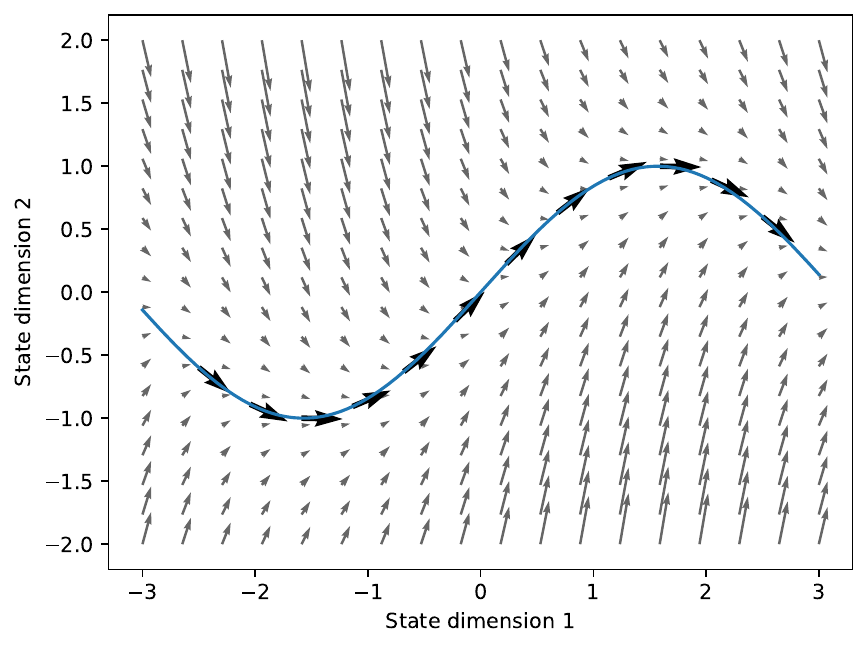}
    \caption{
    Illustration of the distinction between learning the solution manifold and learning the induced dynamics. The curve represents a low-dimensional solution manifold $\mathcal{M}$ embedded in a higher-dimensional space. Tangent vectors along the manifold correspond to the intrinsic dynamics $F(u)\in T_u\mathcal{M}$, while the surrounding vector field illustrates the additional requirement of maintaining consistency with the manifold by attracting perturbed states back toward M. Reconstruction-based learning only needs to learn the geometry of $\mathcal{M}$, whereas dynamics learning also requires learning both tangent and surrounding vectors.
    }
    \label{fig:mainfold}
\end{figure}

Let $\mathcal{H}$ and $\mathcal{M}\subset \mathcal{H}$ denote a high-dimensional function space and a set of admissible PDE solution states $u$, respectively. $\mathcal{M}$ is typically concentrated near a low-dimensional, smooth manifold due to spatial coupling induced by differential operators. Training with the reconstruction target is equivalent to learning a coordinate chart and its inverse on $\mathcal{M}$:
\begin{equation}
\mathcal{E}: \mathcal{M} \rightarrow \mathbb{R}^k,\; \mathcal{D}:\mathbb{R}^k \rightarrow \mathcal{M},\; \mathcal{D}\circ\mathcal{E}\approx\text{Id}_\mathcal{M},
\end{equation}
where $\mathbb{R}^k$ denotes a k-dimensional Euclidean latent space. The above learning target is primarily governed by the geometric features, e.g., dimensionality and regularity, of the solution manifold $\mathcal{M}$ rather than the high-dimensional function space $\mathcal{H}$. In particular, when $\mathcal{M}$ is a smooth manifold, the associated coordinate maps can also be chosen to be smooth, and their neural approximations, i.e., \encoder{} and \decoder{}, can be small and simple. 

In contrast, the PDE evolution can be treated as a dynamical system on $\mathcal{M}$:
\begin{equation}
    \frac{du}{dt}=F(u),
\end{equation}
where $F(u)\in T_u\mathcal{M}$ is a tangent vector assigned to each state $u\in \mathcal{M}$.  Learning a one-step prediction (i.e., \current{} $\rightarrow$ \future{}) corresponds to learning a flow map generated through integrating $F$, which is more challenging for many reasons: 

\begin{itemize}
    \item While reconstruction requires approximating the geometry of $\mathcal{M}$, dynamical prediction also requires learning additional vector fields defined on $\mathcal{M}$ besides the manifold's geometry.
    \item In addition to approximating the tangent vector $F$ on $\mathcal{M}$, a learned dynamics model must remain stable under perturbations, e.g., it should also learn the surrounding vectors, as shown in \cref{fig:mainfold}, to map states in a neighborhood of $M$ back toward $M$, thereby preserving the manifold’s invariance.
    \item Although $\mathcal{M}$ can be smooth and low-dimensional, the vector field $F$ and corresponding surrounding vector fields could still exhibit strong variability, particularly when PDEs have strong nonlinear interactions or multiscale effects.
\end{itemize}

Consequently, reconstruction-based objectives primarily learn the geometric structure of the solution manifold, whereas dynamical models must additionally resolve the induced flow on this manifold. This distinction helps explain why reconstruction-based training is typically easier and exhibits better generalization properties than  directly learning PDEs time-stepping.

\subsection{Methodological Summary of the Tadpole approach}

The discussion of the advantages of reconstruction-based objectives for general \textit{representation learning} highlights key advantages of the proposed approach. 
\begin{tcolorbox}[colback=blue!6!white] 
To summarize, Tadpoles distinguishes itself from existing approaches for scientific foundation models in the following ways:
\begin{itemize} 
\item It focuses on \textit{autoencoding} as generalizable, central objective for represetnation learning.
\item Tadpole employs \textit{online data-generation} with a fast, semi-spectral, GPU-based solver, circumventing storage and I/O bottlenecks.
\item It comes with a highly \textit{flexible architecture}, e.g., supporting arbitrary numbers of channels, and temporal dynamics (via learned skip-connections) for downstream tasks. 
\item Tadpole's capabilities are demonstrated for a wide range of downstream applications, from reconstruction, over generative modeling to temporal dynamics. 
\end{itemize}
\end{tcolorbox}

\vspace{5pt}

These aspects also provide important distinctions from existing approaches for foundation models, in particular from large language models: the representations of LLMs typically emerge implicitly from the next-token prediction task using a pre-determined tokenization codebook. Tadpole instead explicitly optimizes for representation learning via autoencoding. Specifically, we induce meaningful representations by combining the reconstruction task with large streams of synthetic PDE data, such that the resulting latent space captures the low-dimensional solution manifold and allows for generalizing transfers to new downstream tasks. This is more closely aligned with the learned latent spaces of imaging and video FMs, which, in contrast, typically focus on perceptually-driven latent spaces.
\clearpage

\subsection{Comparison between crop-based inference and whole-domain inference \label{sec:additional_results_visualizations:whole_domain_inference}}

\begin{figure}[htbp]
    \centering
    \includegraphics[scale=0.48]{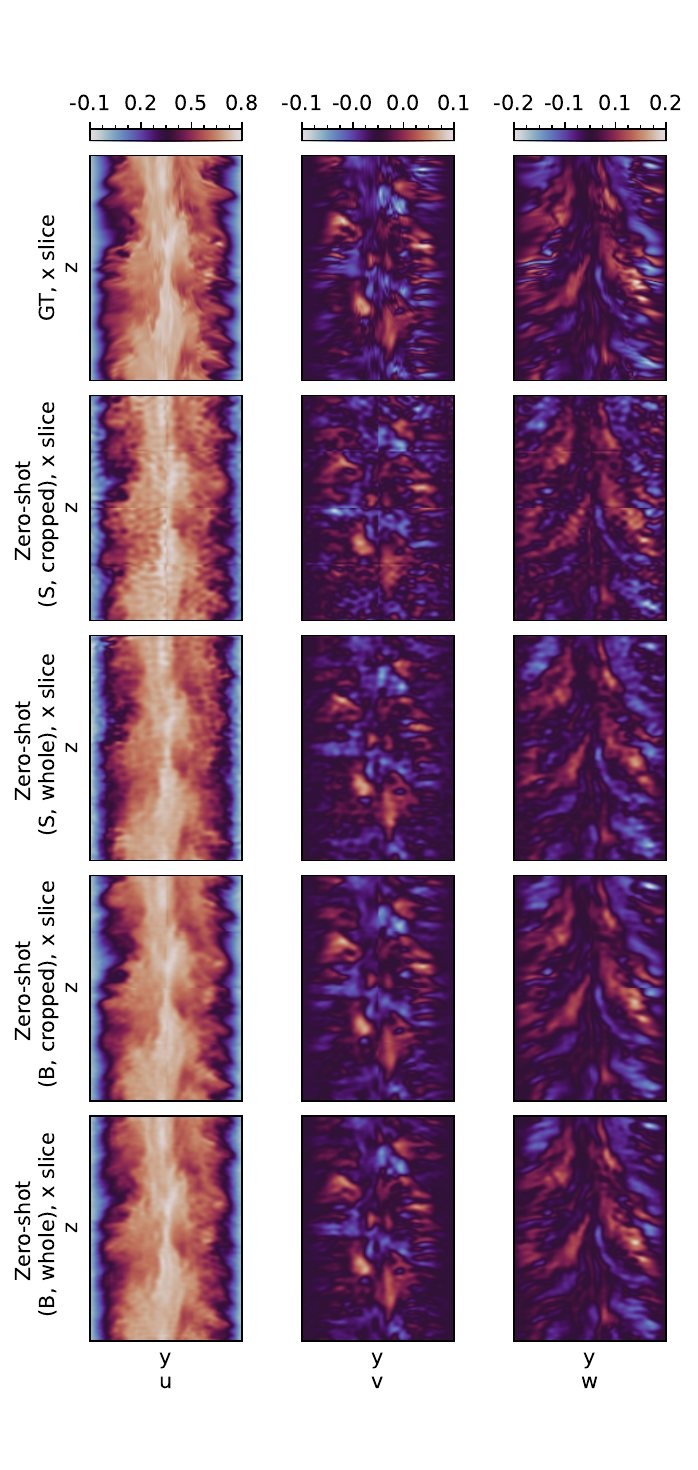}\hspace{20pt}
    \includegraphics[scale=0.48]{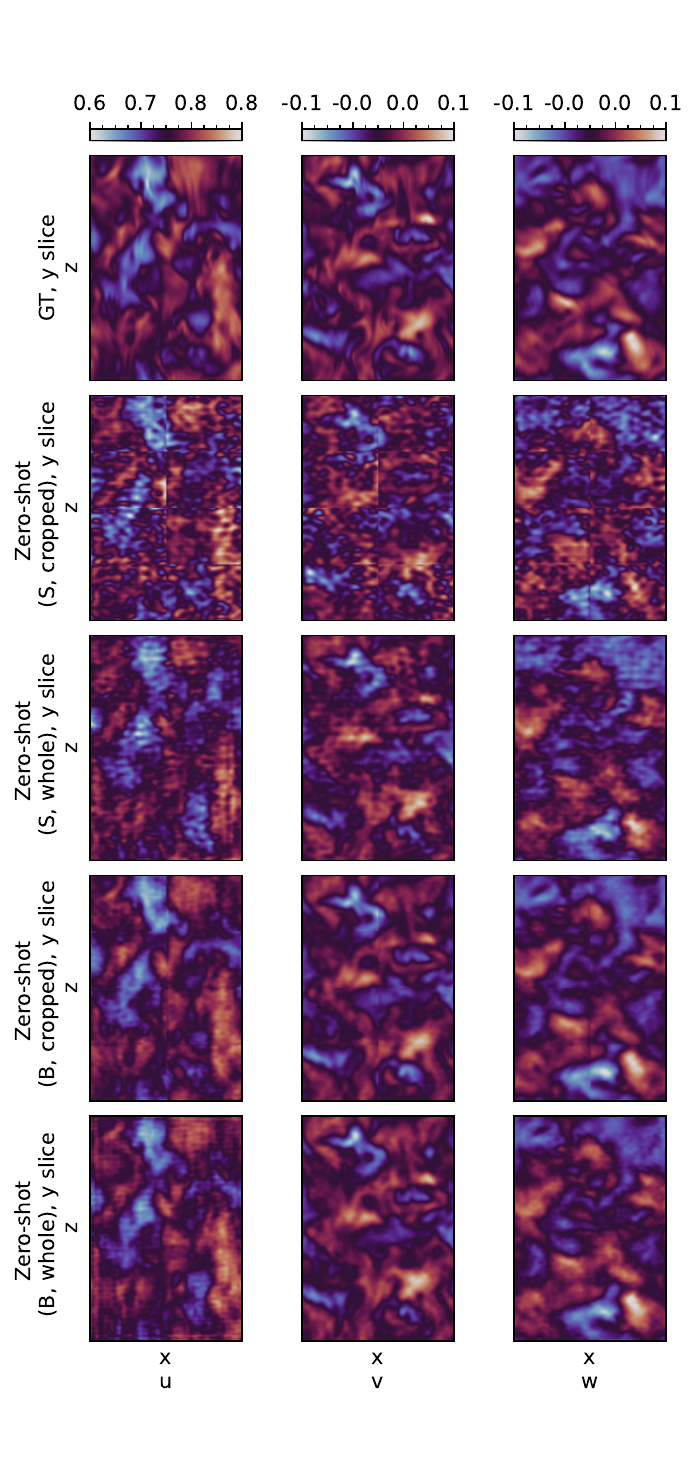}
    \caption{
    Visualization of the reconstruction of \tcf{} with crop-based and whole-domain inference for slices $x=X/2$ (left), and $y=Y/2$ (right). 
    }\label{fig:vis_recon_full_crop_blxs}
\end{figure}

\begin{figure}[p]
    \centering
    \includegraphics[scale=0.53]{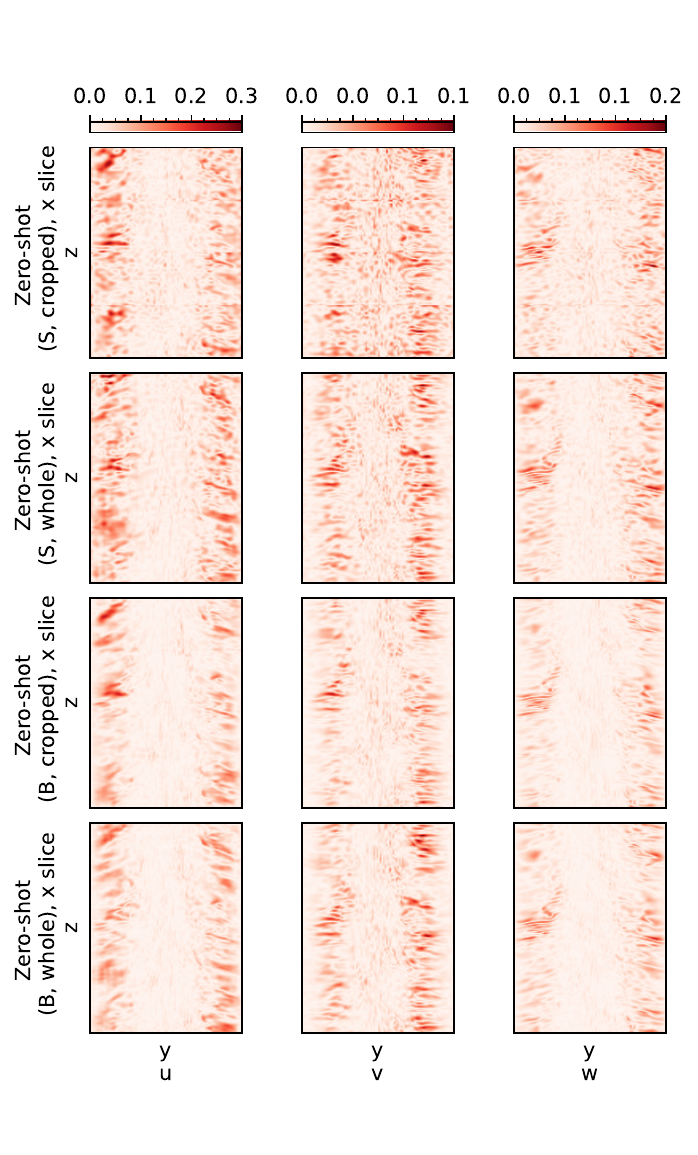}\hspace{20pt}
    \includegraphics[scale=0.53]{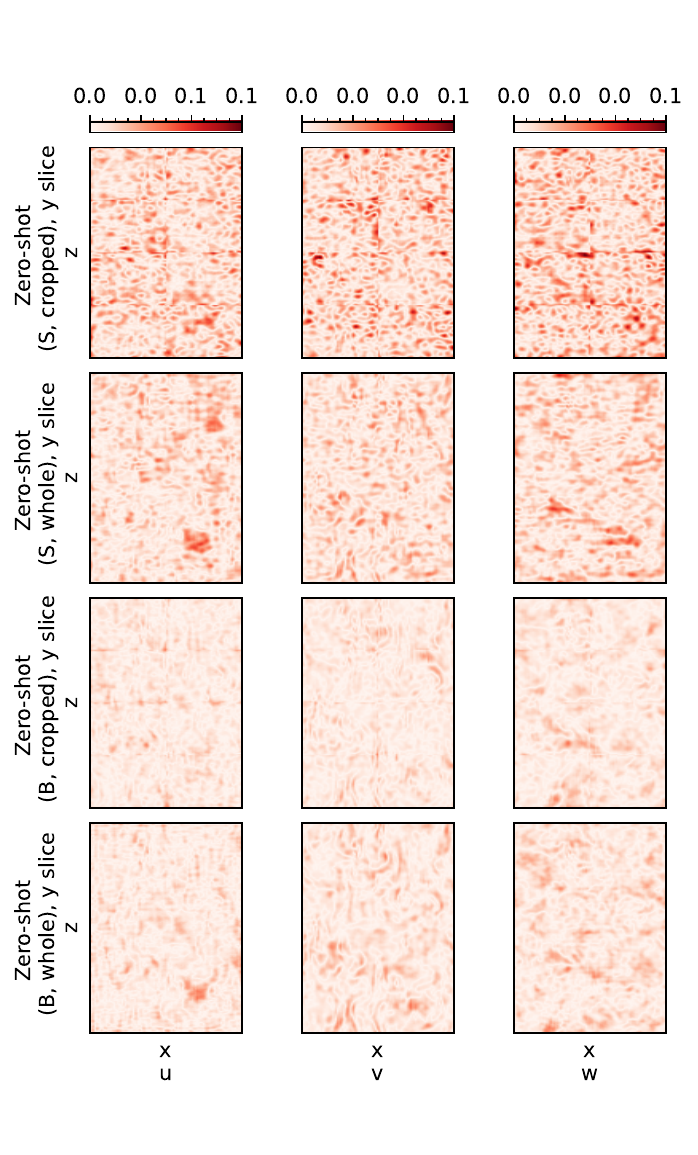}
    \caption{
    Visualization of the absolute error for the reconstruction of \tcf{} with crop-based and whole-domain inference for slices $x=X/2$ (left), and $y=Y/2$ (right).
    }
    \label{fig:vis_recon_full_crop_blxs_error}
\end{figure}

We compare the zero-shot reconstruction RMSE of the Tadpole B-size model across \tcf{} datasets for crop-based and whole-domain inference. For the S-size model, the reconstruction error slightly improved by 2.8\% with whole-domain inference. For the B-size model, the performance degrades slightly by 0.8\%. \cref{fig:vis_recon_full_crop_blxs}
shows the corresponding visualizations of the reconstructions with whole-domain inference and crop-based inference, and \cref{fig:vis_recon_full_crop_blxs_error} shows the corresponding absolute error. We can see that the whole-domain approach helps to remove the inconsistency near crop boundaries in the S-model zero-shot. But in general, the zero-shot reconstruction RMSE on the \tcf{} dataset is not significantly affected by the inference strategy, highlighting Tadpole's strong generalization across different resolutions.

\FloatBarrier

\subsection{Convergence curve of FPFT and LoRA fine-tuning on dynamics learning}
\begin{figure}[htbp]

    \centering
    \includegraphics[width=0.59\linewidth]{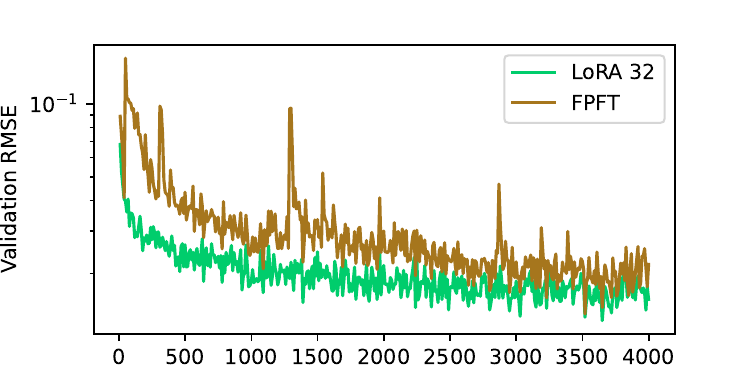}
    \caption{
        Validation RMSE of Tadpole fine-tuned with LoRA-32 and FPFT.
        LoRA yields stable training with reduced errors.
    }
    \label{fig:val_rmse_lora32_pretrained}
\end{figure}

\cref{fig:val_rmse_lora32_pretrained} shows the validation RMSE during FPFT and LoRA fine-tuning. FPFT fine-tuning exhibits many oscillations during training, especially at the start, whereas LoRA fine-tuning remains stable. This highlights LoRA fine-tuning's ability to preserve pretrained knowledge, thereby avoiding training instabilities.

\subsection{Ablation study of dynamics learning in pixel space}
\begin{figure}[htbp]

    \centering
    \includegraphics[width=0.5\linewidth]{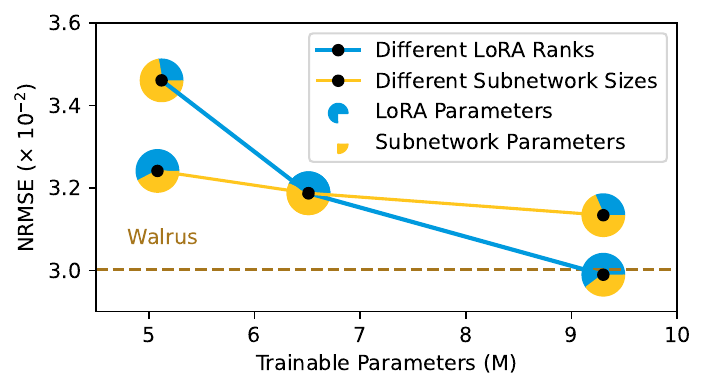}
    \caption{
        One-step NRMSE of the Tadpole-B model with different sub-network sizes and LoRA ranks. Especially the latter positively affects performance.
    }
    \label{fig:dynamics_size_ablation_nrmse}
\end{figure}

\begin{figure}[htbp]
    \centering
    \includegraphics[width=0.5\linewidth]{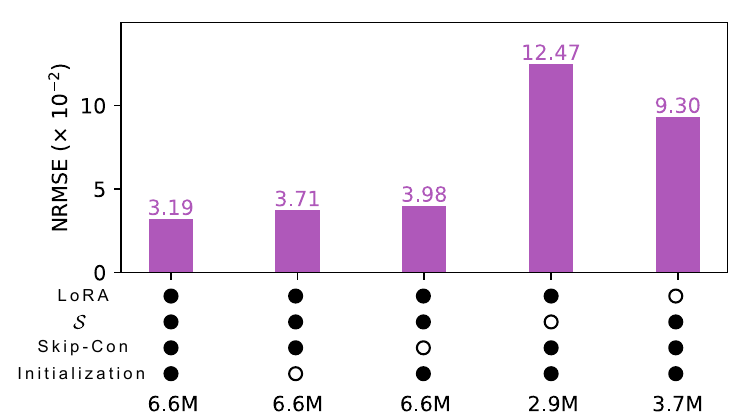}
    \caption{
        One-step NRMSE of Tadpole-B fine-tuned with different Tadpole-DFT components. 
    }
    \label{fig:dynamics_ablation_nrmse}
\end{figure}

In \cref{fig:dynamics_size_ablation} and \cref{fig:dynamics_ablation}, we perform ablation studies in spectrum space, showing that the sub-network has little effect on model performance. Here, we provide additional evaluations in the pixel space. \cref{fig:dynamics_size_ablation_nrmse} and \cref{fig:dynamics_ablation_nrmse} compare the model performance with different sub-network sizes, LoRA ranks, and DFT components. Compared to the results in spectrum space, the effect of the sub-network is more evident in physics space, where improving the sub-network size clearly decreases the NRMSE of the prediction, although increasing the LoRA rank is more efficient. Dropping the sub-network results in the largest increase in pixel-space error compared to removing other components. 

The above results suggest that LoRA has a greater impact on performance in the spectrum space, while the sub-network has a greater impact in the pixel space. This is because LoRA works with the model backbone, where convolutional and vision transformer layers are applied to learn spatial correlations. Improving backbone performance yields a better-learned spatial pattern of PDE solutions. While the sub-network learns in the latent space, spatial correlations are more difficult to learn because of the backbone's downsampling layers. It focuses more on improving model performance by combining information from state variables. Thus, although it can improve performance in pixel space, the spatial patterns of PDE solutions are only slightly affected.

\newpage

\subsection{Detailed metric values of the main experiments\label{sec:exact_metric_values}}
In this section, we summarize the detailed metric values from the previous experiments. \cref{tab:downstream_autoencoder} shows the reconstruction NRMSE of Tadpole models on the autoencoding task. Corresponding results are illustrated in \cref{fig:reconstruction_bar_plot}. \cref{tab:dynamics_bl_spectrum}-\cref{tab:dynamics_iso_nrmse} summarize the NRMSE$^{ES}$ and NRMSE of different methods in dynamics tasks. Corresponding results are illustrated in \cref{fig:dynamics_bar_plot}. Here, we also compare the model's performance with Swin3D~\cite{swinv22022}, AViT~\cite{mpp2024}, and AFNO~\cite{afno2021}, all trained from scratch, where Tadpole and other foundation models outperform them in $NRMSE^{ES}$.

\floatingformatedtable{Tadpole's reconstruction NRMSE ($\times 10^{-2}$) on downstream autoencoding tasks. The zero-shot model surpasses a from-scratch model across three datasets, highlighting the efficacy of pre-training. Fine-tuning further enhances the performance.
}{tab:downstream_autoencoder}{
\begin{tabular}{cccccccc}
\toprule
                           &                    &         & \iso            & \tcf            & \tbl            & \mhd            & \begin{tabular}[c]{@{}c@{}}Trainable\\ Params\end{tabular} \\ \midrule
\multirow{7}{*}{Zero Shot} & \multicolumn{2}{c}{S}        & 6.11            & 9.82            & 5.93            & 7.22            & \multirow{7}{*}{\textbackslash}                            \\
                           & \multicolumn{2}{c}{B}        & 3.23            & 7.87            & 1.27            & 3.42            &                                                            \\
                           & \multicolumn{2}{c}{L}        & 2.52            & 7.27            & 1.17            & \underline {2.93} &                                                            \\ \cline{2-7}
                           & \multirow{4}{*}{B} & 1 PDE   & 112.36          & 14.57           & 18.88           & 6.48            &                                                            \\
                           &                    & 3 PDEs  & 3.35            & 8.50            & 1.86            & 3.57            &                                                            \\
                           &                    & Initial & 3.75            & 8.20            & 1.63            & 4.04            &                                                            \\
                           &                    & Local   & 4.14            & 7.96            & 1.79            & 4.21            &                                                            \\ \hline
Scratch                    &                    &         & 7.17            & 4.40            & 1.67            & 8.22            & \underline {38.1M}                                            \\
FPFT                       & \multicolumn{2}{c}{B}        & \bf 2.73        & \bf 2.94        & \bf 0.50        & \bf 2.50        & \underline {38.1M}                                            \\
LoRA-32                    &                    &         & \underline {3.01} & \underline {3.82} & \underline {0.78} & 3.18            & \textbf{2.8M}                                              \\ \bottomrule
\end{tabular}
}

\begin{landscape}
\formatedtable{Rollout NRMSE${^{ES}}$ ($\times 10^{-2}$) of different models on the dynamics downstream task for \tbl{} dataset. Tadpole with DFT fine-tuning strategy outperforms all other specialized and foundational models. }{tab:dynamics_bl_spectrum}
{
\begin{tabular}{ccccccccccccc}
\toprule
Rollout steps & 1 & 2 & 3 & 4 & 5 & 6 & 7 & 8 & 9 & 10 & Average & Trainable Params (M)\\
\midrule
Swin3D & 1.21 & 1.96 & 2.49 & 2.87 & 3.16 & 3.36 & 3.64 & 3.95 & 4.28 & 4.67 & 3.16 & 50.3M\\
AViT & 7.99 & 8.47 & 9.21 & 10.00 & 10.77 & 11.56 & 12.36 & 12.89 & 13.64 & 14.21 & 11.11 & 60.0M\\
AFNO & 3.62 & 4.95 & 6.53 & 8.19 & 9.82 & 11.19 & 12.24 & 13.17 & 14.02 & 14.78 & 9.85 & 64.1M\\
\midrule
MorphNet-S & 10.95 & 11.89 & 12.36 & 12.98 & 13.22 & 13.35 & 13.83 & 14.29 & 14.69 & 15.04 & 13.26 & 32.8M\\
DPOT-S & 11.96 & 13.60 & 14.80 & 15.92 & 16.64 & 17.20 & 18.01 & 18.70 & 19.41 & 20.05 & 16.63 & 49.5M\\
\midrule
Walrus & 2.44 & 2.92 & 2.97 & 3.05 & 3.68 & 4.45 & 5.55 & 6.76 & 8.20 & 9.58 & 4.96 & 1300.0M\\
\midrule
Tadpole-B-Scratch & 0.94 & 1.83 & 2.71 & 3.58 & 4.48 & 5.37 & 6.42 & 7.62 & 8.80 & 10.05 & 5.18 & 41.6M\\
Tadpole-B-FPFT & 1.16 & 2.24 & 3.31 & 4.35 & 5.40 & 6.36 & 7.36 & 8.37 & 9.31 & 10.27 & 5.81 & 41.6M\\
Latent Dynamics & 51.84 & 65.74 & 69.72 & 71.16 & 71.69 & 71.94 & 72.00 & 72.39 & 72.26 & 72.51 & 69.12 & 7.4M\\
\midrule
Tadpole-S-LoRA-32 & 2.02 & 3.47 & 4.69 & 5.85 & 6.92 & 7.86 & 8.85 & 9.80 & 10.65 & 11.44 & 7.16 & \textbf{3.0M}\\
Tadpole-B-LoRA-16 & 1.37 & 2.38 & 3.27 & 4.14 & 4.91 & 5.57 & 6.28 & 6.93 & 7.53 & 8.09 & 5.05 & \underline{5.1M}\\
Tadpole-B-LoRA-32 & 0.93 & 1.64 & 2.28 & 2.89 & 3.39 & 3.76 & 4.19 & 4.56 & 4.89 & 5.18 & 3.37 & 6.5M\\
Tadpole-B-LoRA-64 & \underline{0.65} & \underline{1.19} & \underline{1.68} & \underline{2.17} & \underline{2.59} & \textbf{2.94} & \underline{3.32} & \underline{3.67} & \underline{3.97} & \underline{4.25} & \underline{2.64} & 9.3M\\
Tadpole-L-LoRA-32 & \textbf{0.58} & \textbf{1.10} & \textbf{1.62} & \textbf{2.13} & \textbf{2.57} & \underline{2.94} & \textbf{3.31} & \textbf{3.65} & \textbf{3.93} & \textbf{4.21} & \textbf{2.60} & 13.6M\\
\bottomrule
\end{tabular}
}

\formatedtable{Rollout NRMSE ($\times 10^{-2}$) of different models on the dynamics downstream task for \tbl{} dataset. Tadpole with DFT fine-tuning strategy outperforms Walrus in one-step prediction with much fewer trainable parameters. }{tab:dynamics_bl_nrmse}
{
\begin{tabular}{ccccccccccccc}
\toprule
Rollout steps & 1 & 2 & 3 & 4 & 5 & 6 & 7 & 8 & 9 & 10 & Average & Trainable Params (M)\\
\midrule
Swin3D & 0.40 & 1.16 & \textbf{1.73} & \textbf{2.44} & \textbf{3.11} & \textbf{3.74} & \textbf{4.30} & \textbf{4.82} & \textbf{5.36} & \textbf{5.86} & \textbf{3.29} & 50.3M\\
AViT & 0.87 & 1.58 & 2.39 & 3.16 & 3.90 & 4.61 & 5.24 & 5.83 & 6.41 & 6.94 & 4.09 & 60.0M\\
AFNO & 0.52 & 1.20 & 2.01 & 2.76 & 3.50 & 4.20 & 4.83 & 5.42 & 6.00 & 6.54 & 3.70 & 64.1M\\
\midrule
MorphNet-S & 0.95 & 1.55 & 2.26 & 2.96 & 3.66 & 4.33 & 4.94 & 5.48 & 6.02 & 6.52 & 3.87 & 32.8M\\
DPOT-S & 1.01 & 1.60 & 2.31 & 3.00 & 3.70 & 4.38 & 5.00 & 5.58 & 6.17 & 6.71 & 3.95 & 49.5M\\
\midrule
Walrus & 0.49 & 1.09 & \underline{1.80} & \underline{2.49} & \underline{3.17} & \underline{3.84} & \underline{4.43} & \underline{4.98} & \underline{5.57} & \underline{6.05} & \underline{3.39} & 1300.0M\\
\midrule
Tadpole-B-Scratch & 0.49 & 1.30 & 2.41 & 3.62 & 4.91 & 6.20 & 7.44 & 8.67 & 9.87 & 11.04 & 5.59 & 41.6M\\
Tadpole-B-FPFT & 0.57 & 1.42 & 2.61 & 3.78 & 4.91 & 5.97 & 6.93 & 7.83 & 8.70 & 9.54 & 5.22 & 41.6M\\
Latent Dynamics & 11.30 & 15.48 & 18.54 & 20.17 & 21.64 & 22.83 & 23.64 & 23.89 & 24.52 & 24.79 & 20.68 & 7.4M\\
\midrule
Tadpole-S-LoRA-32 & 0.55 & 1.35 & 2.27 & 3.14 & 3.99 & 4.81 & 5.58 & 6.32 & 7.06 & 7.79 & 4.29 & \textbf{3.0M}\\
Tadpole-B-LoRA-16 & 0.46 & 1.17 & 2.07 & 2.93 & 3.79 & 4.62 & 5.39 & 6.13 & 6.86 & 7.58 & 4.10 & \underline{5.1M}\\
Tadpole-B-LoRA-32 & 0.41 & 1.11 & 2.05 & 2.95 & 3.81 & 4.61 & 5.32 & 5.99 & 6.65 & 7.27 & 4.02 & 6.5M\\
Tadpole-B-LoRA-64 & \underline{0.38} & \underline{1.05} & 1.90 & 2.70 & 3.46 & 4.18 & 4.85 & 5.47 & 6.09 & 6.69 & 3.68 & 9.3M\\
Tadpole-L-LoRA-32 & \textbf{0.34} & \textbf{1.01} & 1.90 & 2.78 & 3.62 & 4.43 & 5.18 & 5.87 & 6.54 & 7.19 & 3.89 & 13.6M\\
\bottomrule
\end{tabular}
}
\end{landscape}
\newpage
\begin{landscape}
\formatedtable{Rollout NRMSE${^{ES}}$ ($\times 10^{-2}$) of different models on the dynamics downstream task for \iso{} dataset. Besides the Walrus with two orders of magnitude more parameters, Tadpole with DFT fine-tuning strategy outperforms all other specialized and foundational models. }{tab:dynamics_iso_spectrum}{
\begin{tabular}{ccccccccccccc}
\toprule
Rollout steps & 1 & 2 & 3 & 4 & 5 & 6 & 7 & 8 & 9 & 10 & Average & Trainable Params (M)\\
\midrule
Swin3D & 30.23 & 37.46 & 40.14 & 41.65 & 42.69 & 43.45 & 44.13 & 45.01 & 45.40 & 46.19 & 41.64 & 50.3M\\
AViT & 36.57 & 49.76 & 73.28 & 102.54 & 135.52 & 171.18 & 208.46 & 243.67 & 281.19 & 309.62 & 161.18 & 60.0M\\
AFNO & 25.45 & 23.27 & 24.87 & 27.62 & 30.80 & 33.93 & 36.84 & 38.69 & 41.24 & 42.17 & 32.49 & 64.1M\\
\midrule
MorphNet-S & 46.74 & 50.61 & 52.99 & 55.04 & 56.91 & 58.64 & 60.31 & 62.04 & 63.38 & 64.74 & 57.14 & 32.8M\\
DPOT-S & 36.35 & 36.96 & 37.27 & 37.69 & 38.06 & 38.39 & 38.71 & 39.22 & 39.40 & 39.86 & 38.19 & 49.5M\\
\midrule
Walrus & \underline{3.74} & \textbf{4.04} & \textbf{4.04} & \textbf{4.02} & \textbf{4.09} & \textbf{4.32} & \textbf{4.29} & \textbf{4.61} & \textbf{5.01} & \textbf{5.28} & \textbf{4.34} & 1300.0M\\
\midrule
Tadpole-B-Scratch & 9.81 & 12.32 & 13.94 & 15.52 & 16.93 & 18.14 & 19.18 & 20.10 & 20.63 & 20.94 & 16.75 & 41.6M\\
Tadpole-B-FPFT & 8.55 & 10.26 & 11.47 & 12.80 & 14.02 & 15.01 & 15.87 & 16.54 & 16.95 & 16.93 & 13.84 & 41.6M\\
Latent Dynamics & 61.15 & 63.74 & 65.81 & 67.77 & 69.56 & 71.15 & 72.69 & 74.23 & 75.44 & 76.69 & 69.82 & 7.4M\\
\midrule
Tadpole-S-LoRA-32 & 27.27 & 30.25 & 31.07 & 31.48 & 31.70 & 31.88 & 32.14 & 32.61 & 32.65 & 32.91 & 31.40 & \textbf{3.0M}\\
Tadpole-B-LoRA-16 & 12.35 & 14.44 & 15.36 & 16.23 & 16.97 & 17.65 & 18.27 & 19.05 & 19.53 & 20.26 & 17.01 & \underline{5.1M}\\
Tadpole-B-LoRA-32 & 7.78 & 9.60 & 10.94 & 12.35 & 13.65 & 14.83 & 15.89 & 16.90 & 17.59 & 18.18 & 13.77 & 6.5M\\
Tadpole-B-LoRA-64 & 5.35 & 6.82 & 8.07 & 9.40 & 10.62 & 11.80 & 12.98 & 14.21 & 15.08 & 16.22 & 11.05 & 9.3M\\
Tadpole-L-LoRA-32 & \textbf{3.51} & \underline{5.14} & \underline{6.77} & \underline{8.40} & \underline{9.93} & \underline{11.32} & \underline{12.57} & \underline{13.80} & \underline{14.84} & \underline{15.92} & \underline{10.22} & 13.6M\\
\bottomrule
\end{tabular}
}
\formatedtable{Rollout NRMSE ($\times 10^{-2}$) of different models on the dynamics downstream task for \iso{} dataset. Tadpole outperforms the Walrus in one-step prediction}{tab:dynamics_iso_nrmse}{
\begin{tabular}{ccccccccccccc}
\toprule
Rollout steps & 1 & 2 & 3 & 4 & 5 & 6 & 7 & 8 & 9 & 10 & Average & Trainable Params (M)\\
\midrule
Swin3D & 5.53 & 9.26 & 11.97 & 14.07 & 15.76 & \underline{17.21} & \underline{18.50} & \underline{19.70} & \textbf{20.64} & \textbf{21.70} & \underline{15.43} & 50.3M\\
AViT & 51.54 & 52.70 & 54.33 & 56.22 & 58.39 & 60.95 & 63.73 & 66.91 & 70.44 & 74.13 & 60.93 & 60.0M\\
AFNO & 6.06 & 10.19 & 13.41 & 16.13 & 18.55 & 20.80 & 22.95 & 25.01 & 26.89 & 28.93 & 18.89 & 64.1M\\
\midrule
MorphNet-S & 9.37 & 13.00 & 16.63 & 20.25 & 23.85 & 27.43 & 30.83 & 34.15 & 37.35 & 40.46 & 25.33 & 32.8M\\
DPOT-S & 6.14 & 8.36 & 10.85 & 13.41 & 16.05 & 18.74 & 21.41 & 24.06 & 26.67 & 29.36 & 17.50 & 49.5M\\
\midrule
Walrus & 3.00 & \textbf{4.67} & \textbf{6.68} & \textbf{8.92} & \textbf{11.41} & \textbf{13.99} & \textbf{16.65} & \textbf{19.59} & \underline{22.46} & \underline{25.42} & \textbf{13.28} & 1300.0M\\
\midrule
Tadpole-B-Scratch & 4.19 & 7.98 & 11.92 & 15.87 & 19.89 & 24.06 & 28.31 & 32.80 & 37.43 & 42.53 & 22.50 & 41.6M\\
Tadpole-B-FPFT & 3.66 & 6.94 & 10.47 & 14.11 & 17.90 & 21.89 & 26.04 & 30.47 & 35.05 & 40.01 & 20.65 & 41.6M\\
Latent Dynamics & 10.30 & 13.88 & 17.72 & 21.47 & 25.14 & 28.75 & 32.19 & 35.59 & 38.93 & 42.32 & 26.63 & 7.4M\\
\midrule
Tadpole-S-LoRA-32 & 4.41 & 7.28 & 10.64 & 14.36 & 18.48 & 22.99 & 27.88 & 33.31 & 39.09 & 45.66 & 22.41 & \textbf{3.0M}\\
Tadpole-B-LoRA-16 & 3.46 & 6.26 & 9.53 & 13.17 & 17.11 & 21.29 & 25.62 & 30.10 & 34.69 & 39.59 & 20.08 & \underline{5.1M}\\
Tadpole-B-LoRA-32 & 3.19 & 5.84 & 8.94 & 12.35 & 16.04 & 19.98 & 24.07 & 28.34 & 32.71 & 37.41 & 18.89 & 6.5M\\
Tadpole-B-LoRA-64 & \underline{2.99} & 5.54 & 8.54 & 11.85 & 15.47 & 19.38 & 23.50 & 27.88 & 32.36 & 37.14 & 18.47 & 9.3M\\
Tadpole-L-LoRA-32 & \textbf{2.85} & \underline{5.29} & \underline{8.07} & \underline{11.06} & \underline{14.23} & 17.59 & 21.04 & 24.58 & 28.19 & 32.08 & 16.50 & 13.6M\\
\bottomrule
\end{tabular}

}    
\end{landscape}

\subsection{Pre-training gains of other baselines \label{sec:pre_training_gain_others}}

The current from-scratch version of Tadpole already demonstrates strong advantages over pre-trained DPOT and Morph models, indicating that the P3D network itself serves as a strong baseline. To distinguish the performance gains attributable to pre-training and network architecture, \cref{tab:rollout_enstrophy,tab:rollout_direct} compare the NRMSE${^{ES}}$ and NRMSE for DPOT and Morph trained from scratch with their respective fine-tuned models in the \iso{} case. DPOT achieves average improvements of 2\% and 14\% in frequency and physics space, respectively, while Morph achieves improvements of 5\% and 15\%. For Tadpole, the corresponding improvements are 18\% in frequency space and 16\% in physics space for a LoRA 32 configuration (\cref{fig:dynamics_improvement_bar_plot,tab:dynamics_iso_spectrum,tab:dynamics_iso_nrmse}). Increasing the LoRA rank to 64 further increases the improvement in frequency space to 34\%. These gains are achieved with 6.5 million (LoRA 32) or 9.3 million (LoRA 64) trainable parameters, compared to the full network's 41.6 million. Overall, pre-training and fine-tuning the Tadpole models result in substantially greater improvements, particularly in the frequency domain, even when starting from the stronger P3D baseline.

\begin{table}[htbp]
    \centering
    \caption{Rollout NRMSE$^{ES}
    (\times 10^{-2})$ of DPOT and Morph on the
    \iso{} dynamics downstream task.}
    \label{tab:rollout_enstrophy}
    \begin{tabular}{lccccc}
        \toprule
        Rollout Step & 1 & 2 & 5 & 10 & Average 1--10 \\
        \midrule
        DPOT (Scratch)       & 36.58 & 36.97 & 38.86 & 41.53 & 39.03 \\
        DPOT (Finetuned)     & 36.35 & 36.96 & 38.06 & 39.86 & 38.19 (-2\%) \\
        Morph (Scratch)      & 55.52 & 58.22 & 60.69 & 63.68 & 60.57 \\
        Morph (Finetuned)    & 46.74 & 50.61 & 56.91 & 64.74 & 57.14 (-5\%) \\
        \bottomrule
    \end{tabular}
\end{table}

\begin{table}[htbp]
    \centering
    \caption{Rollout direct NRMSE$
    (\times 10^{-2})$ of DPOT and Morph on the
    \iso{} dynamics downstream task.}
    \label{tab:rollout_direct}
    \begin{tabular}{lccccc}
        \toprule
        Rollout Step & 1 & 2 & 5 & 10 & Average 1--10 \\
        \midrule
        DPOT (Scratch)       & 7.13 & 9.71 & 18.56 & 34.22 & 20.32 \\
        DPOT (Finetuned)     & 6.14 & 8.36 & 16.05 & 29.36 & 17.50 (-14\%) \\
        Morph (Scratch)      & 10.96 & 15.31 & 27.44 & 47.74 & 29.92 \\
        Morph (Finetuned)    & 9.37 & 13.00 & 23.85 & 40.46 & 25.33 (-15\%) \\
        \bottomrule
    \end{tabular}
\end{table}

The results for DPOT and MORPH trained from scratch yield important insights into their respective architectures. When considered alongside the previous comparison between fine-tuned Tadpole and from-scratch Tadpole (\cref{fig:dynamics_improvement_bar_plot,tab:dynamics_bl_spectrum,tab:dynamics_bl_nrmse,tab:dynamics_iso_spectrum,tab:dynamics_iso_nrmse}), these findings clarify the contributions of each component in Tadpole. Specifically, a portion of the absolute performance can be attributed to the backbone, while the controlled same-architecture comparison isolates the additional benefits provided by pretrained representations and the fine-tuning method, which account for up to 49\% improvement in LoRA-64 DFT fine-tuning in the TBL case.

\subsection{Computation Scaling Relative to Data Size \label{sec:computation_scaling}}

In this section, we evaluate the computation scaling of Tadpole models with data size. All the experiments are performed on an NVIDIA RTX PRO 6000 Blackwell GPU.

\floatingformatedtable{
Computational cost and peak allocated memory for different PDE state dimensions $D$.
}{tab:flops_memory}{
\begin{tabular}{ccc}
        \toprule
        $D/64^3$ & GFLOPs & Peak allocated memory (GB) \\
        \midrule
        1   & 10.336    & 0.339  \\
        2   & 20.672    & 0.483  \\
        4   & 41.345    & 0.772  \\
        8   & 82.689    & 1.349  \\
        16  & 165.379   & 2.503  \\
        32  & 330.758   & 4.811  \\
        64  & 661.516   & 9.429  \\
        128 & 1,323.032 & 18.664 \\
        256 & 2,646.063 & 37.131 \\
        \bottomrule
    \end{tabular}
}

\cref{tab:flops_memory} summarizes the changes in inference computational cost and peak VRAM consumption of Tadpole-B as the dimension of the 3D PDE state $D=C\times X\times Y\times Z$ increases. In this experiment, the process crop size is fixed at $64^3$, and Tadpole processes the data in a single batch, where the batch size equals $D/64^3$. The results demonstrate that inference cost increases linearly with the PDE state $D$.

\floatingformatedtable{
Trade-off between computation time and memory usage for internal minibatch inference when $D/64^3=256$. An internal batch size of 1 indicates that Tadpole processes a single crop at a time, requiring 256 batches to process the entire dataset. In contrast, an internal batch size of 256 indicates that Tadpole processes all data in a single internal batch.
}{tab:batch_time_memory}{
\begin{tabular}{ccc}
        \toprule
        Internal batch size & Median time (s)
        & Peak allocated memory (GB) \\
        \midrule
        1 & 4.169 & 0.945 \\
        2   & 3.521 & 0.980  \\
        4   & 3.491 & 1.263  \\
        8   & 3.159 & 1.834  \\
        16  & 4.142 & 2.971  \\
        32  & 3.288 & 4.799  \\
        64  & 3.036 & 8.048  \\
        128 & 3.027 & 12.854 \\
        256
            & 2.891 & 37.131 \\
        \bottomrule
    \end{tabular}
}

However, since Tadpole incorporates the channel and crop dimensions into the batch dimension, crops can be processed in internal mini-batches. \cref{tab:batch_time_memory} presents the changes in calculation time and peak memory usage as a function of internal minibatch size when $D/64^3=256$. The internal minibatch operation does not change the total number of floating-point operations, but it significantly reduces peak memory usage, with only a slight increase in computation time.

\FloatBarrier

\subsection{Visualizations of autoencoding reconstructions \label{sec:additional_results_visualizations:autoencoding}}

\begin{figure}[htbp]
    \centering
    \includegraphics[scale=0.6]{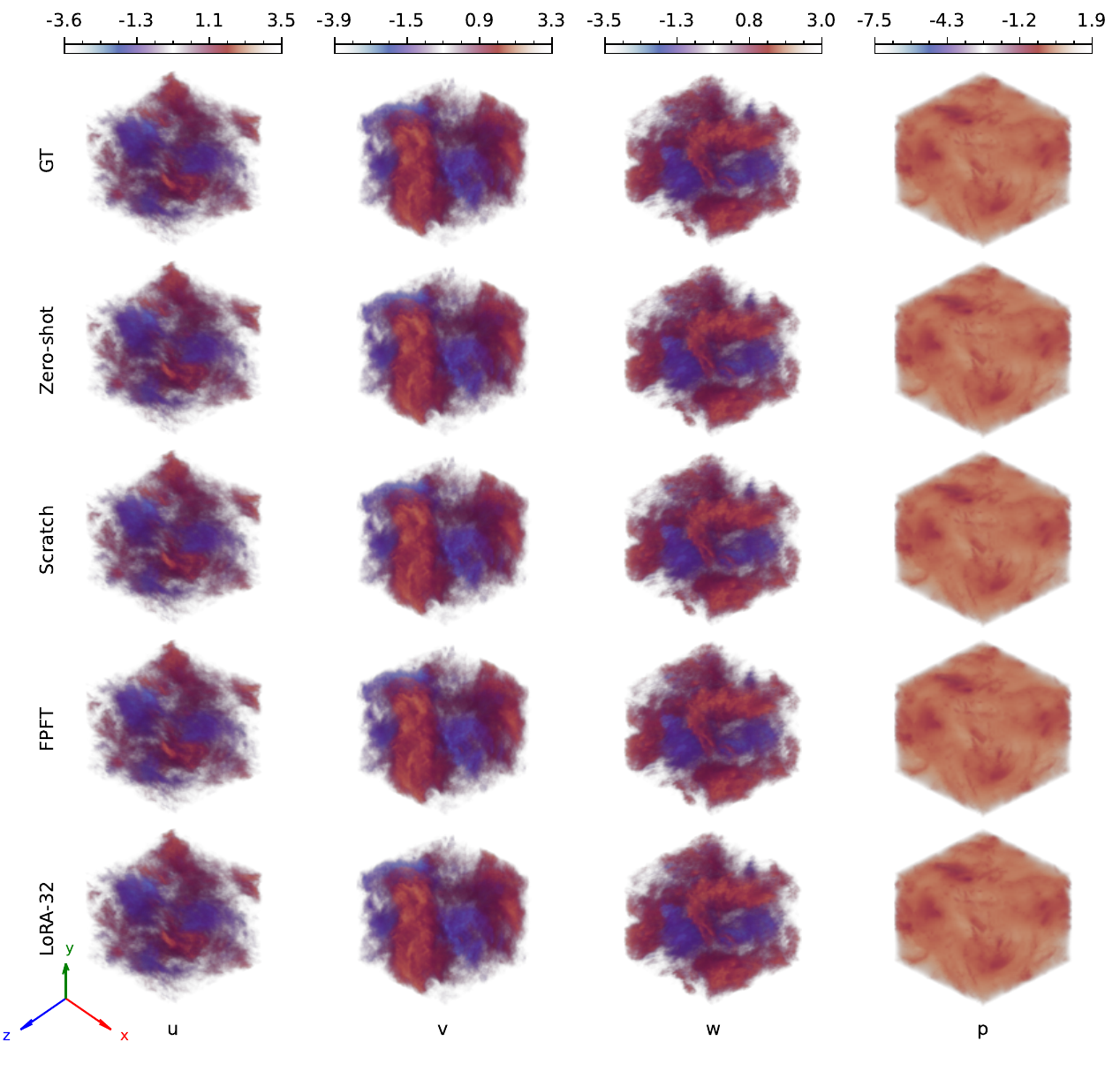}
    \caption{
        Volume rendering of the reconstructed $1024^3$ \iso{} fields generated by different Tadpole training methods.
    }
    \label{fig:vis_recon_iso3d}
\end{figure}

\begin{figure}[htbp]
    \centering
    \includegraphics[scale=0.56]{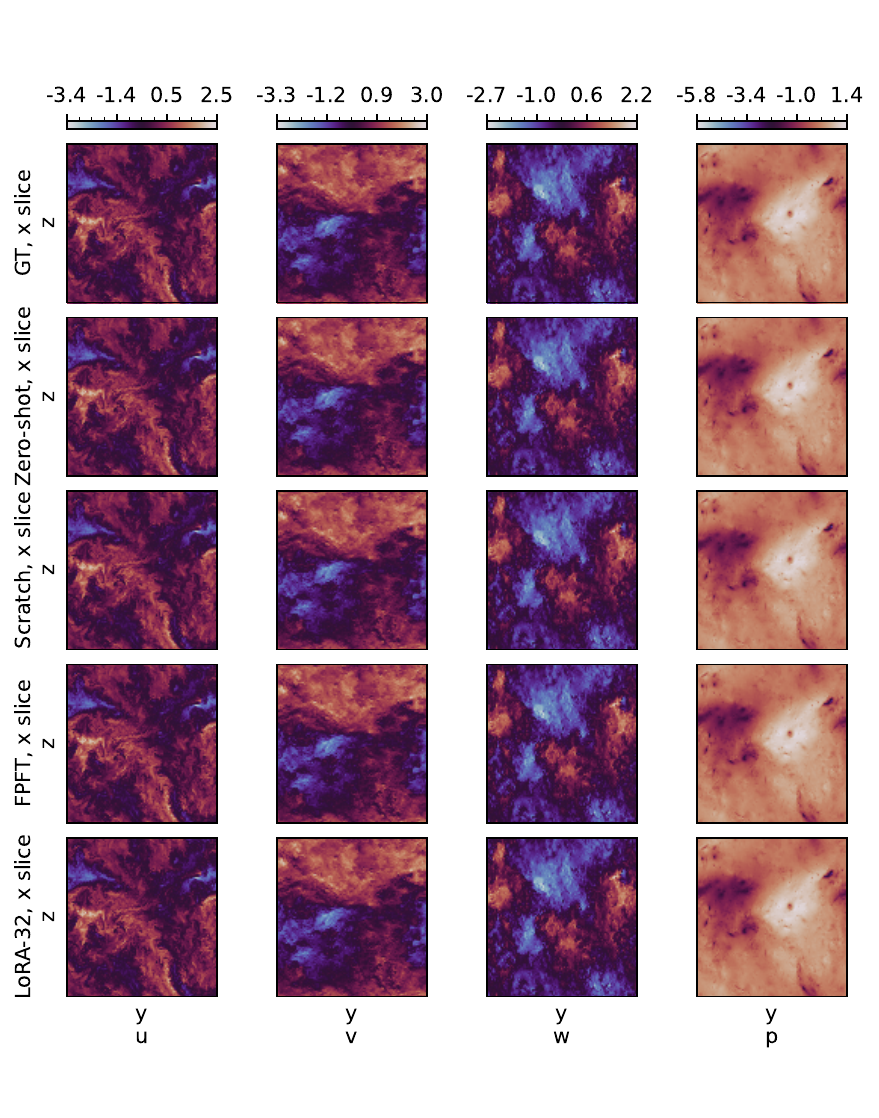}
    \caption{
    Visualization of the reconstruction of \iso{} at the slice where $x=X/2$.
    }
    \label{fig:vis_recon_isoxs}
\end{figure}

\begin{figure}[htbp]
    \centering
    \includegraphics[scale=0.56]{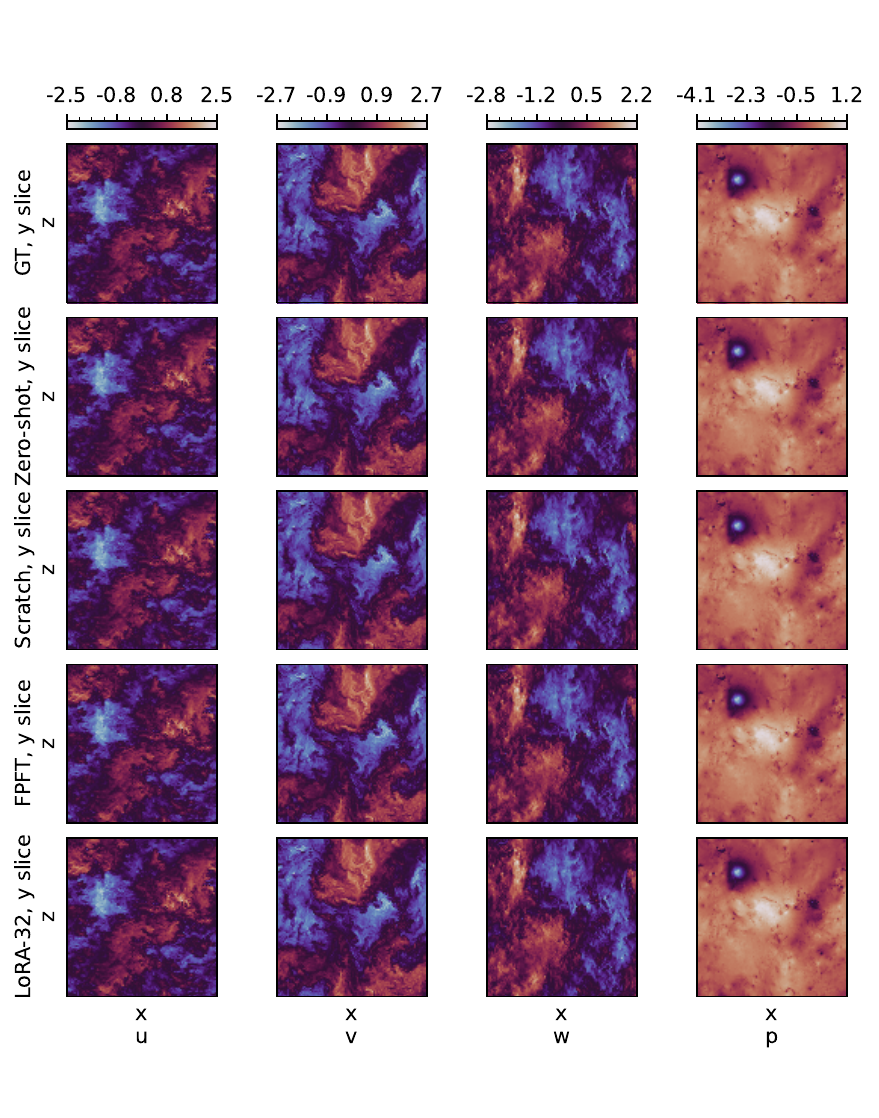}
    \caption{
    Visualization of the reconstruction of \iso{} at the slice where $y=Y/2$.
    }
    \label{fig:vis_recon_isoys}
\end{figure}

\begin{figure}[htbp]
    \centering
    \includegraphics[scale=0.56]{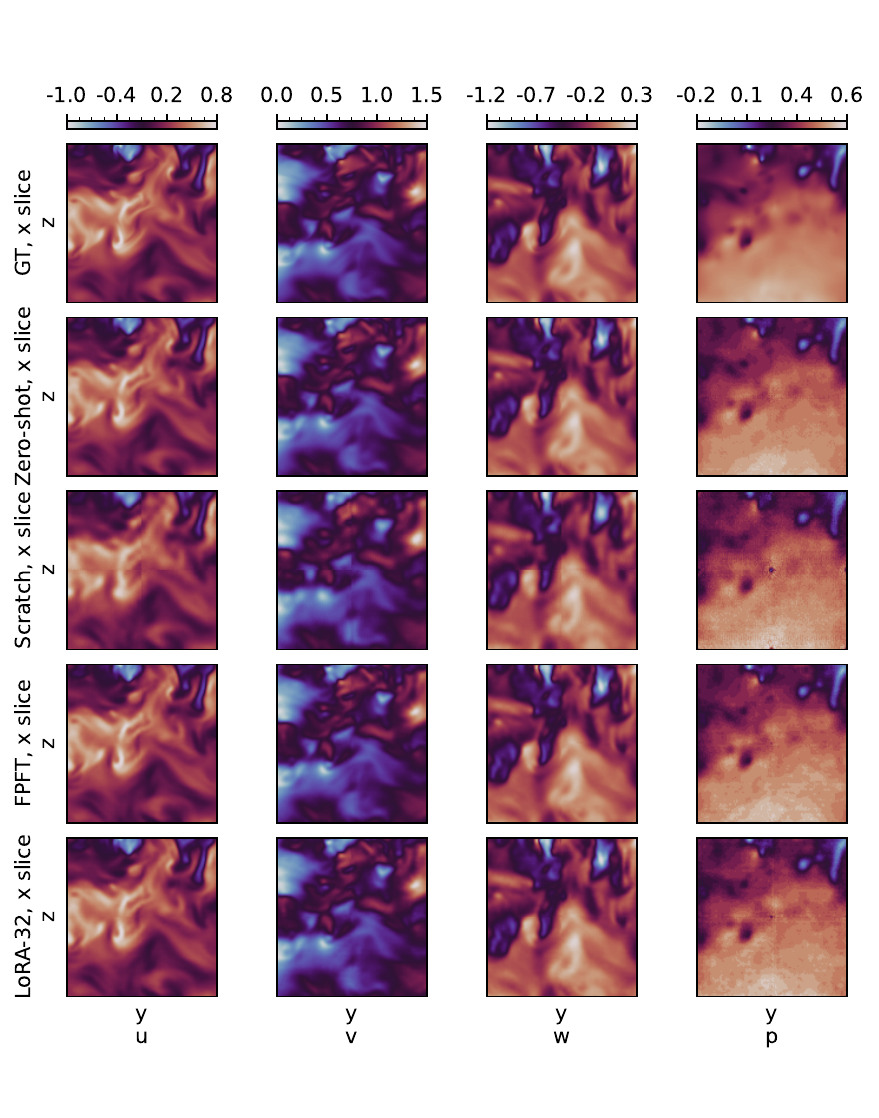}
    \caption{
    Visualization of the reconstruction of \iso{} $128^3$ crops at the slice where $x=X/2$.
    }
    \label{fig:vis_recon_isoxs_128}
\end{figure}

\begin{figure}[htbp]
    \centering
    \includegraphics[scale=0.56]{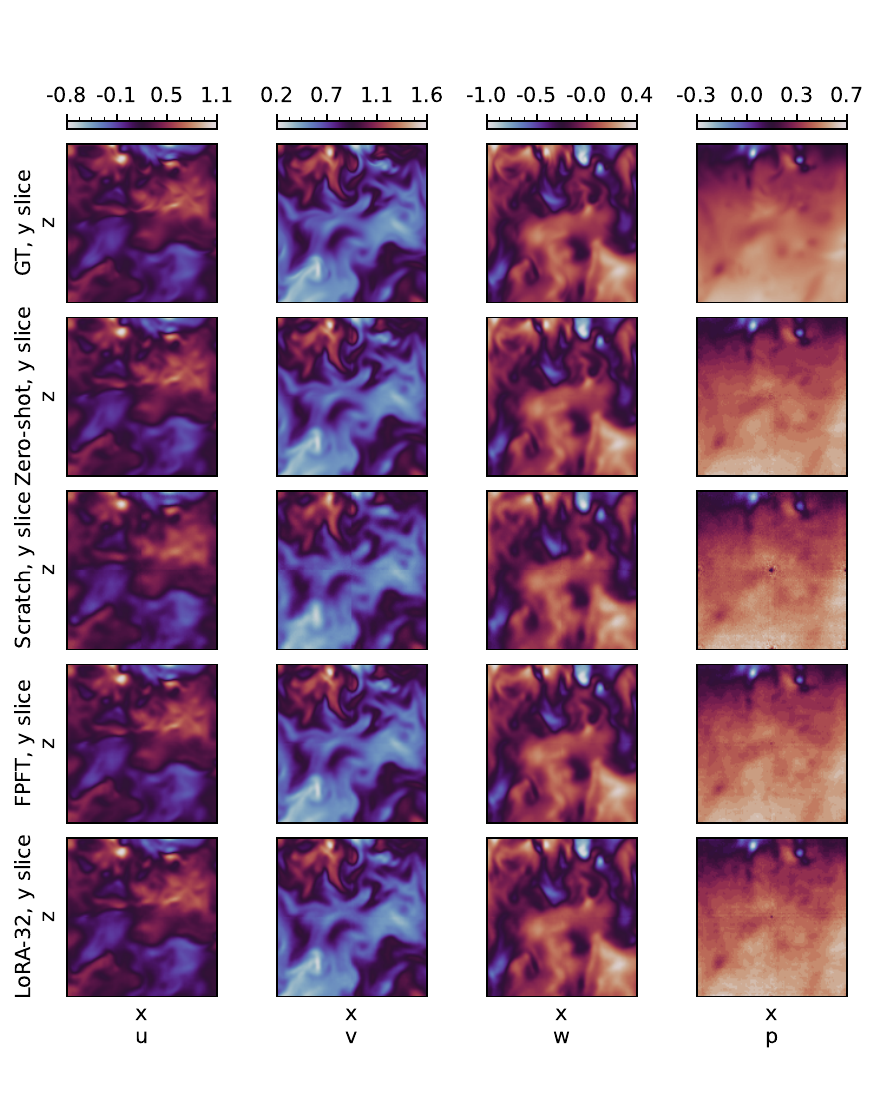}
    \caption{
    Visualization of the reconstruction of \iso{} $128^3$ crops at the slice where $y=Y/2$.
    }
    \label{fig:vis_recon_isoys_128}
\end{figure}

\begin{figure}[htbp]
    \centering
    \includegraphics[scale=0.56]{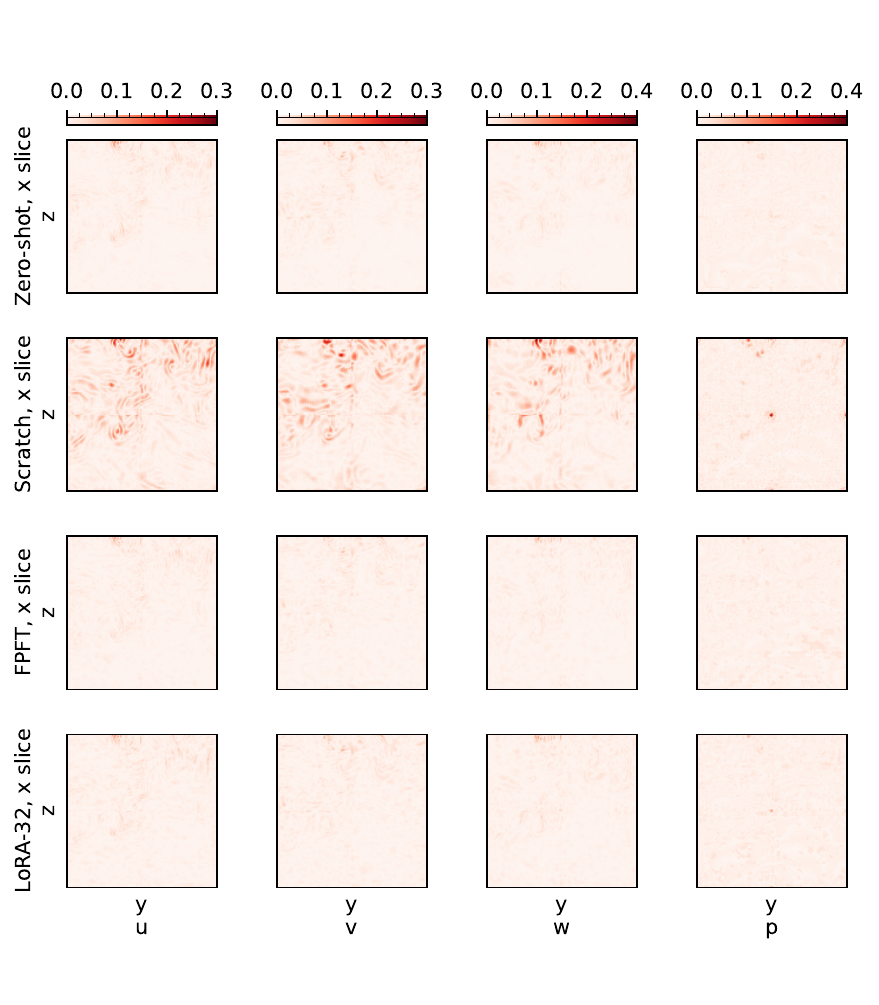}
    \caption{
    Visualization of the absolute error for the reconstruction of \iso{} $128^3$ crops at the slice where $x=X/2$.
    }
    \label{fig:vis_recon_isoxs_128_error}
\end{figure}

\begin{figure}[htbp]
    \centering
    \includegraphics[scale=0.56]{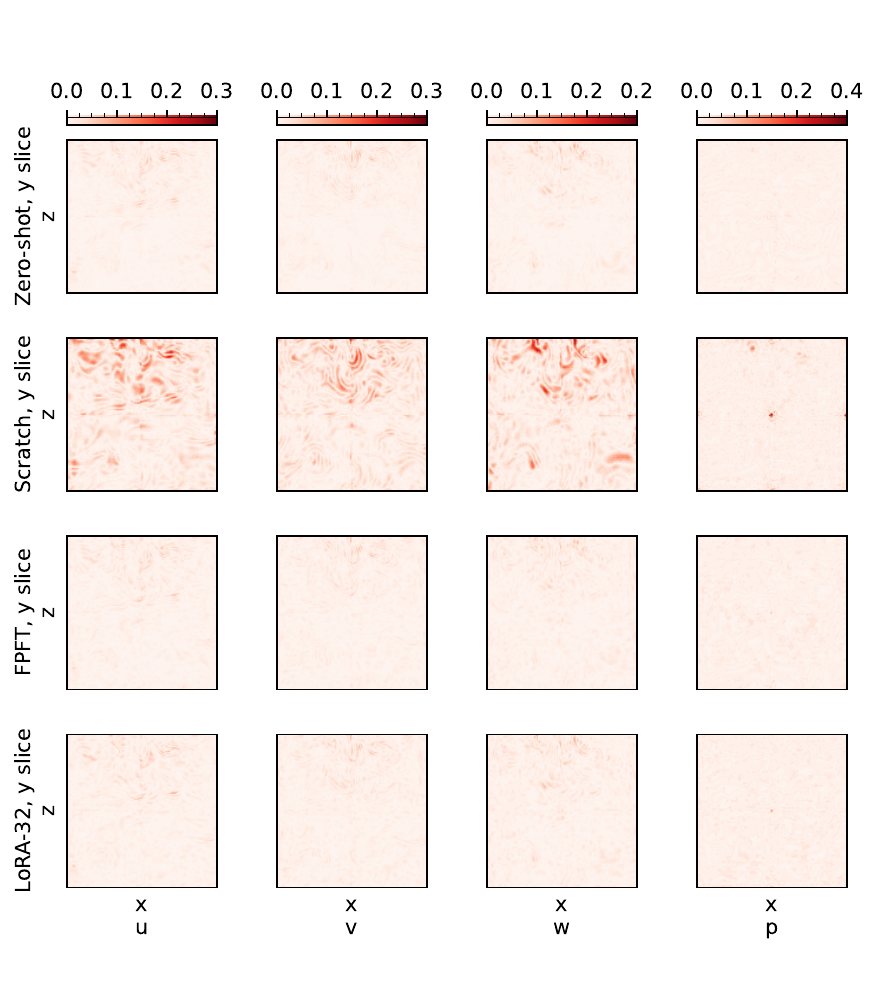}
    \caption{
    Visualization of the absolute error for the reconstruction of \iso{} $128^3$ crops at the slice where $y=Y/2$.
    }
    \label{fig:vis_recon_isoys_128_error}
\end{figure}

\begin{figure}[htbp]
    \centering
    \includegraphics[scale=0.7]{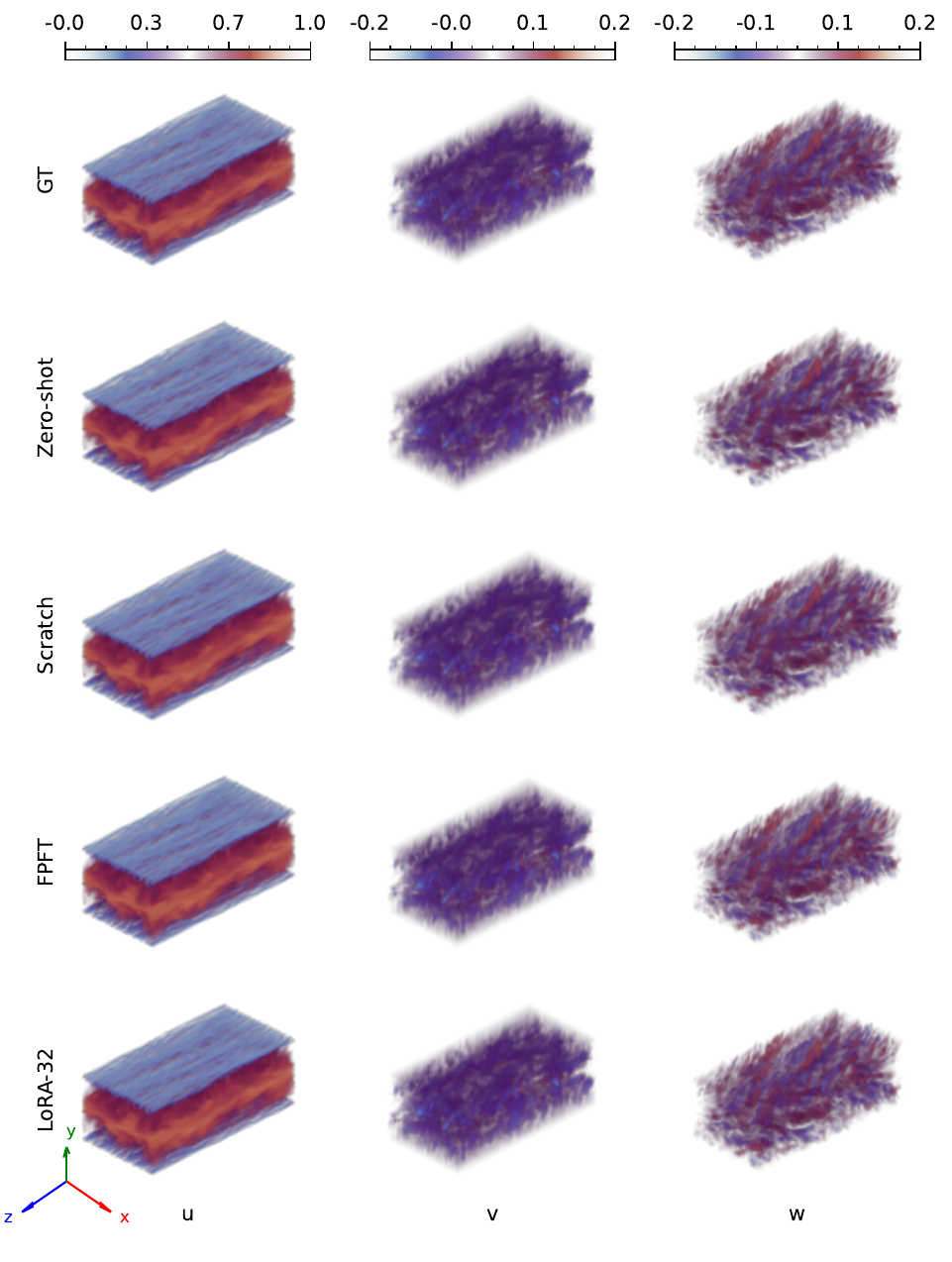}
    \caption{
        Volume rendering of the reconstructed $96^2 \times 192$ \tcf{} fields generated by different Tadpole training methods.
        This visualization confirms that all methods have successfully learned the large-scale structures of the data. 
        Differences become apparent in the following visualizations of individual slices through the volume.
    }
    \label{fig:vis_recon_tcf3d}
\end{figure}

\begin{figure}[htbp]
    \centering
    \includegraphics[scale=0.5]{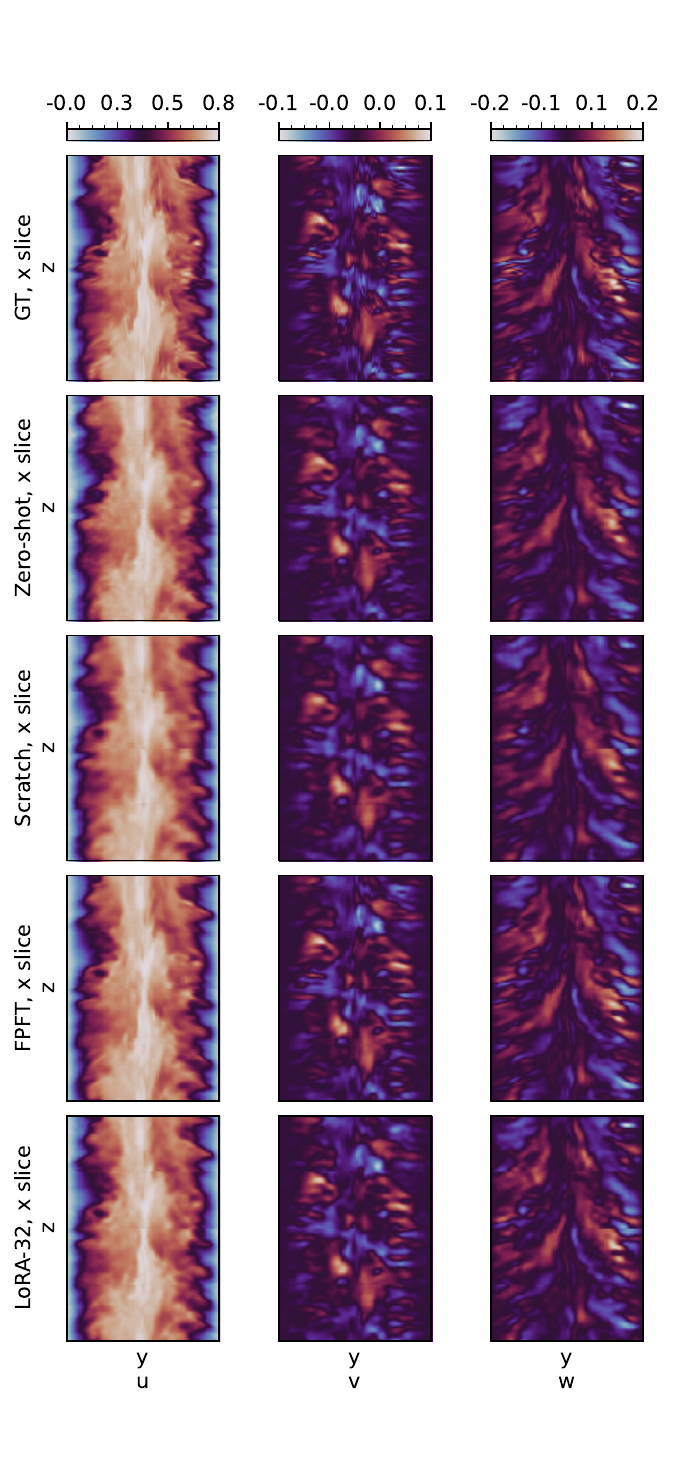}\hspace{10pt}
    \includegraphics[scale=0.5]{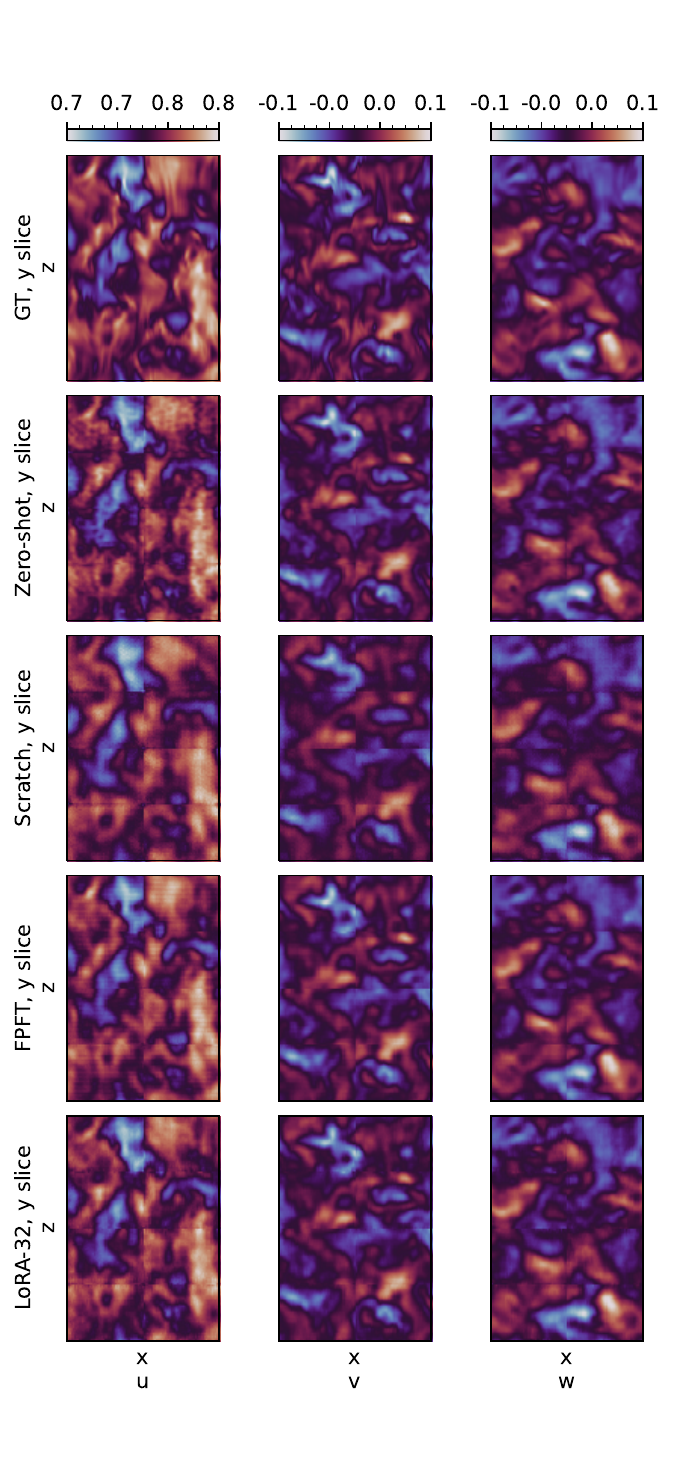}
  \caption{
    Visualization of the reconstruction of \tcf{} at the slice where $x=X/2$ and $y=Y/2$.
    }
    \label{fig:vis_recon_tcfxs}
\end{figure}

\begin{figure}[htbp]
    \centering
    \includegraphics[scale=0.5]{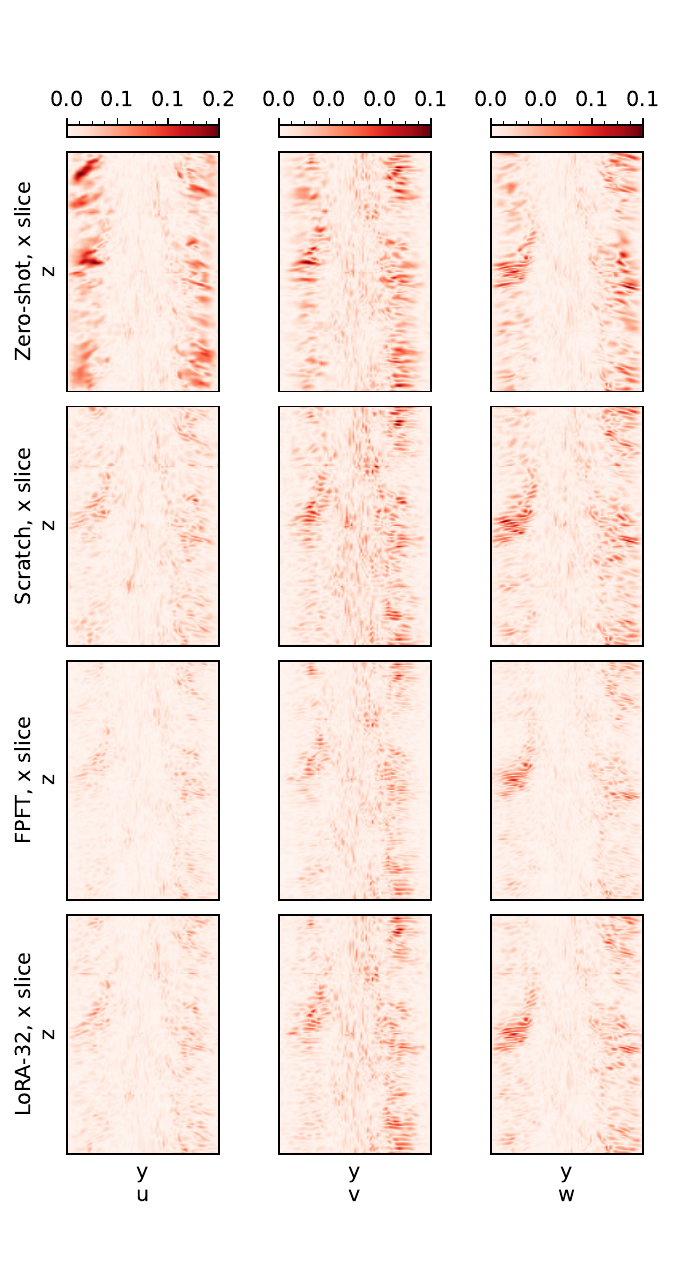}\hspace{10pt}
    \includegraphics[scale=0.5]{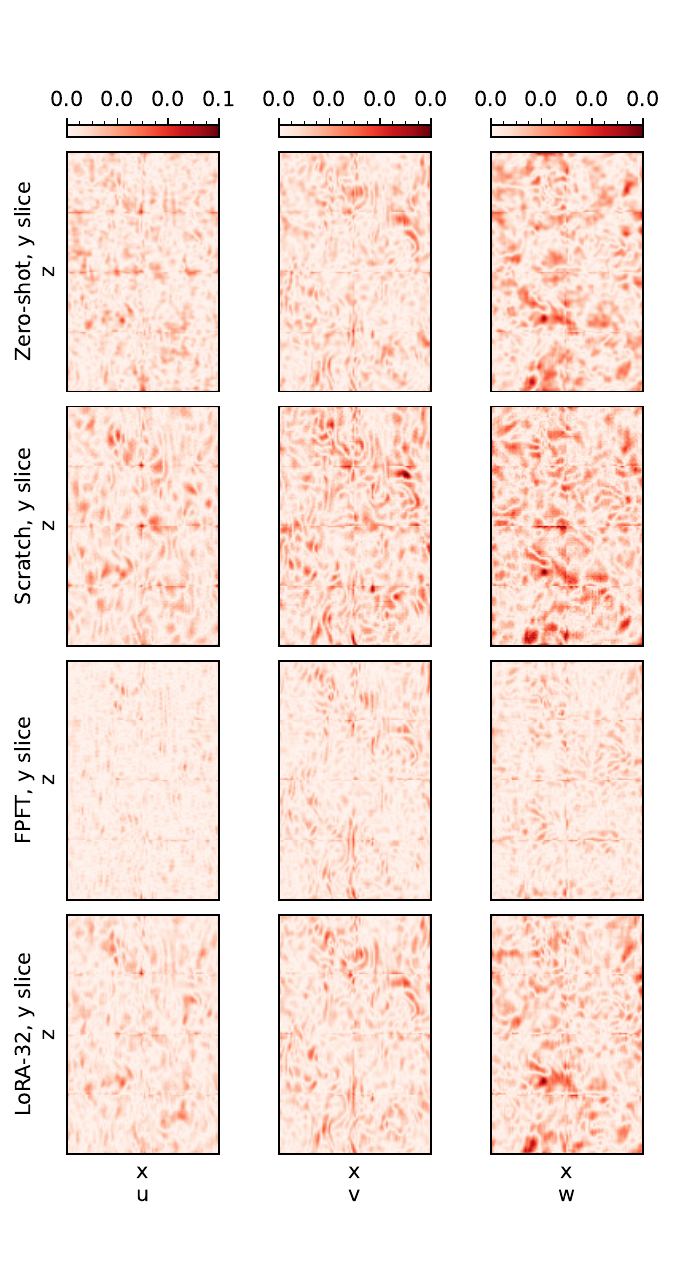}
  \caption{
    Visualization of the absolute error for the reconstruction of \tcf{} at the slice where $x=X/2$ and $y=Y/2$.
    }
    \label{fig:vis_recon_tcfxs_error}
\end{figure}

\begin{landscape}
\begin{figure}[htbp]
    \centering
    \includegraphics[scale=0.6]{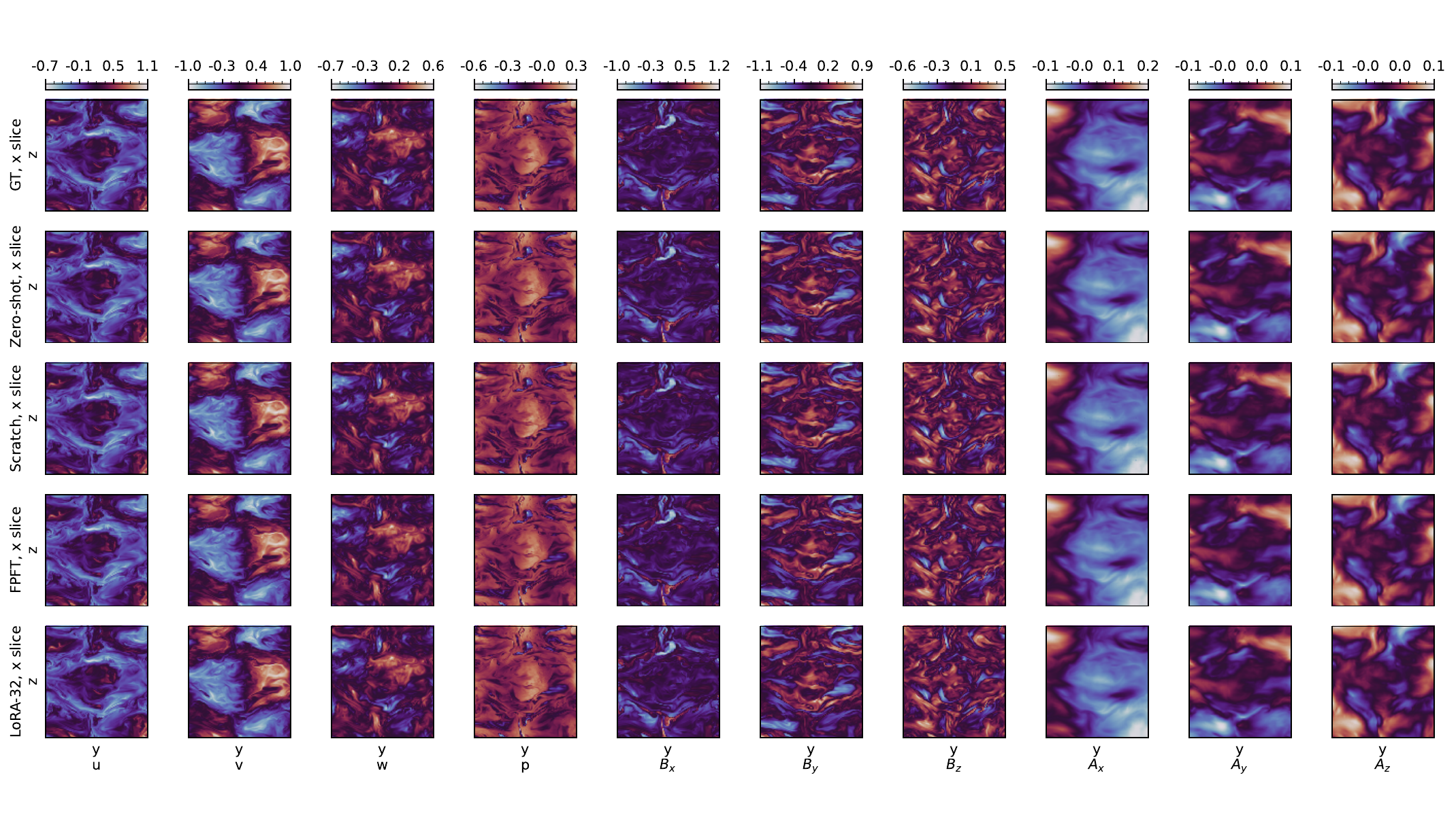}
    \caption{
    Visualization of the reconstruction of \mhd{} at the slice where $x=X/2$.
    }
    \label{fig:vis_recon_mhdxs}
\end{figure}
\end{landscape}

\begin{landscape}
\begin{figure}[htbp]
    \centering
    \includegraphics[scale=0.6]{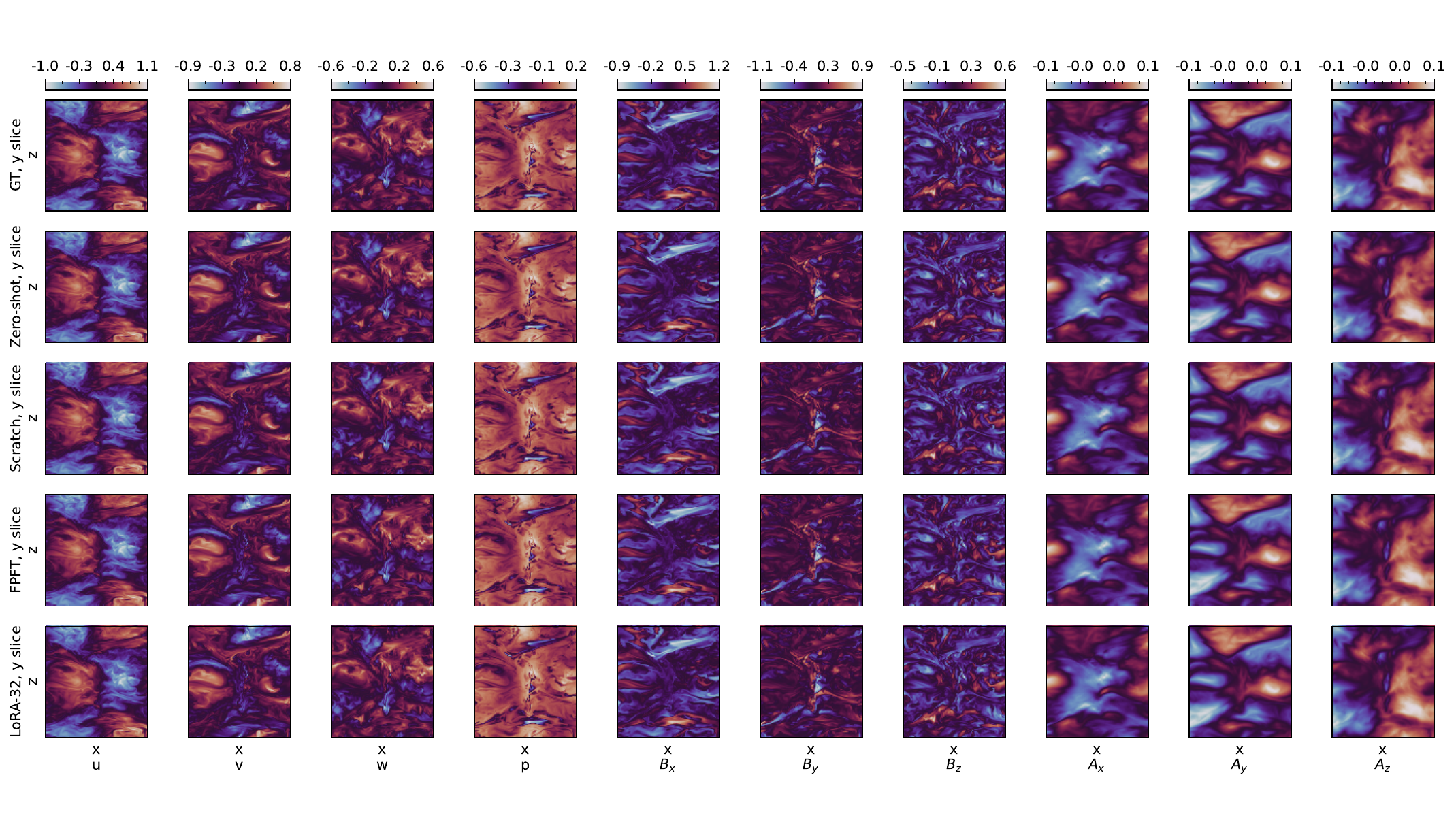}
    \caption{
    Visualization of the reconstruction of \mhd{} at the slice where $y=Y/2$.
    }
    \label{fig:vis_recon_mhdys}
\end{figure}
\end{landscape}

\begin{landscape}
\begin{figure}[htbp]
    \centering
    \includegraphics[scale=0.6]{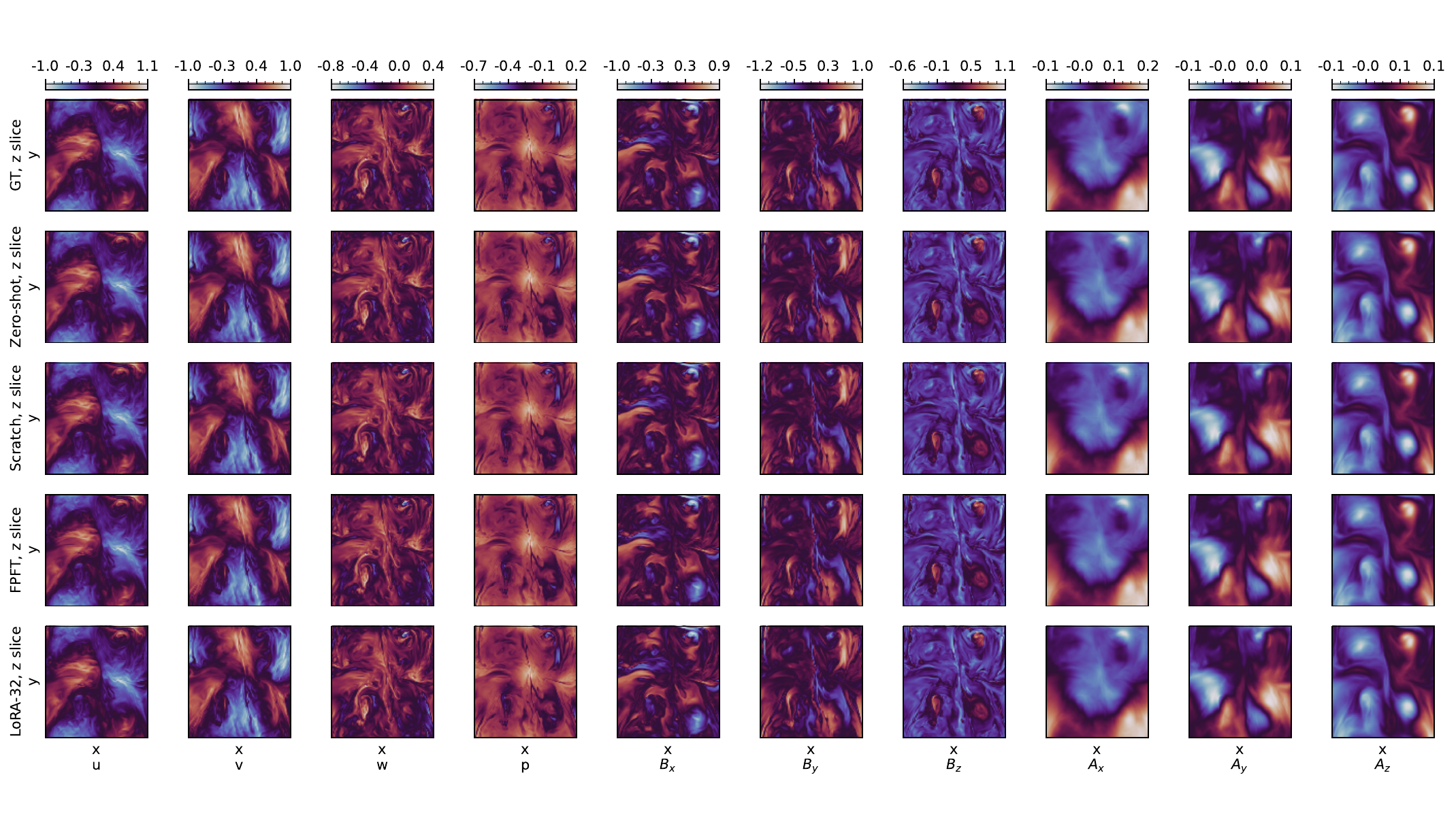}
    \caption{
    Visualization of the reconstruction of \mhd{} at the slice where $z=Z/2$.
    }
    \label{fig:vis_recon_mhdzs}
\end{figure}
\end{landscape}

\begin{landscape}
\begin{figure}[htbp]
    \centering
    \includegraphics[scale=0.6]{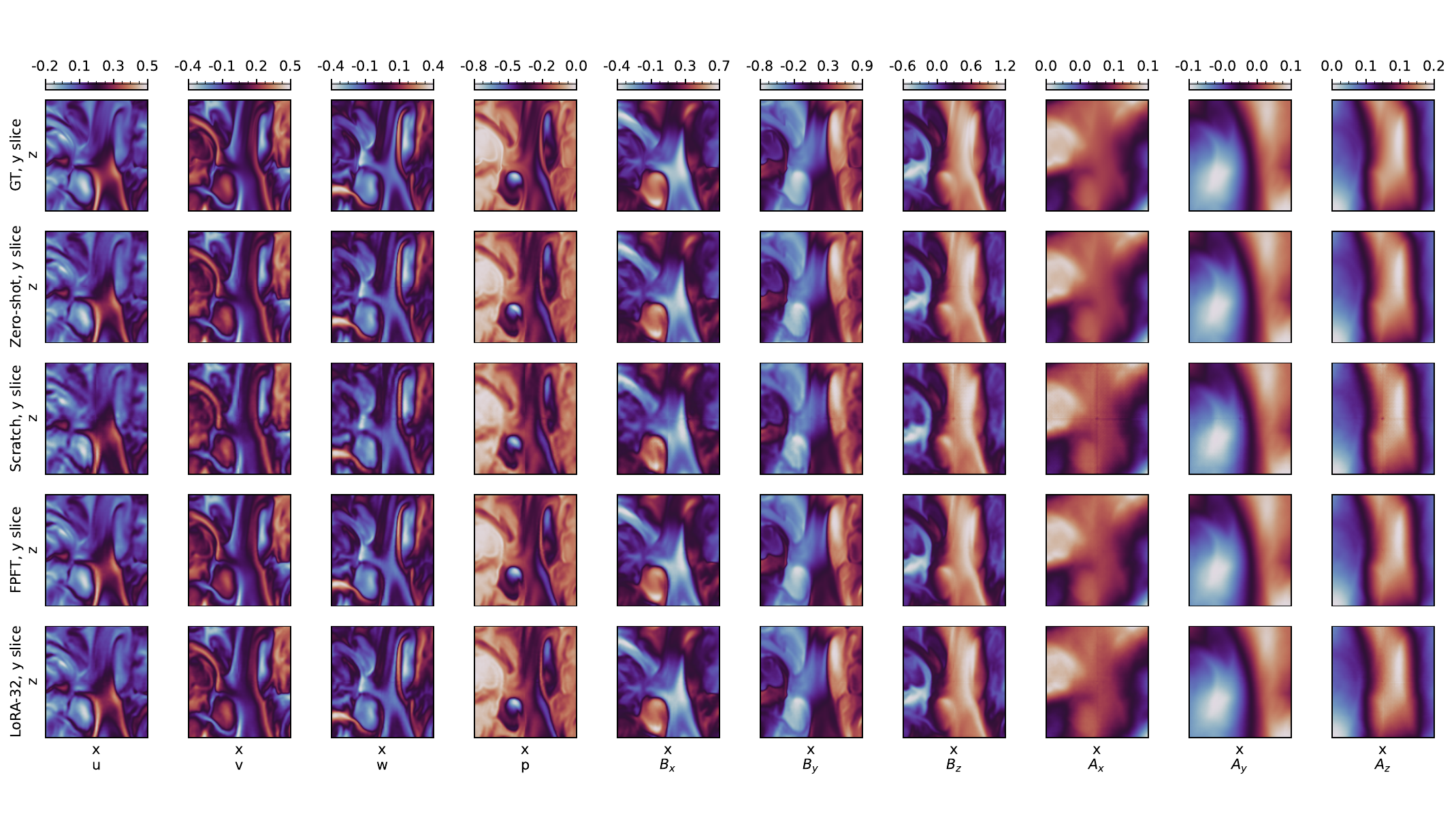}
    \caption{
    Visualization of the reconstruction of \mhd{} $128^3$ crops at the slice where $y=Y/2$.
    }
    \label{fig:vis_recon_mhdys_128}
\end{figure}
\end{landscape}

\begin{landscape}
\begin{figure}[htbp]
    \centering
    \includegraphics[scale=0.6]{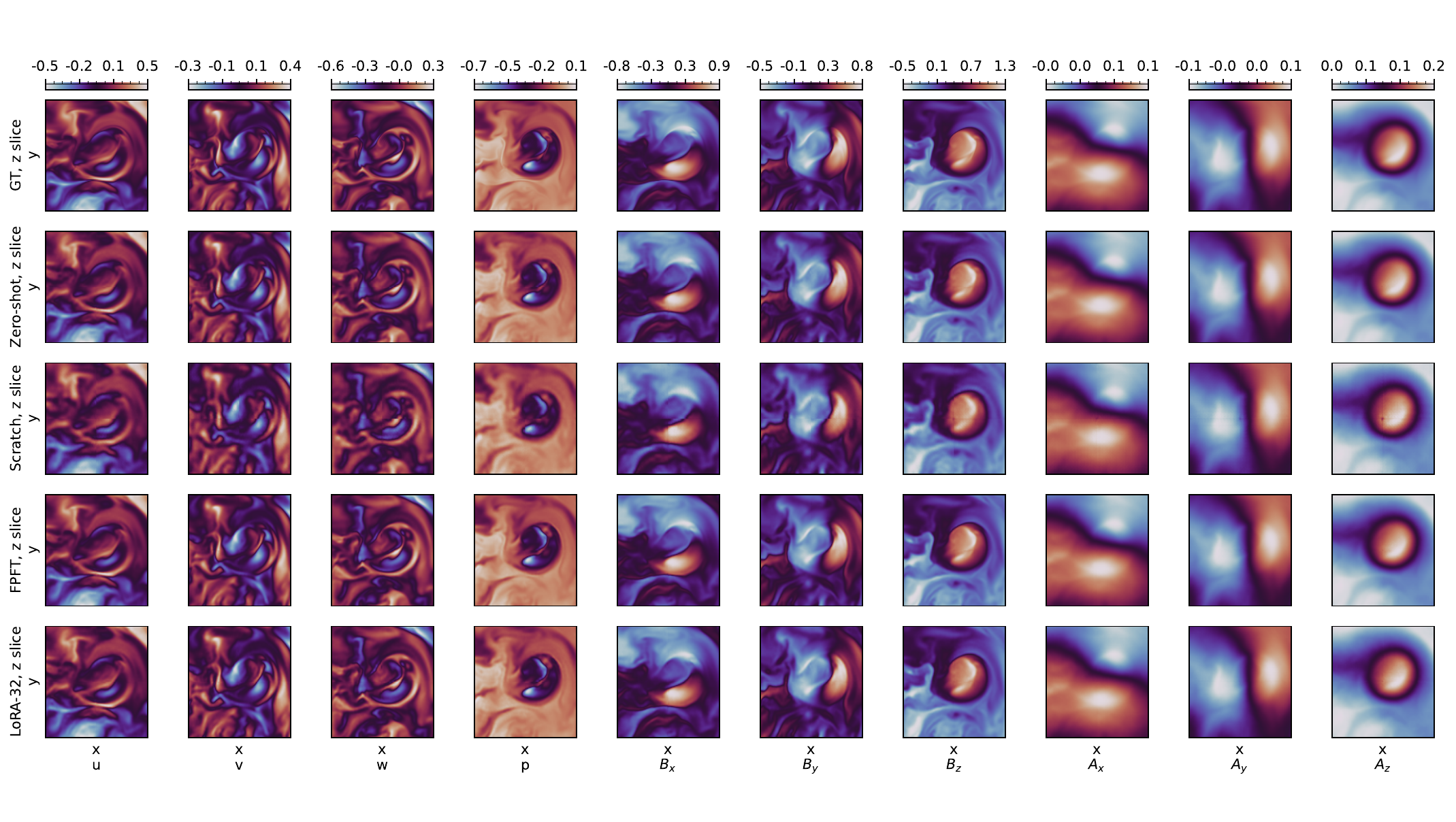}
    \caption{
    Visualization of the reconstruction of \mhd{} $128^3$ crops at the slice where $z=Z/2$.
    }
    \label{fig:vis_recon_mhdzs_128}
\end{figure}
\end{landscape}

\begin{figure}[htbp]
    \centering
    \includegraphics[scale=0.65]{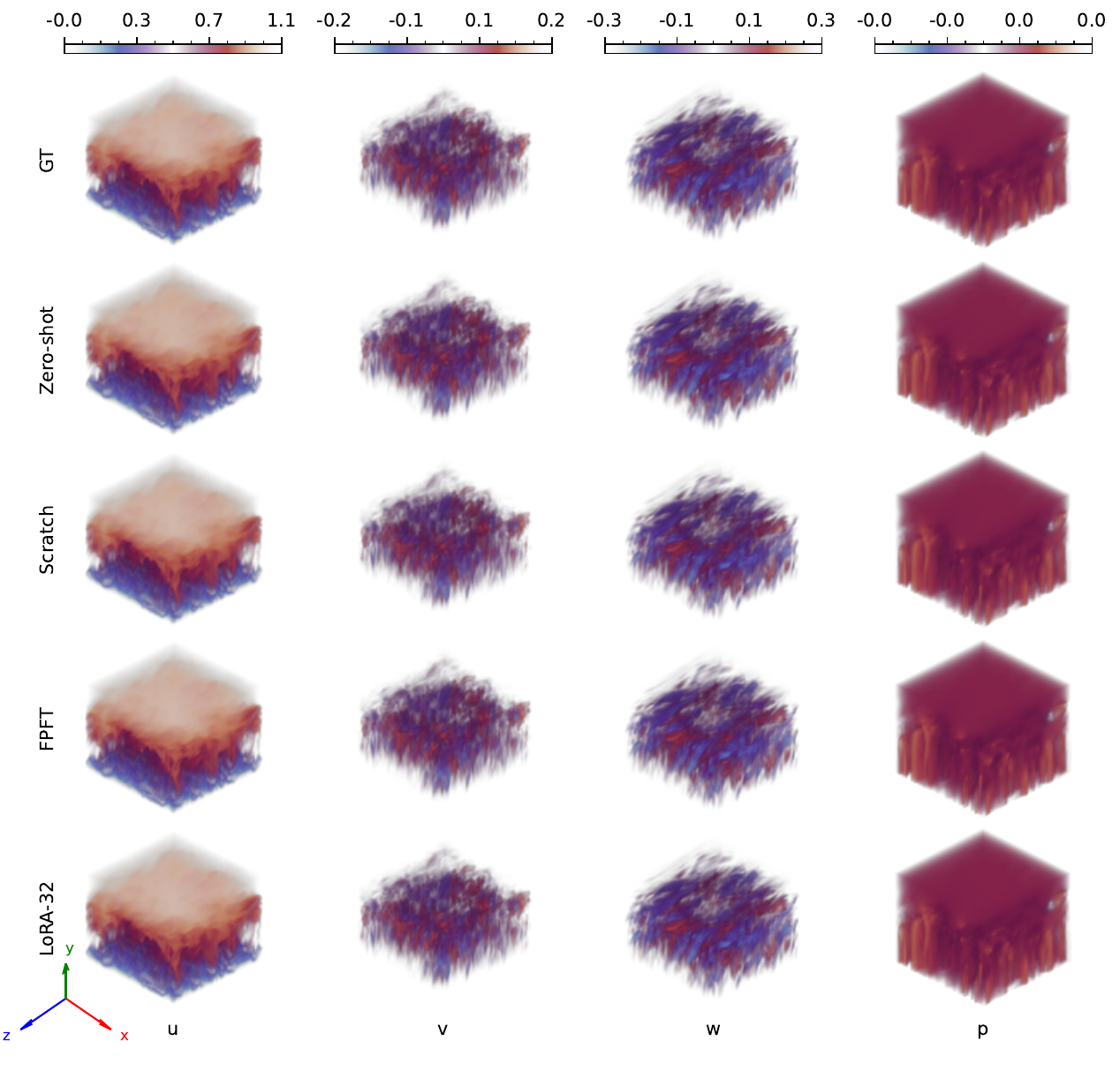}
    \caption{
        Volume rendering of the reconstructed $224^3$ \tbl{} fields generated by different Tadpole training methods.
    }
    \label{fig:vis_recon_bl3d}
\end{figure}

\begin{figure}[htbp]
    \centering
    \includegraphics[scale=0.56]{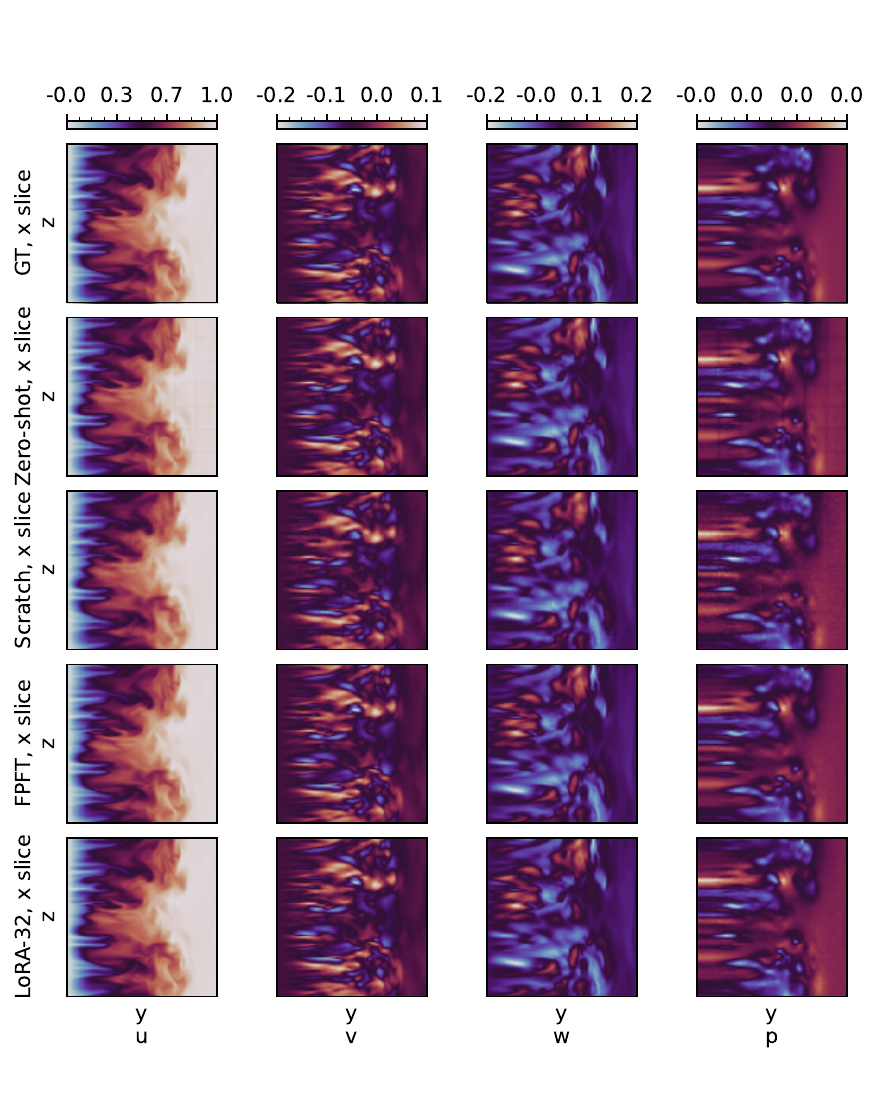}
    \caption{
    Visualization of the reconstruction of \tbl{} at the slice where $x=X/2$.
    }
    \label{fig:vis_recon_blxs}
\end{figure}

\begin{figure}[htbp]
    \centering
    \includegraphics[scale=0.56]{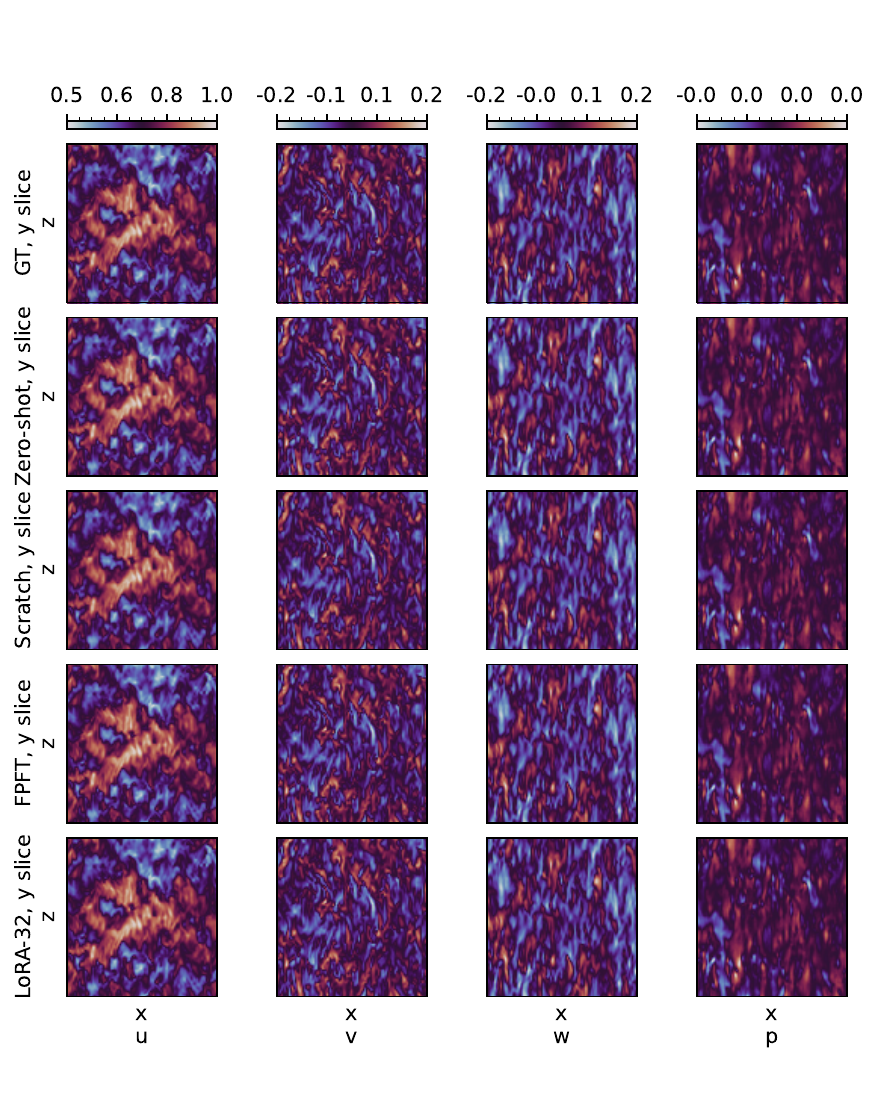}
    \caption{
    Visualization of the reconstruction of \tbl{} at the slice where $y=Y/2$.
    }
    \label{fig:vis_recon_blys}
\end{figure}

\begin{figure}[htbp]
    \centering
    \includegraphics[scale=0.56]{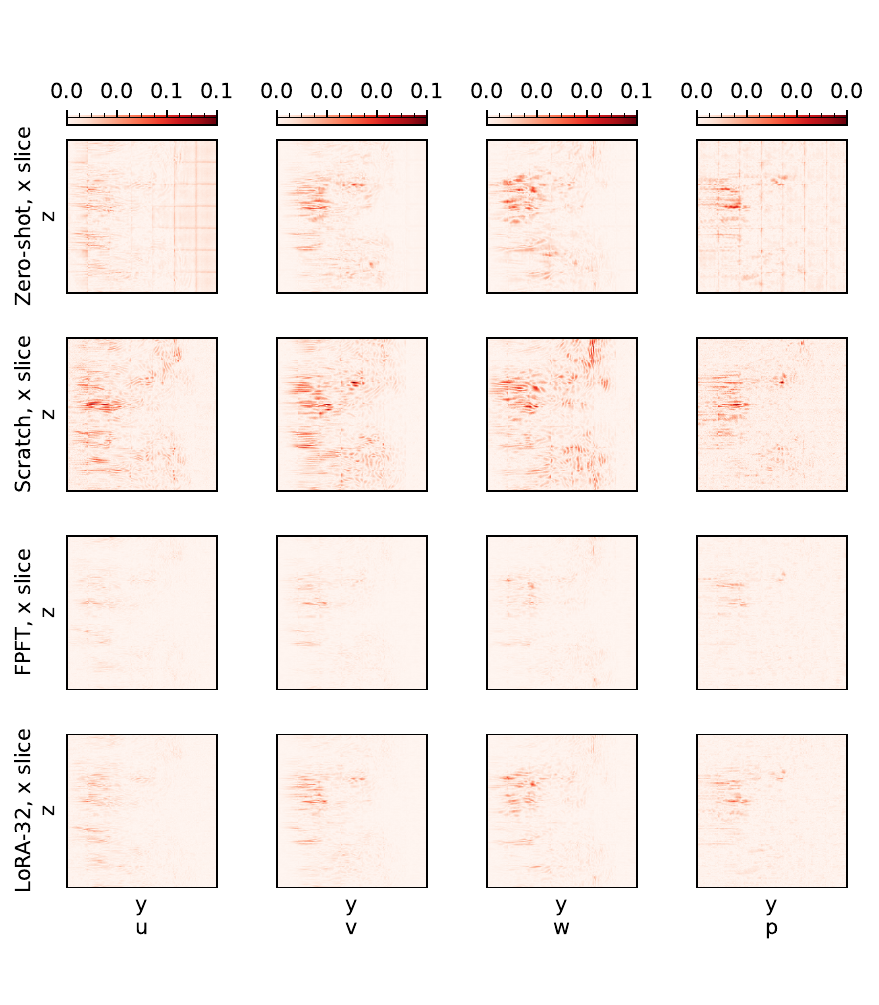}
    \caption{
    Visualization of absolute error for the reconstruction of \tbl{} at the slice where $x=X/2$.
    }
    \label{fig:vis_recon_blxs_error}
\end{figure}

\begin{figure}[htbp]
    \centering
    \includegraphics[scale=0.56]{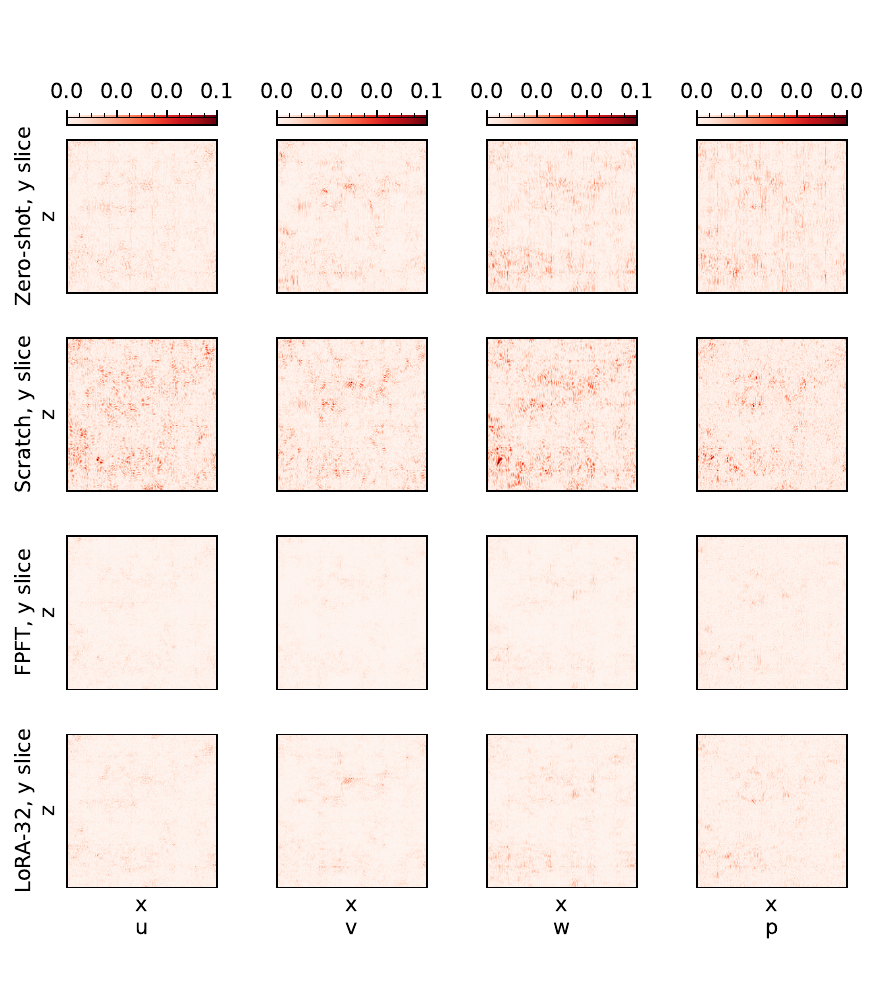}
    \caption{
    Visualization of absolute error for the reconstruction of \tbl{} at the slice where $y=Y/2$.
    }
    \label{fig:vis_recon_blys_error}
\end{figure}

Figures~\ref{fig:vis_recon_iso3d}–\ref{fig:vis_recon_blys_error} present qualitative reconstruction results for different datasets under various training strategies. Due to the high spatial resolution of \iso{} and \mhd, differences in the reconstructions are difficult to discern at lower visualization resolutions. We therefore additionally provide visualizations of $128^3$ crops for clearer comparison. Overall, models trained from scratch exhibit higher reconstruction errors than the other variants. The zero-shot model occasionally shows inconsistencies at crop boundaries; however, these artifacts disappear after fine-tuning on the corresponding dataset.

\FloatBarrier{}

\FloatBarrier{}

\begin{landscape}
\vspace*{\fill}
\begin{figure}[htbp]
    \centering
    \includegraphics[scale=0.4]{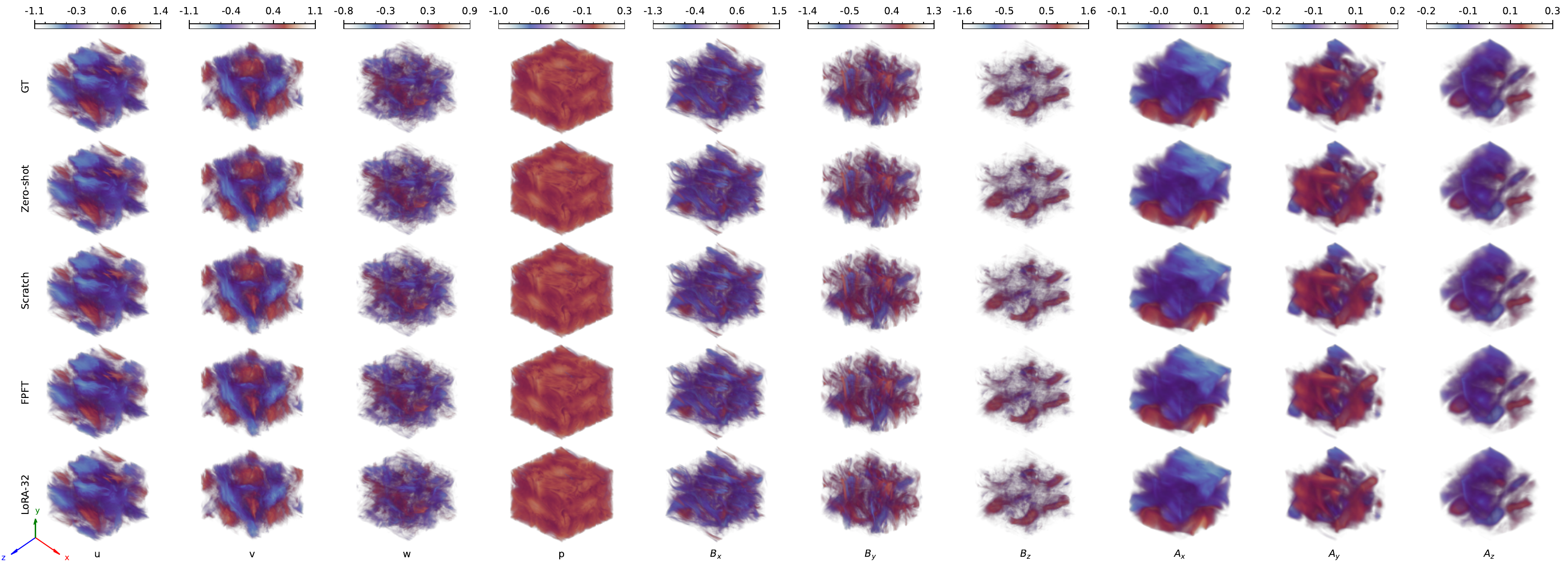}
    \caption{
        Volume rendering of the reconstructed $512^3$ \mhd{} fields generated by different Tadpole training methods.
    }
    \label{fig:vis_recon_mhd3d}
\end{figure}
\vspace*{\fill}
\end{landscape}
\begin{landscape}
\vspace*{\fill}
\begin{figure}[htbp]
    \centering
    \includegraphics[scale=0.6]{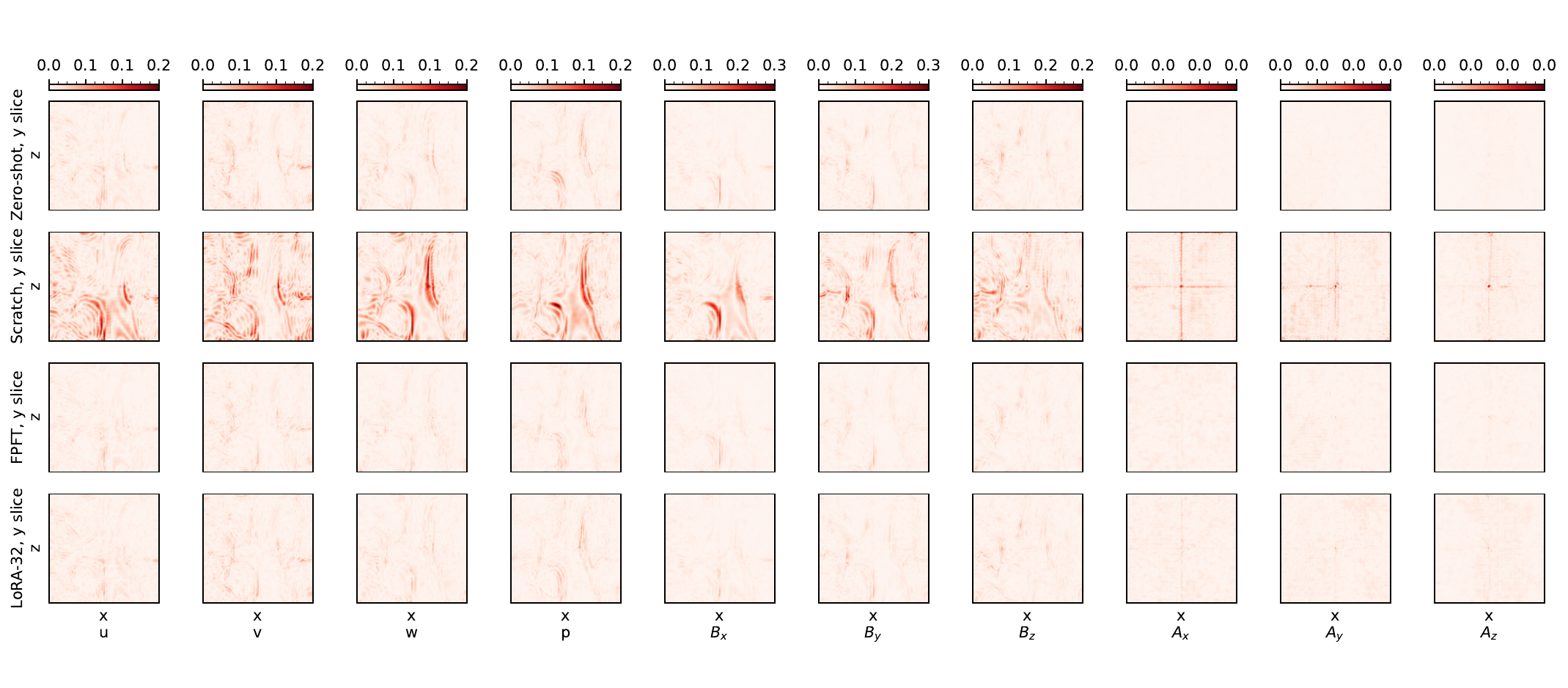}
    \caption{
    Visualization of the absolute error for the reconstruction of \mhd{} $128^3$ crops at the slice where $y=Y/2$.
    }
    \label{fig:vis_recon_mhdys_128_error}
\end{figure}
\vspace*{\fill}
\end{landscape}
\begin{landscape}
\vspace*{\fill}
\begin{figure}[htbp]
    \centering
    \includegraphics[scale=0.6]{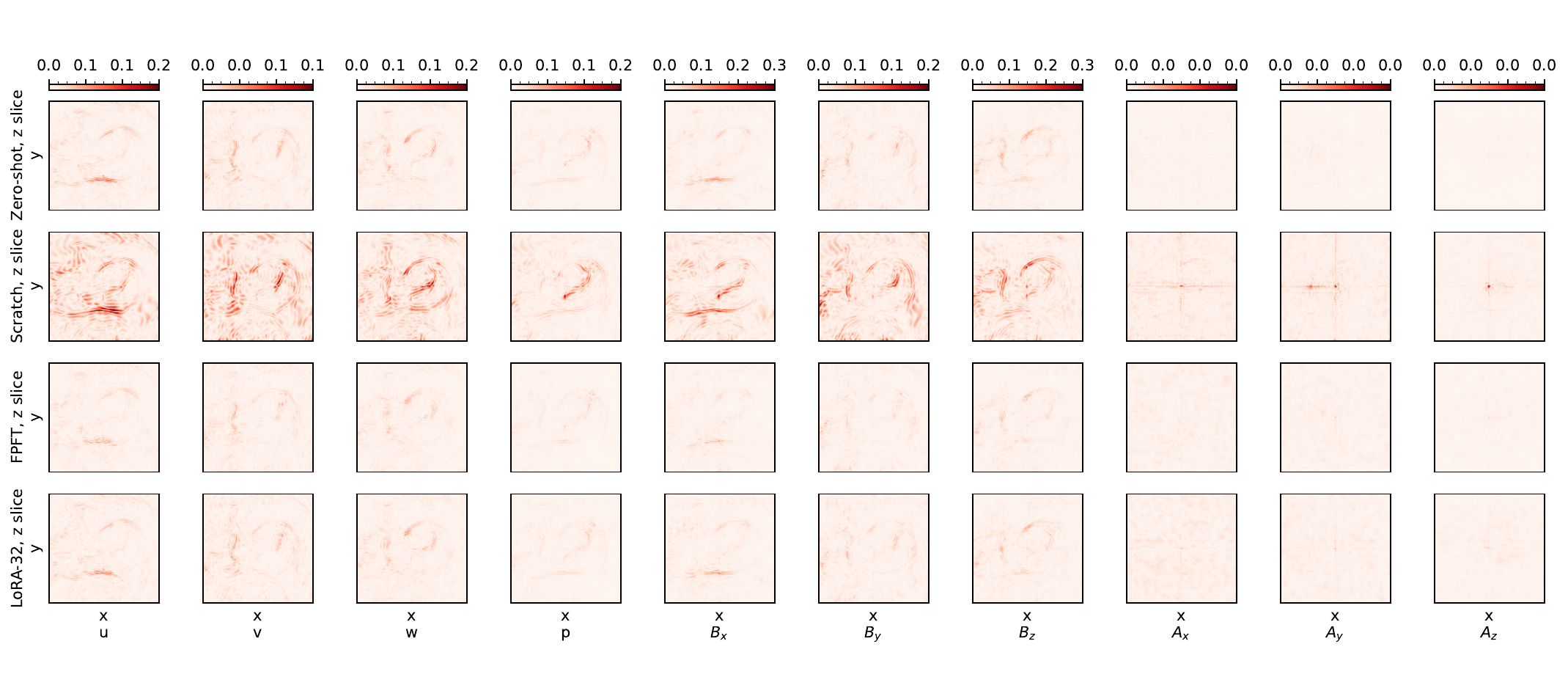}
    \caption{
    Visualization of the absolute error for the reconstruction of \mhd{} $128^3$ crops at the slice where $z=Z/2$.
    }
    \label{fig:vis_recon_mhdzs_128_error}
\end{figure}
\vspace*{\fill}
\end{landscape}

\FloatBarrier{}

\FloatBarrier
\subsection{Visualizations on dynamics learning \label{sec:additional_results_visualizations:dynamics}}

\newcommand{\figspace}{\vspace{-15pt}}

The following images provide qualitative samples for the dynamics rollout task over the course of three time steps. Each image compares the ground truth result (at the top) with the three baseline models (next three rows, top to bottom: Walrus, DPOT, and MORPH), with the Tadpole-DFT result shown in the bottom row. 

As indicated by the MSE measurements in the main text, both DPOT and MORPH exhibit a worse performance. The outputs are relatively smooth, and both methods show clear patch boundaries induced by the underlying architectures. The MORPH model induces a substantial amount of smoothing in the first step, while DPOT loses detail more gradually. In contrast, both Walrus and Tadpole produce outputs that are hard to distinguish from the ground truth. Importantly, Tadpole achieves this with a more than $100\times$ smaller number of trainable weights. 

\begin{figure}[htbp]
    \centering
    \includegraphics[scale=0.33]{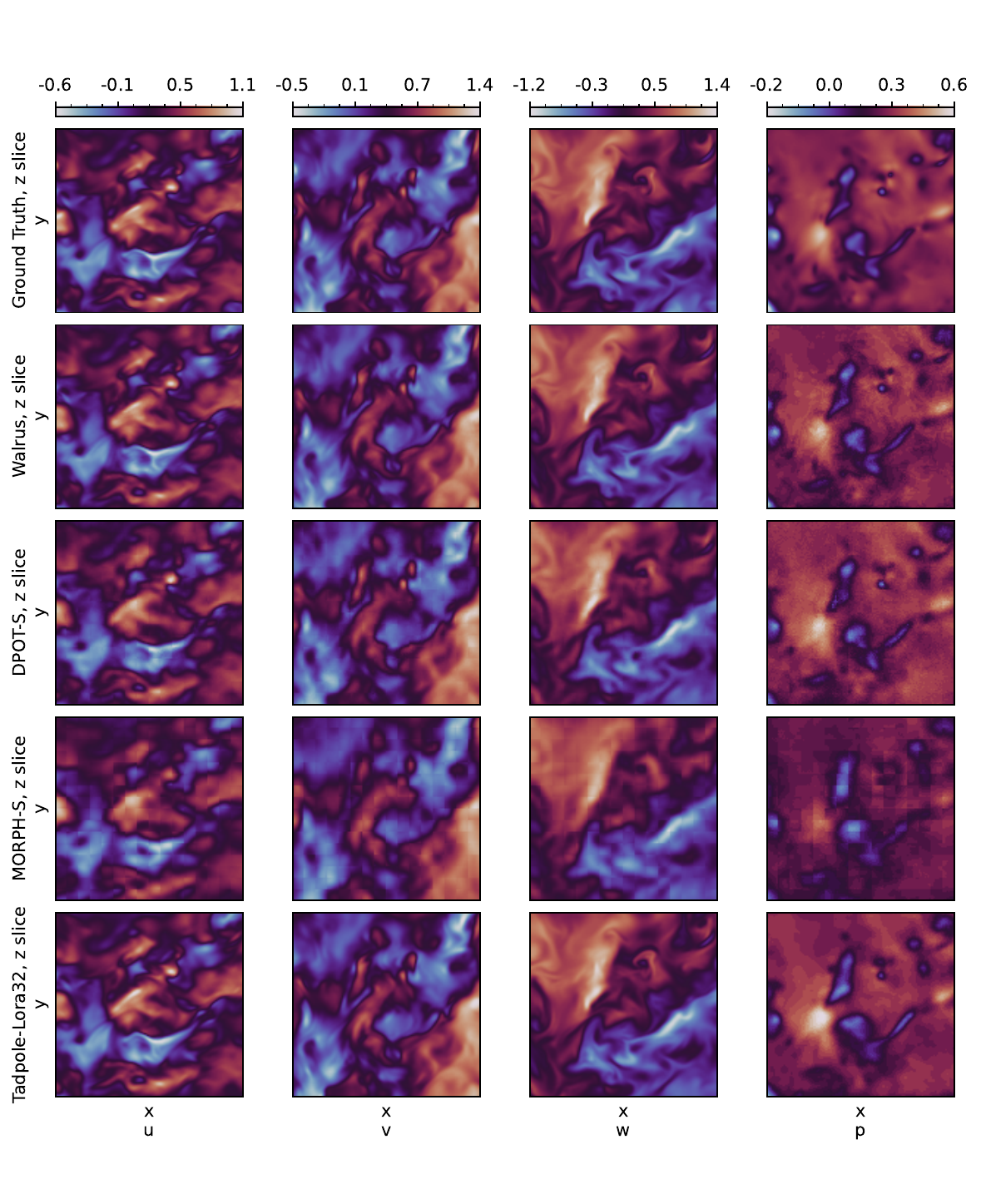}\hspace{10pt}
    \includegraphics[scale=0.33]{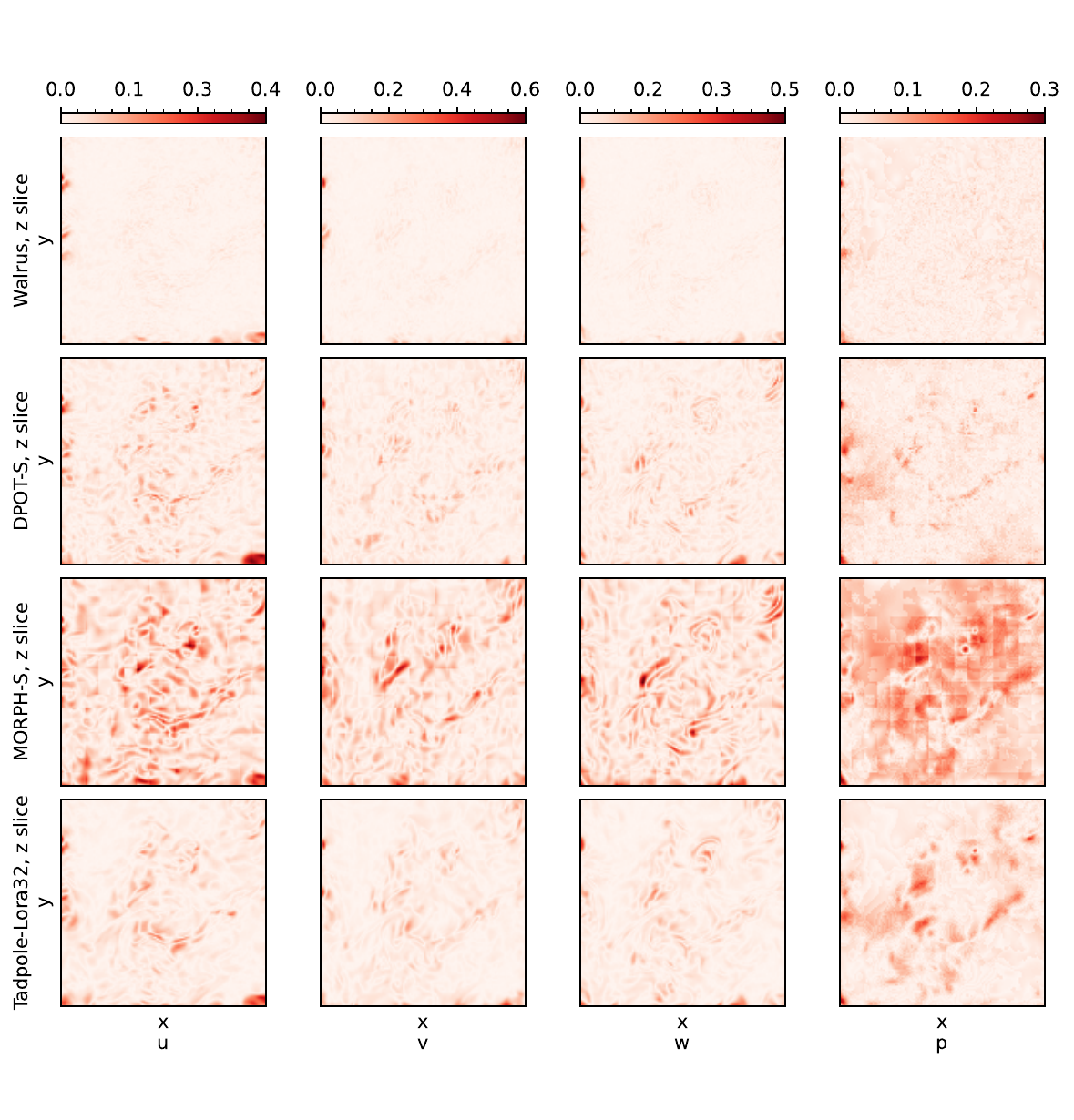}
  \caption{
    Visualization of the prediction (left) of \iso{} and the corresponding absolute error (right) at the \textbf{third rollout step} and the slice where $z=Z/2$.
    }
    \label{fig:vis_dynamics_z_3}
\end{figure}

In addition, the Walrus results feature noticeable noise in the pressure channel (e.g., fourth column in \cref{fig:vis_dynamics_z_3}). This is most likely caused by the Walrus model having difficulties in adjusting to the different scales of velocity and pressure channels. The outputs of the Tadpole model are significantly smoother without losing detail.

\begin{figure}[htbp]
    \centering
    \includegraphics[scale=0.33]{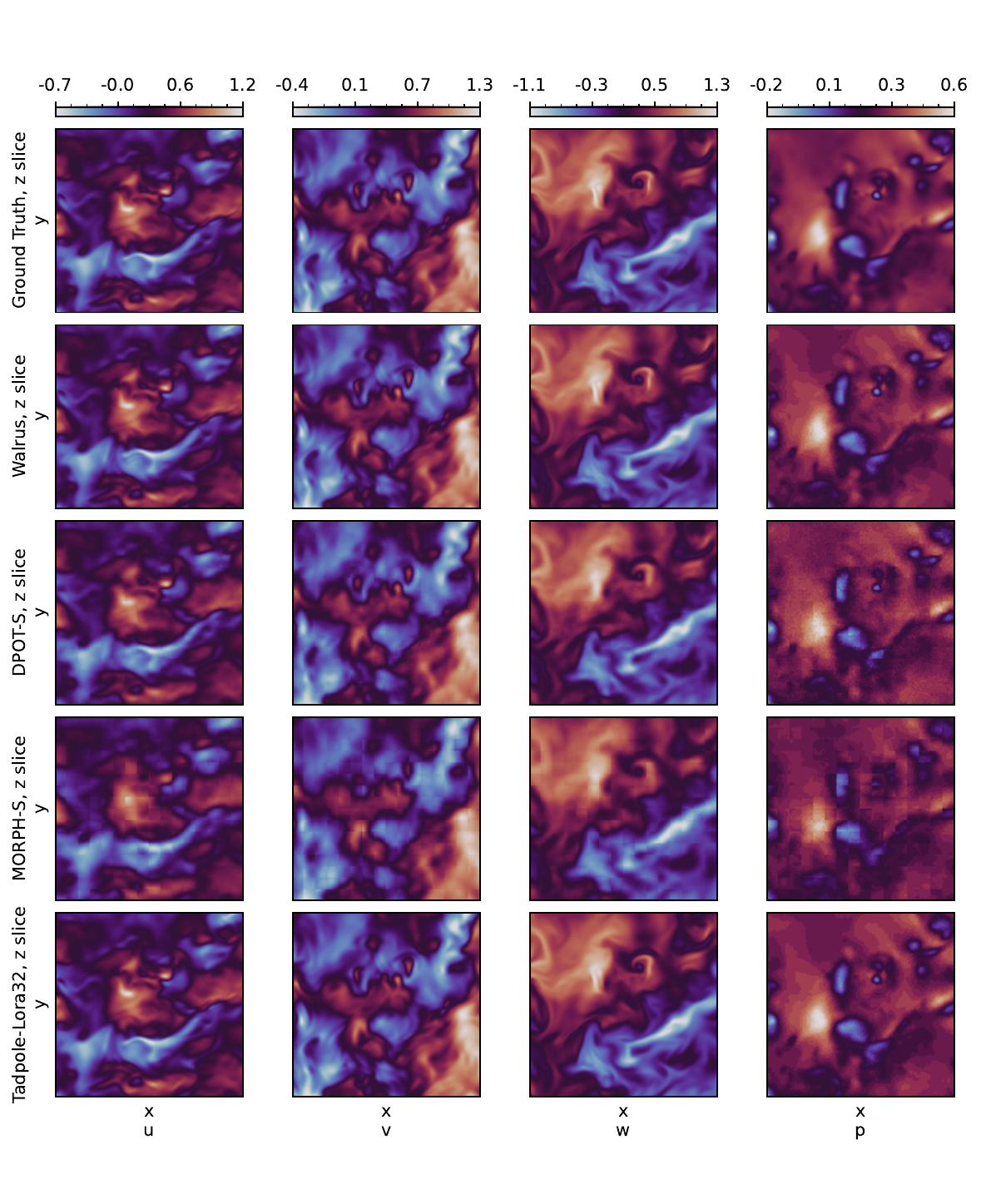}
    \hspace{10pt}
    \includegraphics[scale=0.33]{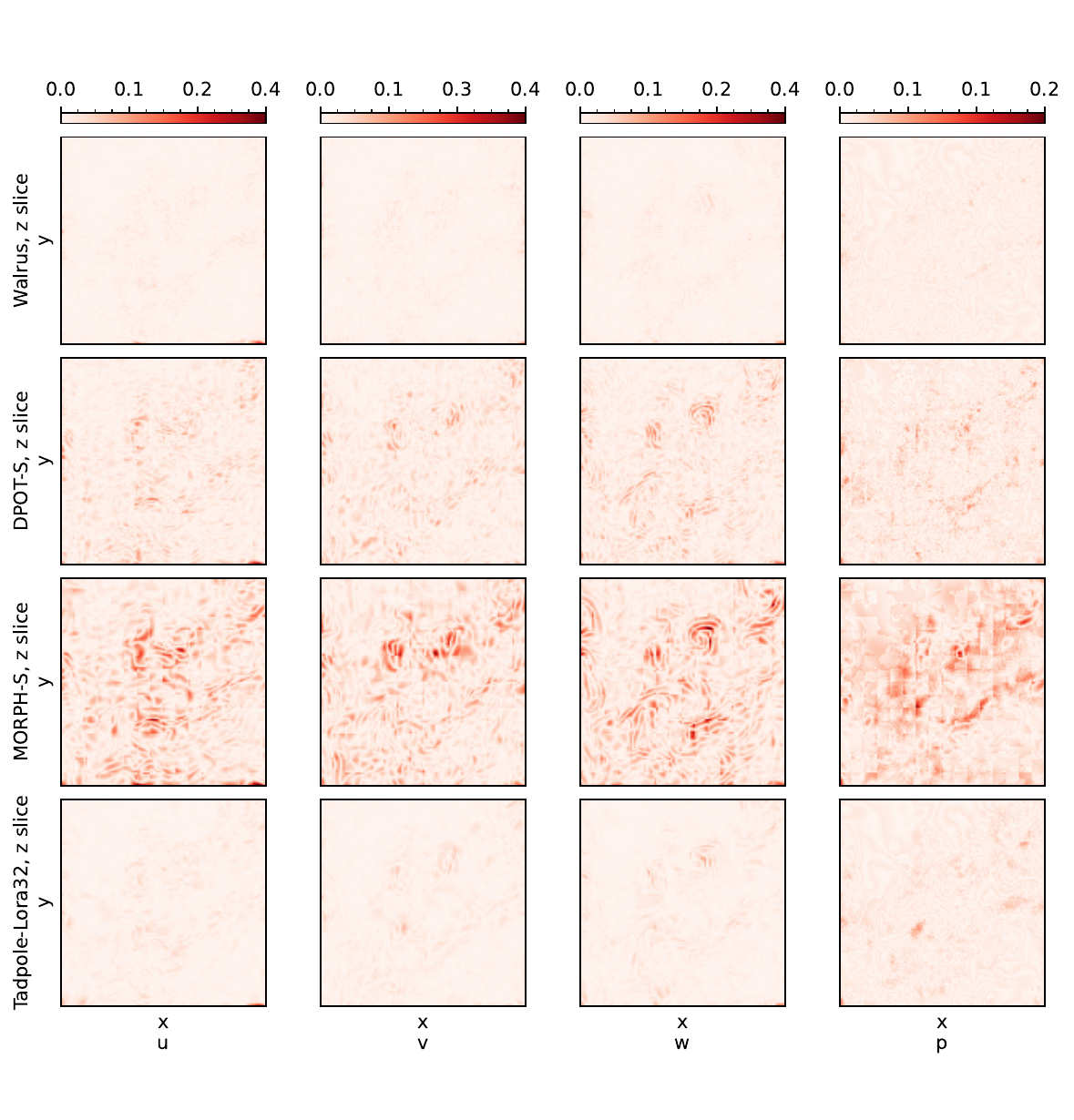}
  \caption{
    Visualization of the prediction (left) of \iso{} and the corresponding absolute error (right) at the \textbf{first rollout step} and the slice where $z=Z/2$.
    }
    \label{fig:vis_dynamics_z_1}
\end{figure}

\begin{figure}[htbp]
    \centering
    \includegraphics[scale=0.33]{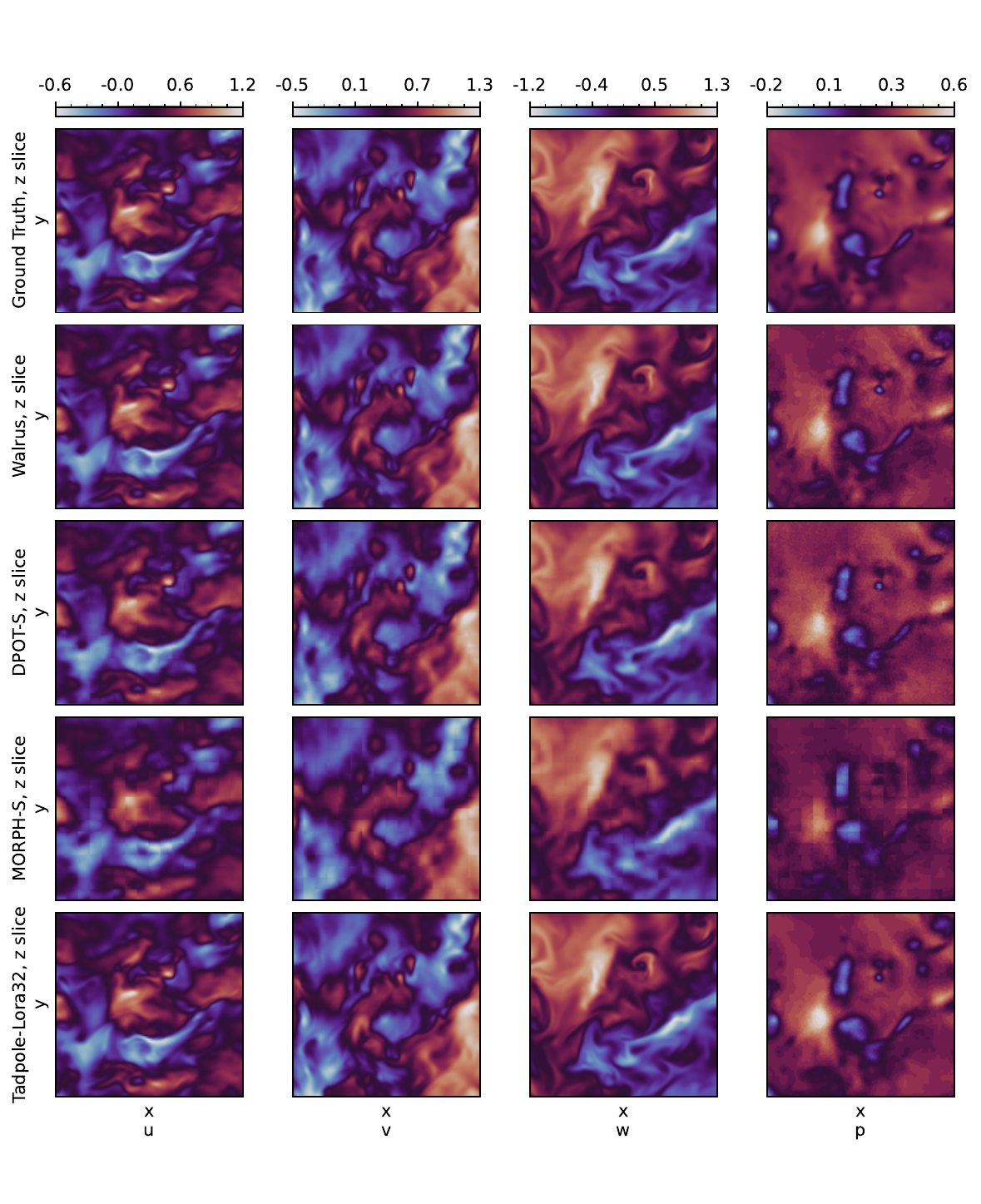}\hspace{10pt}
    \includegraphics[scale=0.33]{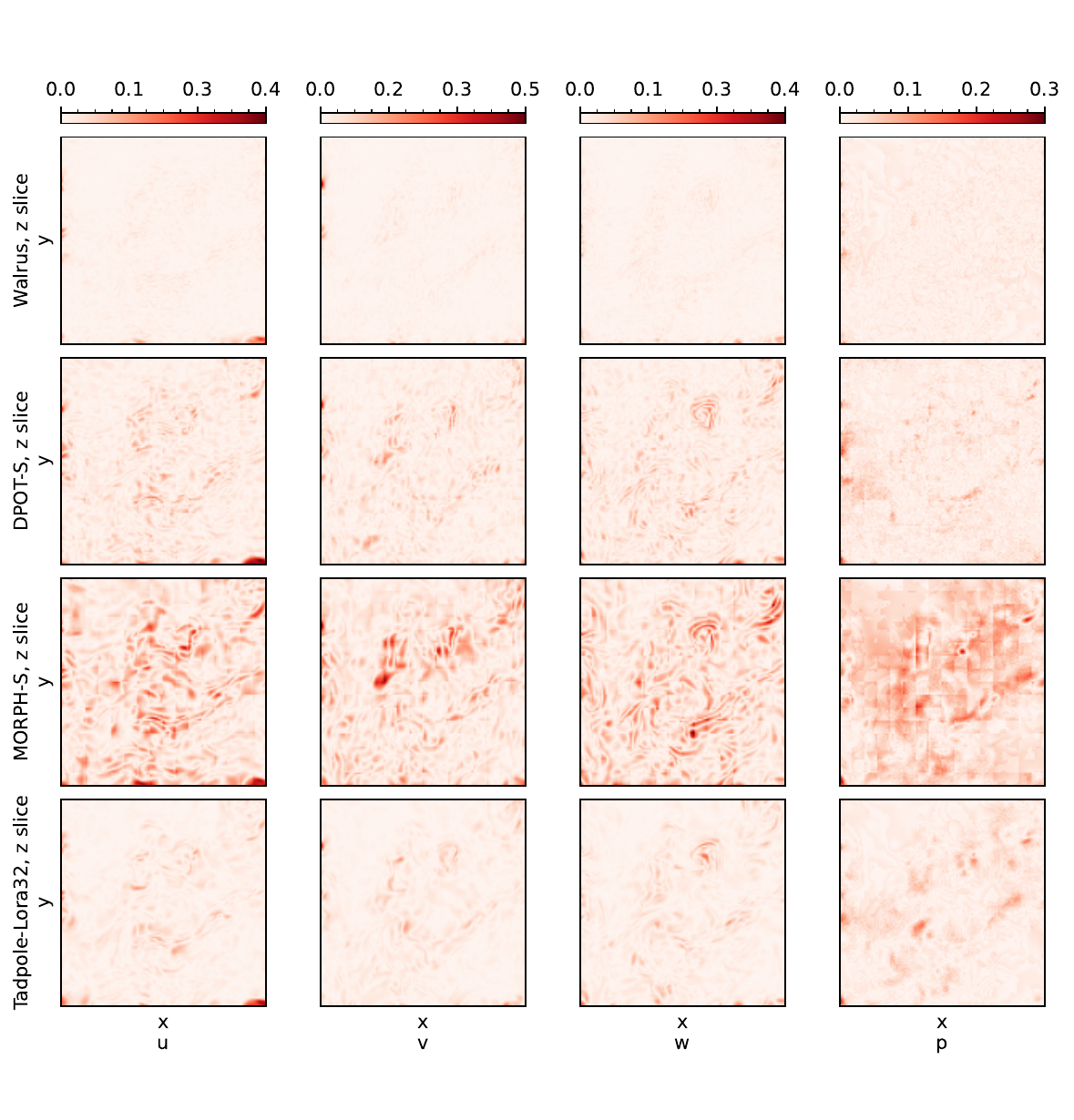}
  \caption{
    Visualization of the prediction (left) of \iso{} and the corresponding absolute error (right) at the \textbf{second rollout step} and the slice where $z=Z/2$.
    }
    \label{fig:vis_dynamics_z_2}
\end{figure}

\begin{figure}[htbp]
    \centering
    \includegraphics[scale=0.4]{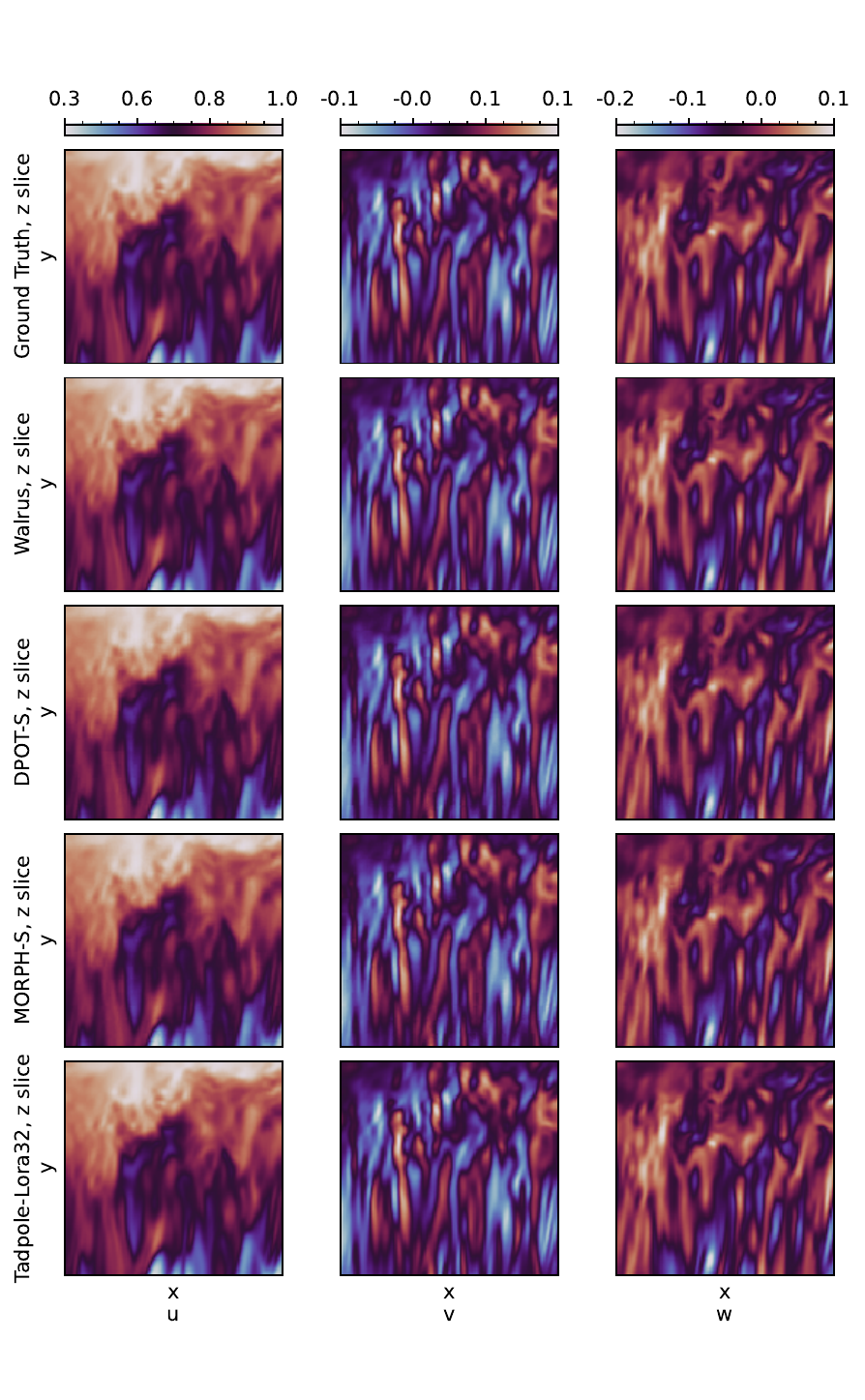}\hspace{10pt}
    \includegraphics[scale=0.4]{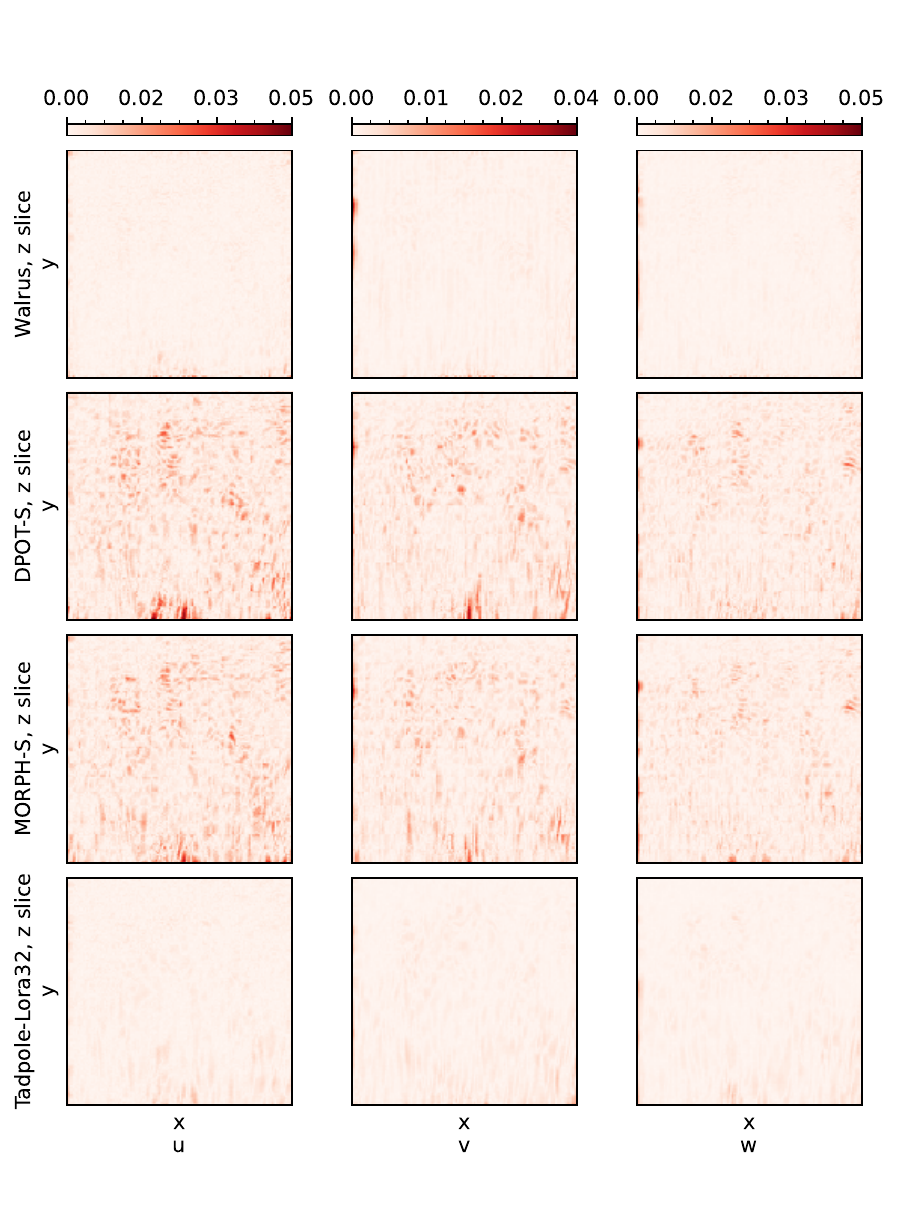}
  \caption{
    Visualization of the prediction (left) of \tbl{} and the corresponding absolute error (right) at the \textbf{first rollout step} and the slice where $z=Z/2$.
    }
    \label{fig:vis_dynamics_z_1_bl}
\end{figure}

\begin{figure}[htbp]
    \centering
    \includegraphics[scale=0.4]{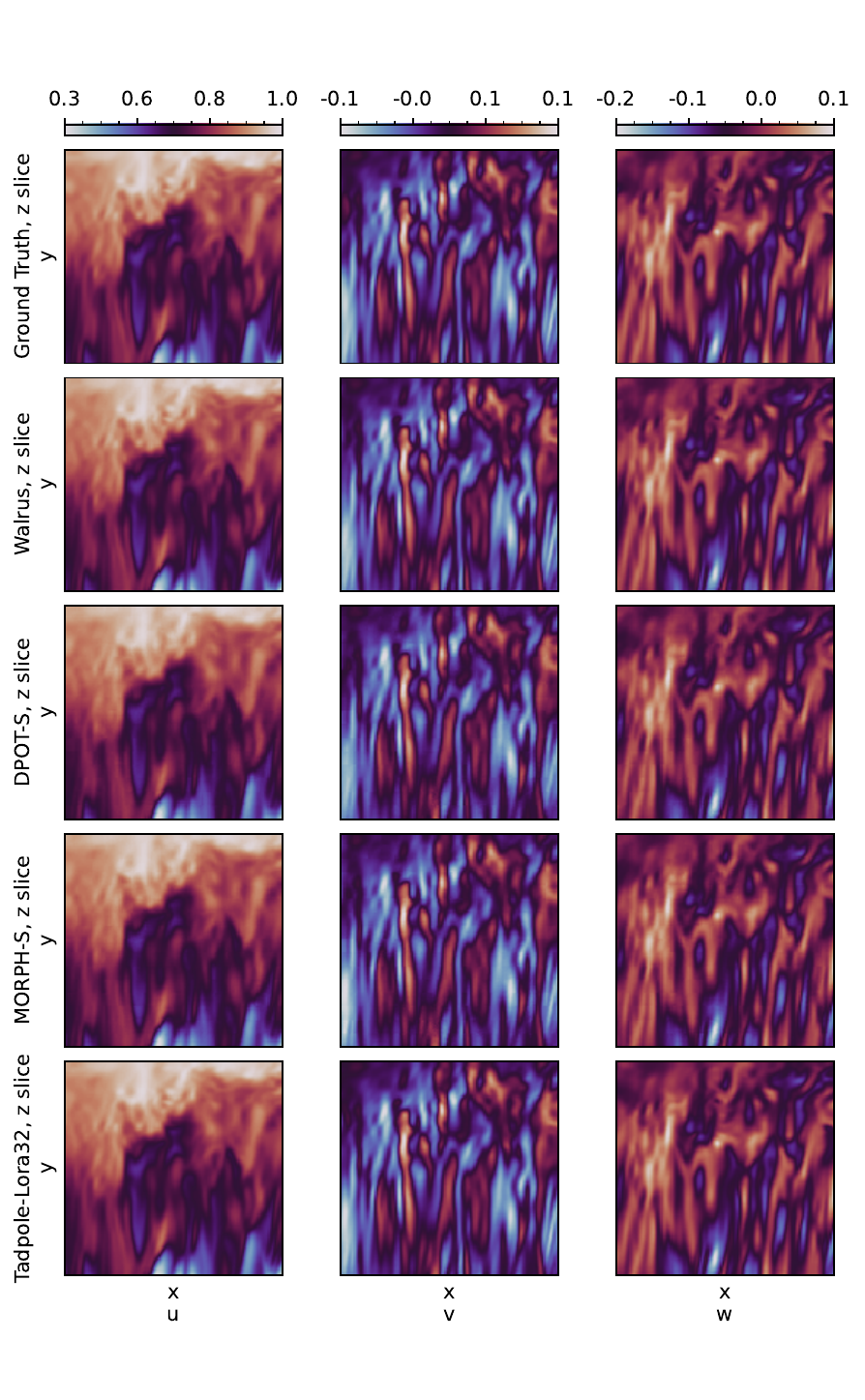}\hspace{10pt}
    \includegraphics[scale=0.4]{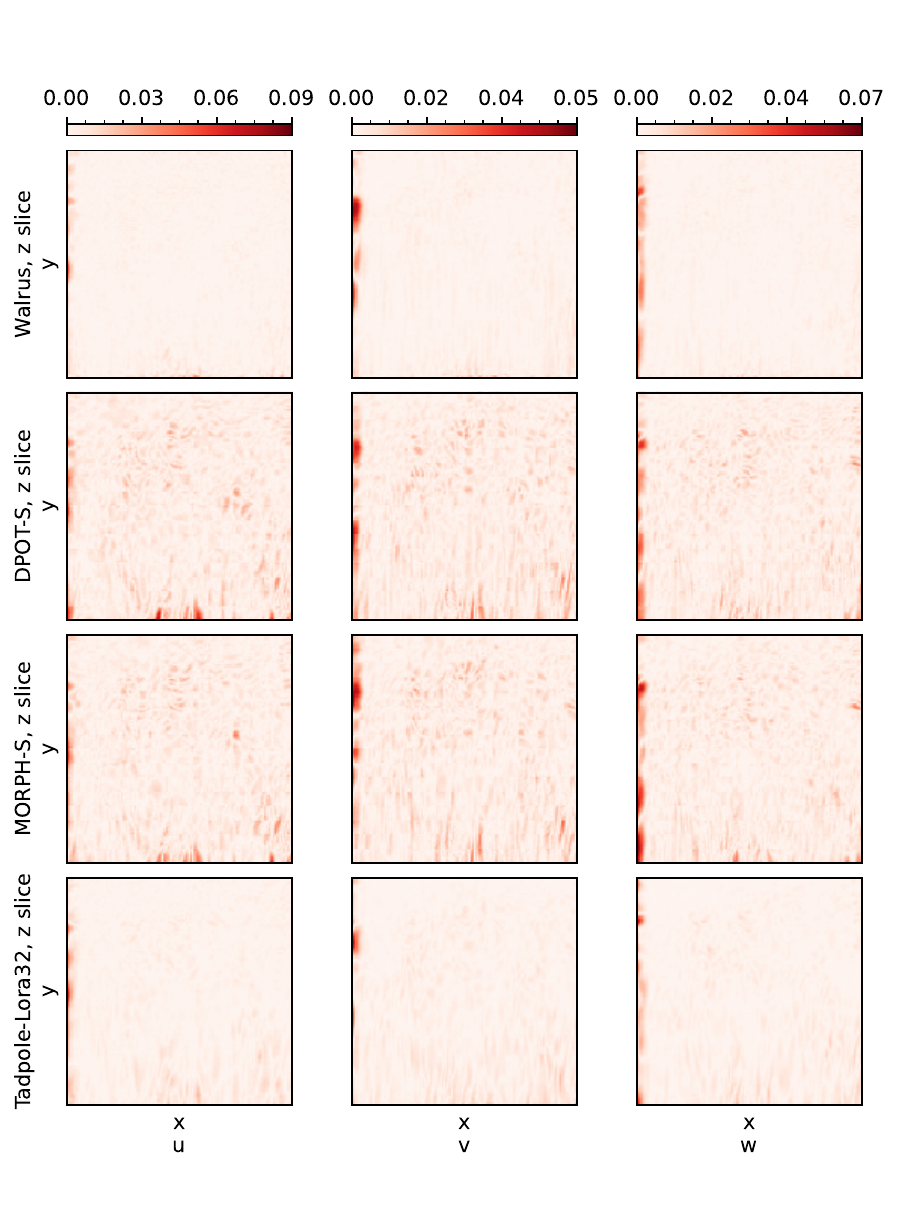}
  \caption{
    Visualization of the prediction (left) of \tbl{} and the corresponding absolute error (right) at the \textbf{second rollout step} and the slice where $z=Z/2$.
    }
    \label{fig:vis_dynamics_z_2_bl}
\end{figure}

\begin{figure}[htbp]
    \centering
    \includegraphics[scale=0.4]{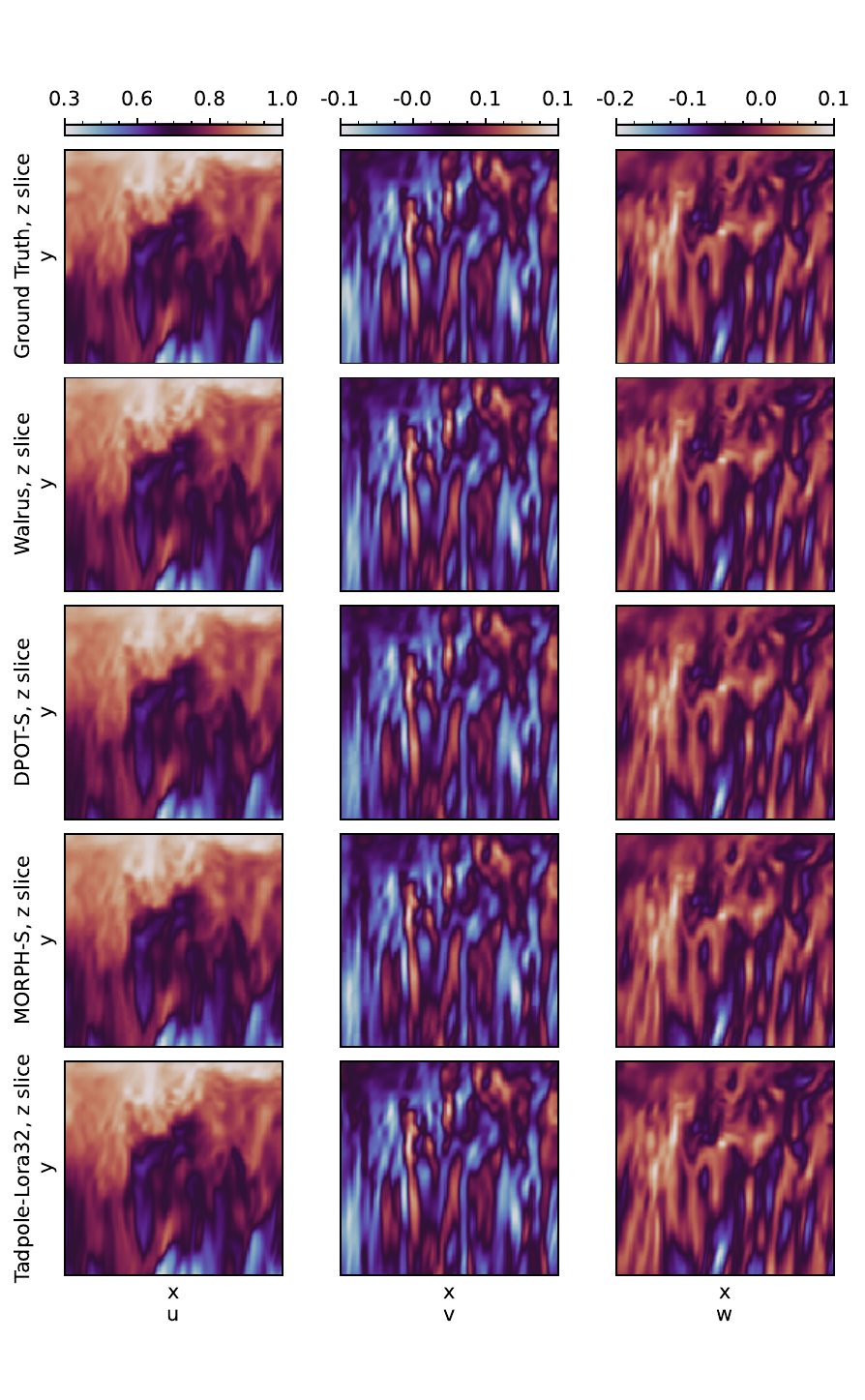}\hspace{10pt}
    \includegraphics[scale=0.4]{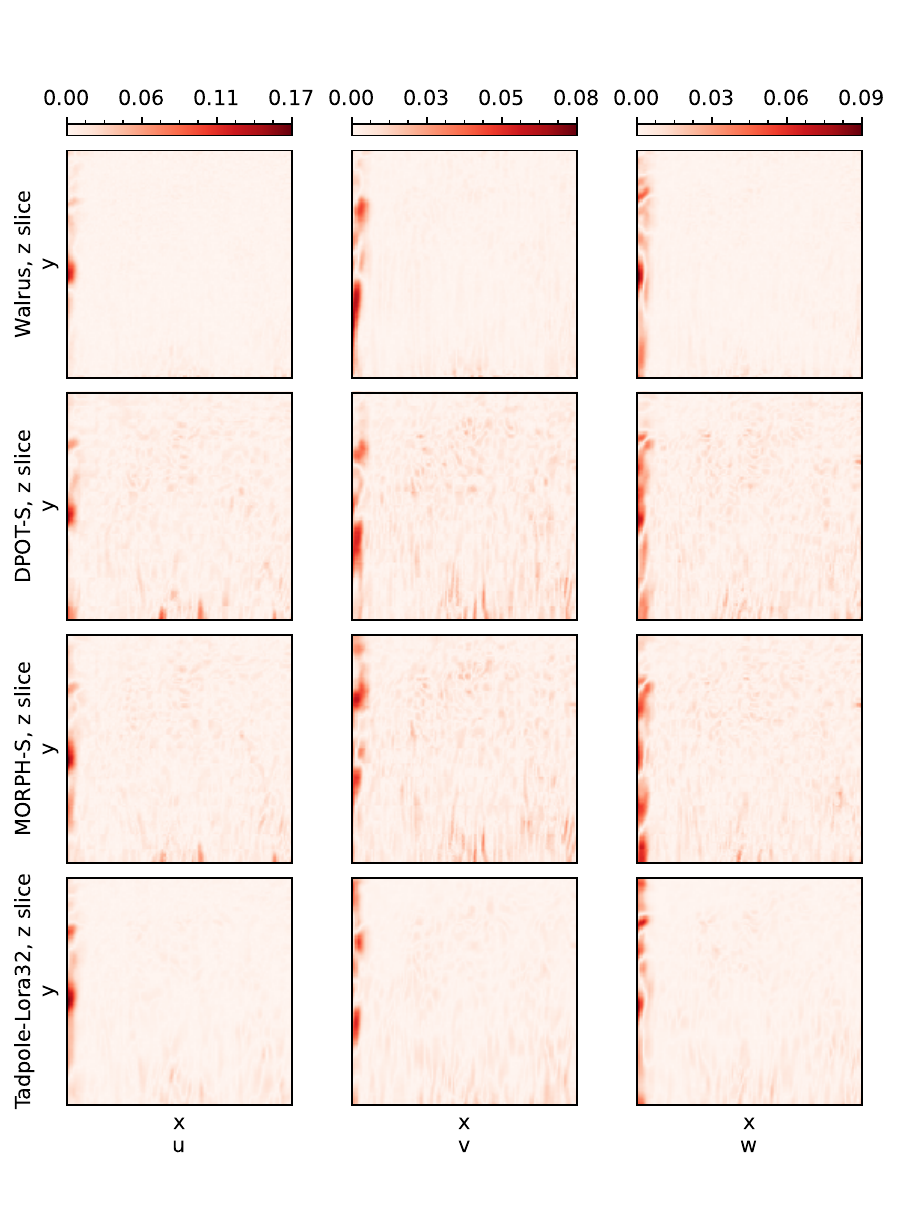}
  \caption{
    Visualization of the prediction (left) of \tbl{} and the corresponding absolute error (right) at the \textbf{third rollout step} and the slice where $z=Z/2$.
    }
    \label{fig:vis_dynamics_z_3_bl}
\end{figure}

\FloatBarrier
\newpage
\section{Dataset and online learning setups\label{sec:dataset_details}}
\subsection{Pre-training dataset\label{sec:dataset_details:pre-training} }

A significant challenge in training 3D physics foundation models is the production, storage, and efficient loading of large-scale spatiotemporal datasets. To illustrate the magnitude of this bottleneck, consider a single trajectory of $100$ frames on a $384^3$ grid for a vector field with three components. In half-precision floating-point format, a single trajectory requires approximately $30\,\text{GB}$ of storage. Scaling this to a dataset of just $100$ trajectories per phenomenon across $10$ different physical phenomena results in a storage requirement between $10\,\text{TB}$ and $100\,\text{TB}$. 

Beyond storage, the I/O overhead required to shuffle and stream this data to GPUs often exceeds the computational cost of the training step itself. While many 3D phenomena are computationally expensive to simulate (requiring at least minutes per frame), there exists a class of semi-linear Partial Differential Equations (PDEs) that admit fast, stable solutions via spectral methods. By leveraging efficient spectral solvers on modern GPUs, we decouple the training process from disk I/O. This procedural, online data generation strategy enables access to a theoretically infinite dataset with great spectral diversity and significantly lower engineering overhead than traditional offline storage.

This appendix details the mathematical formulation of the governing equations, the spectral solver implementation, and the communication strategy used to bridge the simulation and training environments. Corresponding numerical solvers have been independently released as TorchFSM~\cite{liu_2025_torchfsm} and can be used in other physics-based deep learning research.

\subsubsection{Overview of equations and their configurations\label{sec:pretraining_equation}}

Consider a scaled three-dimensional unit cube $\Omega = (0, L)^3 \subset \mathbb{R}^3$ in which $L$ describes the extent along each dimension. We assume periodic boundary conditions in each direction. On this domain, we want to solve equations of the form
\begin{equation}\label{eq:semi-linear-pde}
  \partial_t u = \mathcal{L} u + \mathcal{N}(u),
\end{equation}
where $\mathcal{L}$ describes a linear differential operator and $\mathcal{N}(\cdot)$ a nonlinear differential operator. Specifically, we are interested in equations in which the order of derivatives in $\mathcal{L}$ is higher than in $\mathcal{N}(\cdot)$. PDEs of this form are also called \emph{semi-linear}. We select a diverse set of equations to cover a wide range of spectral characteristics, including diffusive, dispersive, shock-forming, chaotic, and pattern-forming dynamics. \cref{tab:pde-overview} summarizes the mathematical formulation of the selected equations. The specific sampling parameters (time-steps $\Delta t$, grid resolutions $N$, and warmup periods) are detailed in the subsequent implementation section (see \cref{tab:pde-discretization-configs} and \cref{tab:pde-trajectory-configs}).
\formatedtable{Mathematical formulation of the considered PDEs. We denote scalar fields by $u$ and vector fields by $\mathbf{u}$. Parameters drawn from distributions are sampled per simulator instance, see \cref{sec:sampling-strategies}.}{tab:pde-overview}{
\begin{tabular}{l c c c}
\toprule
\textbf{Equation} & \textbf{Linear Term} $\mathcal{L}$ & \textbf{Nonlinear Term} $\mathcal{N}$ & \textbf{Parameters} \\
\midrule
Diffusion & $\nu \Delta u$ & $0$ & $\nu \sim \mathcal{U}(5\text{e-}4, 5\text{e-}3)$ \\
Hyper-Diffusion & $-\zeta \Delta^2 u$ & $0$ & $\zeta \sim \mathcal{U}(5\text{e-}4, 5\text{e-}3)$ \\
Burgers & $\nu \Delta \mathbf{u}$ & $-\mathbf{u} \cdot \nabla \mathbf{u}$ & $\nu \sim \mathcal{U}(1\text{e-}3, 5\text{e-}3)$ \\
Korteweg-de Vries (KdV) & $\xi (\mathbf{1} \cdot \nabla) \Delta \mathbf{u}$ & $-\mathbf{u} \cdot \nabla \mathbf{u}$ & $\xi = -6$ \\
Kuramoto-Sivashinsky (KS) & $-(\Delta + \Delta^2) u$ & $-\frac{1}{2} \|\nabla u\|^2$ & None \\
Fisher-KPP & $\nu \Delta u + r u$ & $-r u^2$ & $\nu \sim \text{Log}\;\mathcal{U}(1\text{e-}4, 0.02)$ \\
& & & $r \sim \mathcal{U}(5, 15)$ \\
Swift-Hohenberg & $r u - (1+\Delta)^2 u$ & $u^2 - u^3$ & $r=0.1$ \\
\bottomrule
\end{tabular}%
}

\textbf{Diffusion and Hyper-Diffusion Equation}

The diffusion equation describes an isotropic smoothing process where the initial energy dissipates over time. The spectrum decays radially as $|\hat{u}(\mathbf{k})| \propto \exp(-\nu \|\mathbf{k}\|_2^2 t)$. To introduce higher-order damping effects often seen in turbulence modeling, we also include the hyper-diffusion equation, which applies a fourth-order Laplacian, resulting in a steeper quartic spectral decay $|\hat{u}(\mathbf{k})| \propto \exp(-\zeta \|\mathbf{k}\|_2^4 t)$.

\textbf{Burgers Equation}

We utilize the viscous 3D Burgers' equation to model shock formation and the transfer of energy from large to small scales. The solution is a three-dimensional vector field. This nonlinear PDE generates sharp gradients, which are essential for training the model to resolve high-frequency features.

\textbf{Korteweg-de-Vries (KdV) Equation}

While classically a 1D equation describing soliton waves, we extend the KdV dynamics to 3D by employing a similar nonlinearity as the Burgers equation and by using an extension of the dispersion effect to higher dimensions. This system balances nonlinearity with dispersion rather than diffusion.

\textbf{Kuramoto-Sivashinsky (KS) Equation}

The KS equation is characterized by a negative diffusion term (instability at low wavenumbers) stabilized by hyper-diffusion at high wavenumbers. This balance between energy production, energy movement based on the nonlinearity and high-frequency dissipation is known for exhibiting spatiotemporal chaos, yielding rich spectra.

\textbf{Fisher-KPP Equation}

The Fisher-Kolmogorov-Petrovsky-Piscounov (Fisher-KPP) equation combines diffusion with a logistic reaction term. The polynomial nonlinearity contributes interesting spectral components.

\textbf{Swift-Hohenberg (SH) Equation}

This equation is a canonical model for pattern formation leading to the emergence of complex spatial structures. We select parameters to ensure the system remains in the pattern-forming regime. These patterns are characterized by spectral richness.

\formatedtable{Discretization configurations for each considered PDE system. Note that values differ depending on the resolution $N$ on which the system is simulated. The effective $\Delta t$ realized when recording the trajectory is  the one given here multiplied by the \textbf{Save Frequency}, allowing substepping on stiffer configurations. After data recording, each frame is voxel-wise normalized to be within \textbf{Value Range}, which was precomputed based on reasonable limits seen in the data distribution.}{tab:pde-discretization-configs}{
\begin{small}
    \begin{tabular}{c c c c c}
        \toprule
        \textbf{Equation} & \textbf{Extent} $L$  & \textbf{Time-Step Size} $\Delta t$ & \textbf{Save Frequency} & \textbf{Value Range} \\
        \midrule
        Diffusion
        &
        $1$
        &
        $\begin{aligned}
            5\text{e-}4 \;& N=64 \\
            5\text{e-}5 \;& N \in \{128, 256, 384\}
        \end{aligned}$
        &
        1
        &
        $[-1, 1]$
        \\
        \midrule
        Hyper-Diffusion
        &
        $1$
        &
        $\begin{aligned}
            5\text{e-}4 \;& N=64 \\
            5\text{e-}5 \;& N \in \{128, 256, 384\}
        \end{aligned}$
        &
        1
        &
        $[-1, 1]$
        \\
        \midrule
        Burgers
        &
        1
        &
        $\begin{aligned}
            5\text{e-}3 \;& N = 64\\
            2\text{e-}3 \;& N = 128\\
            1\text{e-}3 \;& N \in \{256, 384\}\\
        \end{aligned}$
        &
        $\begin{aligned}
            1 \;& N =64\\
            2 \;& N = 128\\
            5 \;& N \in \{256, 384\}\\
        \end{aligned}$
        &
        $[-1, 1]$
        \\
        \midrule
        Korteweg-de-Vries 
        &
        1
        &
        2\text{e-}6
        &
        2
        &
        $[-1.25, 1.25]$
        \\
        \midrule
        Kuramoto-Sivashinsky
        &
        64
        &
        0.1
        &
        1
        &
        $[-25, 25]$
        \\
        \midrule
        Fisher-KPP
        &
        1
        &
        $\begin{aligned}
            1\text{e-}3 \;& N \in \{64, 128\}\\
            5\text{e-}4 \;& N \in \{256, 384\}\\
        \end{aligned}$
        &
        $\begin{aligned}
            1 \;& N \in \{64, 128\}\\
            2 \;& N \in \{256, 384\}\\
        \end{aligned}$
        &
        $[0, 1]$
        \\
        \midrule
        Swift-Hohenberg
        &
        20
        &
        $\begin{aligned}
            0.1 \;& N =64\\
            0.02 \;& N \in \{128, 256, 384\}\\
        \end{aligned}$
        &
        $\begin{aligned}
            1 \;& N=64\\
            2 \;& N=128\\
            5 \;& N \in \{256, 384\}\\
        \end{aligned}$
        &
        $[-2, 3]$
        \\
        \bottomrule
    \end{tabular}
\end{small}
}

\formatedtable{Configurations for recording trajectories. \textbf{Warmup} is the number of frames starting from the initial condition that are being discarded \emph{before} recording starts. This ensures we are within the physics state space we deem most interesting. The number of frames recorded per trajectory (after warmup) is given by \textbf{Trajectory Length}. Before moving on to the next simulator configuration, we repeat the recording process \textbf{Num Runs} times. As such, each setup always contributes $30$ frames. We also conducted an ablation with an offline local training dataset of approximately 500GB. The \textbf{Offline Train Num Trajectory} settings denote how many trajectories are produced for each split of the data in the offline setup. }{tab:pde-trajectory-configs}{
\begin{small}
    \begin{tabular}{c c c c c}
        \toprule
        \textbf{Equation} & \textbf{Warmup} & \textbf{Traj. Length} & \textbf{Num Runs} & \textbf{Offline Train Num Traj.} \\
        \midrule
        Diffusion
        &
        0
        &
        2
        &
        15
        &
        75
        \\
        \midrule
        Hyper-Diffusion
        &
        0
        &
        2
        &
        15
        &
        75
        \\
        \midrule
        Burgers
        &
        $\begin{aligned}
            30 \;& N=64\\
            60 \;& N=128\\
            150 \;& N \in \{256, 384\}\\
        \end{aligned}$
        &
        30
        &
        1
        &
        5
        \\
        \midrule
        Korteweg-de-Vries
        &
        40
        &
        10
        &
        3
        &
        15
        \\
        \midrule
        Kuramoto-Sivashinsky
        &
        500
        &
        30
        &
        1
        &
        5
        \\
        \midrule
        Fisher-KPP 
        &
        $\begin{aligned}
            20 \;& N \in \{64, 128\}\\
            40 \;& N \in \{256, 384\}\\
        \end{aligned}$
        &
        10
        &
        3
        &
        15
        \\
        \midrule
        Swift-Hohenberg 
        &
        0
        &
        30
        &
        1
        &
        5
        \\
        \bottomrule
    \end{tabular}
\end{small}

}

\subsubsection{Fast semi-linear PDE solvers in PyTorch \label{sec:solver}}

We solve the semi-linear PDEs $\partial_t u = \mathcal{L}u + \mathcal{N}(u)$ using Exponential Time Differencing Runge-Kutta (ETDRK) methods. These schemes are particularly well-suited for stiff PDEs where the linear term $\mathcal{L}$ contains high-order derivatives (e.g., $\Delta^2$). By treating the linear part exactly via an integrating factor, we avoid the severe time-step restrictions typical of explicit schemes.

We discretize the scaled unit cube $\Omega = (0, L)^3$ into $N$ intervals of size $\Delta x$ per dimension (yielding $N^3$ total number of voxels). Then we consider the left end of each interval a nodal degree of freedom, i.e., in one dimension the grid points are located at positions $\left[0, \Delta x, 2 \Delta x, \dots, (N-1)
\Delta x\right]^T \in \mathbb{R}^N$. This also means that the left end of the domain is considered a degree of freedom, while the right end is not, which naturally encodes periodic boundary conditions and is a prerequisite for most Fast Fourier Transform implementations. The three-dimensional grid is given by the tensor product of the one-dimensional coordinates.

We denote by $\mathbf{u}_t \in \mathbb{R}^{C \times N \times N \times N}$ a state array at time point $t$ with $C$ channels and a value for each valid degree of freedom. Applying a three-dimensional discrete Fourier $\mathcal{F}$ transformation yields $\hat{\mathbf{u}}_t \in \mathbb{C}^{C \times N \times N \times N}$ which is an array of the same shape but with complex-valued entries \footnote{Note that given all our PDEs are real-valued, one could have used a real-valued FFT that typically halves the number of entries in the last array axis and halves the compute cost. However, for simplicity, we opted for the general FFT since we did not observe significant performance degradations in our online learning setup.}.

In state space, time integration with a fixed time step size $\Delta t$ can be represented by the time-stepping operator $\mathcal{P}$ which yields
\begin{equation}
    \mathbf{u}_{t+\Delta t} = \mathcal{P}(\mathbf{u}_t).
\end{equation}
Due to the diagonalization of derivative operators in Fourier space, we choose to perform time integration in the spectral domain via
\begin{equation}
    \mathbf{u}_{t+\Delta t} = \mathcal{F}^{-1}\left(
        \hat{\mathcal{P}}\left(
            \mathcal{F}\left(
                \mathbf{u}_t
            \right)
        \right)
        \right),
\end{equation}
in which the spectral time-stepping operator $\hat{\mathcal{P}}(\cdot)$ is implemented via a two-stage process
\begin{align}
  \hat{\mathbf{u}}_* &= \exp(\hat{\mathcal{L}} \Delta t) \odot \hat{\mathbf{u}}_t + \frac{\exp(\hat{\mathcal{L}} \Delta t) - 1}{\hat{\mathcal{L}}} \odot \hat{\mathcal{N}}(\hat{\mathbf{u}}_t),\label{eq:etdrk}
  \\
  \hat{\mathbf{u}}_{t+\Delta t} &= \hat{\mathbf{u}}_* + \frac{\exp(\hat{\mathcal{L}} \Delta t) - 1 - \hat{\mathcal{L}} \Delta t}{\hat{\mathcal{L}}^2 \Delta t} \left( \hat{\mathcal{N}}(\hat{\mathbf{u}}_*) - \hat{\mathcal{N}}(\hat{\mathbf{u}}_t) \right).\label{eq:etdrk_2}
\end{align}
Here, $\hat{\mathcal{L}}$ and $\hat{\mathcal{N}}(\cdot)$ are the linear and nonlinear differential operators in Fourier space, respectively. Both can be built using spectral derivative operators.
As a property of spectral differentiation, these operators diagonalize, which allows all the operations in \cref{eq:etdrk,eq:etdrk_2} to be evaluated pointwise. This includes an elementwise exponentiation that is at the core of the ETDRK methods. For the nonlinear operator in Fourier space $\hat{\mathcal{N}}(\cdot)$, a pseudo-spectral evaluation strategy using transformation to the state space, evaluating the nonlinearity, and transforming back into Fourier space is used. We supplement this with appropriate anti-aliasing. The time integration procedure of \cref{eq:etdrk,eq:etdrk_2} is of second order consistency. We use this ETDRK2 method for most equations because it offers the best compromise between speed, accuracy, and stability in single precision across our tested systems. Only for the KdV equation, we use the fourth-order ETDRK version \citep{ETDRK2002}.

Since these methods are based on array computations, they operate efficiently within modern acceleration frameworks like PyTorch and can therefore be easily ported to GPUs. For more details on Fourier pseudo-spectral ETDRK methods, their implementation, and a discussion of their limitations, we refer to \citep{ETDRK2002,ETDRK2005,koehler2024apebench}.

\subsubsection{Overview of initializers\label{sec:initial_condition}}

For many equations, their long-term behavior throughout the trajectory is influenced by the distribution of the initial state. For Tadpole, we used five initialization routines chosen to ensure a wide range of spectral representations. The primary difference between them is how the spectrum behaves as a function of wavenumber $\mathbf{k}$.
Some of the initialization routines additionally sample hyperparameters according to \cref{tab:ic-configs}.

For most initializers, we apply a post-processing step that clamps the state to the range $(c_{\text{min}}, c_{\text{max}})$. These limits $c_{\text{min}}$ and $c_{\text{max}}$ can either be fixed or randomly drawn. We normalize the initial conditions to ensure that their order of magnitude is around $1$. Note that this normalization is different from the normalization of each recorded frame according to \cref{tab:pde-discretization-configs}.

Each PDE system uses a distinct set of initializers and their respective hyperparameters. This is to enable sufficiently wide spectral representation in the created frames while ensuring stable time integration. We note down the match in \cref{tab:pde-and-initializers}.

\begin{figure}[t]
    \centering
    \includegraphics[width=\linewidth]{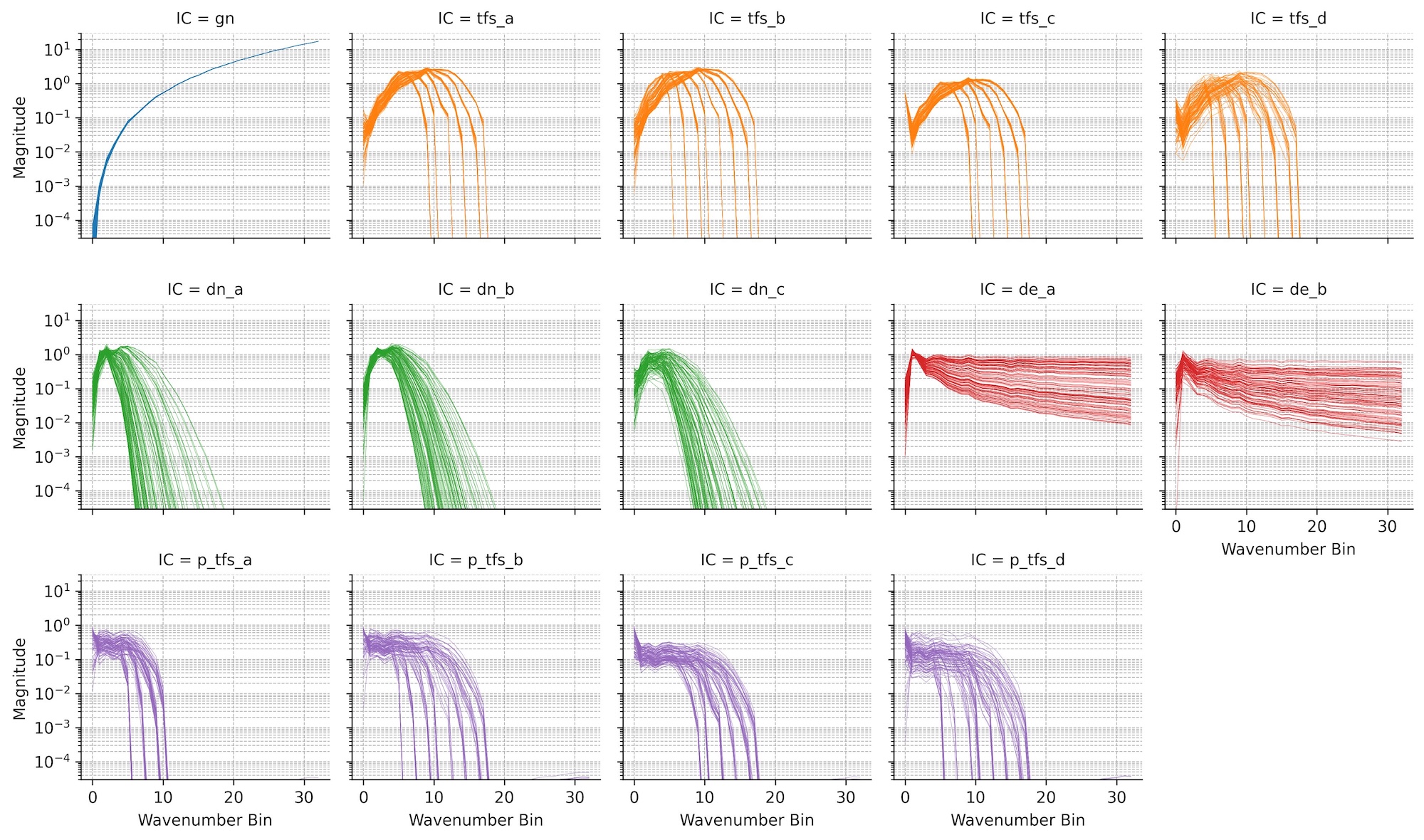}
    \caption{
        Radially \emph{shell-aggregated} magnitude spectra (100 samples per distribution). 
        \textbf{Row 1 (GN \& TFS):} Gaussian (white) noise (GN) exhibits quadratic growth due to the volume of spherical shells in 3D Fourier space. The Truncated Fourier Series (TFS) initializers show distinct spectral cutoffs determined by $k_{\text{limit}}$. Note that TFS-D exhibits vertical variation due to its randomized normalization bounds.
        \textbf{Row 2 (DN \& DE):} Diffused Noise (DN) follows an exponential quadratic decay ($\sim \exp(-\nu \|\mathbf{k}\|_2^2)$), with DN-A appearing to have more spectrally compact samples due to higher diffusivity sampling. Decayed Energy (DE) follows a rough power-law distribution with significant high-frequency contributions.
        \textbf{Row 3 (P-TFS):} The Poisson (P-TFS) initializers retain the cutoff frequency of the source TFS but exhibit smoother maxima and a steeper decay within the active modes, consistent with the smoothing properties of the inverse Laplacian.
    }
    \label{fig:initial-condition-spectra}
\end{figure}

In \cref{fig:initial-condition-spectra}, we display 100 samples for each initial condition distribution.

\textbf{Pixel-Wise Gaussian Noise (GN)} 

The state array for a single channel $\mathbf{u}_0 \in \mathbb{R}^{1 \times N \times N \times N}$ is built by drawing each degree of freedom value identically and independently from a standard normal distribution
\begin{equation}
    u_j \sim \mathcal{N}(0, 1).
\end{equation}
This corresponds to white noise. We use this initializer for the KS equation whose long-term behavior is independent of the initial state. For the other PDE systems, this initializer contains too much high-frequency content. We only use it as a starting point to apply spectral modifications.

\textbf{Truncated Fourier Series (TFS)}

We generate a spectrally compact IC by filtering white noise in the Fourier domain. Let $\hat{\mathbf{u}} = \mathcal{F}(\mathbf{u}_{\text{noise}})$. We apply a three-dimensional binary mask $M(\mathbf{k})$ which is $1$ if $\mathbf{k} \in [0, k_\text{limit}]^3$ and $0$ otherwise:
\begin{equation}
    \mathbf{u}_{\text{TFS}} = \mathcal{F}^{-1}( M \odot \hat{\mathbf{u}} ).
\end{equation}
This ensures energy is distributed only among specific low-to-mid frequency modes. Since the mask does not alter the spectrum within its active region, energy is equally distributed among all active modes. We draw the limit $k_\text{limit}$ from a discrete uniform distributions with extents $k_\text{min}$ and $k_\text{max}$.

\textbf{Diffused Noise (DN)}

To generate smooth fields with physically natural decay, we initialize white noise and integrate the linear diffusion operator $\partial_t u = \nu \Delta u$ for a single time step of size $\Delta t=1$. The resulting magnitude spectrum follows an exponential quadratic decay:
\begin{equation}
    |\hat{u}(\mathbf{k})| \propto \exp(-\nu \|\mathbf{k}\|_2^2).
\end{equation}
We vary the smoothness of the IC by sampling the diffusivity parameter $\nu$.

\textbf{Decayed Energy (DE)}

This initializer creates Gaussian Random Fields (GRF) with a specific power-law spectral density. We explicitly enforce the amplitude of the Fourier modes to follow $|\hat{u}(\mathbf{k})| \propto \|\mathbf{k}\|_2^{-\alpha}$, where the exponent $\alpha$ is drawn uniformly from the range $(-5, -2)$. This directly controls the field's roughness.

\textbf{Poisson (P)}

We solve a Poisson equation $\Delta u = -f$, where the source term $f$ is generated via the Truncated Fourier Series (TFS) method described above. In the spectral domain, inversion of the Laplacian acts as a low-pass filter, scaling the spectrum by $\|\mathbf{k}\|_2^{-2}$. This results in smoother initial conditions than the raw TFS source. Since Poisson inversion on periodic boundaries does not move energy between the modes, we retain the characteristics of a compact spectrum (i.e., non-zero energy only in the modes leftover by the mask). However, within this patch, the magnitude follows an isotropic polynomial decay. We denote these configurations as P-TFS.

\formatedtable{These hyperparameter configurations for the initialization schemes are used to instantiate the initial condition distribution.
Normalization bounds are used to clamp the initial condition and to keep their order of magnitude consistent.
}{tab:ic-configs}{
\begin{tabular}{l l l}
\toprule
\textbf{Config Name} & \textbf{Parameters} & \textbf{Normalization Bounds} \\
\midrule
TFS-A & $k_\text{min}, k_\text{max} = \{3, 5\}$ & Fixed: $[-1, 1]$ \\
TFS-B & $k_\text{min}, k_\text{max} = \{3, 9\}$ & Fixed: $[-1, 1]$ \\
TFS-C & $k_\text{min}, k_\text{max} = \{5, 9\}$ & Fixed: $[0, 1]$ \\
TFS-D & $k_\text{min}, k_\text{max} = \{3, 9\}$ & Random: $\{c_{\text{min}}, c_{\text{max}} \} \sim \{\mathcal{U}(-1,-0.1),  \mathcal{U}(0.1, 1)\}$ \\
\midrule
DN-A & $\nu \sim \mathcal{U}(0.001, 0.01)$ & Fixed: $[-1, 1]$ \\
DN-B & $\nu \sim \mathcal{U}(0.001, 0.005)$ & Fixed: $[-1, 1]$ \\
DN-C & $\nu \sim \mathcal{U}(0.001, 0.005)$ & Random: $\{c_{\text{min}}, c_{\text{max}} \} \sim \{\mathcal{U}(-1,-0.1),  \mathcal{U}(0.1, 1)\}$ \\
\midrule
DE-A & $\alpha \sim \mathcal{U}(-5, -2)$ & Fixed: $[-1, 1]$ \\
DE-B & $\alpha \sim \mathcal{U}(-5, -2)$ & Random: $\{c_{\text{min}}, c_{\text{max}} \} \sim \{\mathcal{U}(-1,-0.1),  \mathcal{U}(0.1, 1)\}$ \\
\bottomrule
\end{tabular}%
}

\formatedtable{To achieve maximal spectral diversity, we pair specific initializers and certain PDE systems. See \cref{tab:ic-configs} for the specifics of each initialization configuration.}{tab:pde-and-initializers}{
    \begin{tabular}{c|c}
        \toprule
        \textbf{Equation} & \textbf{Eligible Initializers} \\
        \midrule
        Diffusion & TFS-D, DN-C, DE-B, P-TFS-D \\
        Hyper-Diffusion & TFS-D, DN-C, DE-B, P-TFS-D \\
        Burgers & TFS-A, DN-A, DE-A, P-TFS-A \\
        KdV & TFS-A, DN-A, DE-A, P-TFS-A \\
        KS & GN (Gaussian Noise) \\
        KPP-Fisher & TFS-C \\
        Swift-Hohenberg & TFS-B, DN-B, DE-A, P-TFS-B \\
        \bottomrule
    \end{tabular}
}

\subsubsection{Spectral distributions and the beneficial effect of cropping \label{sec:spectral_distribution}}

The combinations of partial differential equations, initialization distributions, and integration horizons were specifically chosen to expose the foundation model to a wide range of plausible physics states. To confirm this, we scraped 100,000 samples from a simulation server and analyzed their spectra. This equals $\approx 3.5\,\text{TB}$ of full-resolution simulation data, and this is approximately a tenth of what the Tadpole B-size model was exposed to during its pre-training.

\begin{figure}
    \centering
    \includegraphics[width=0.89\textwidth]{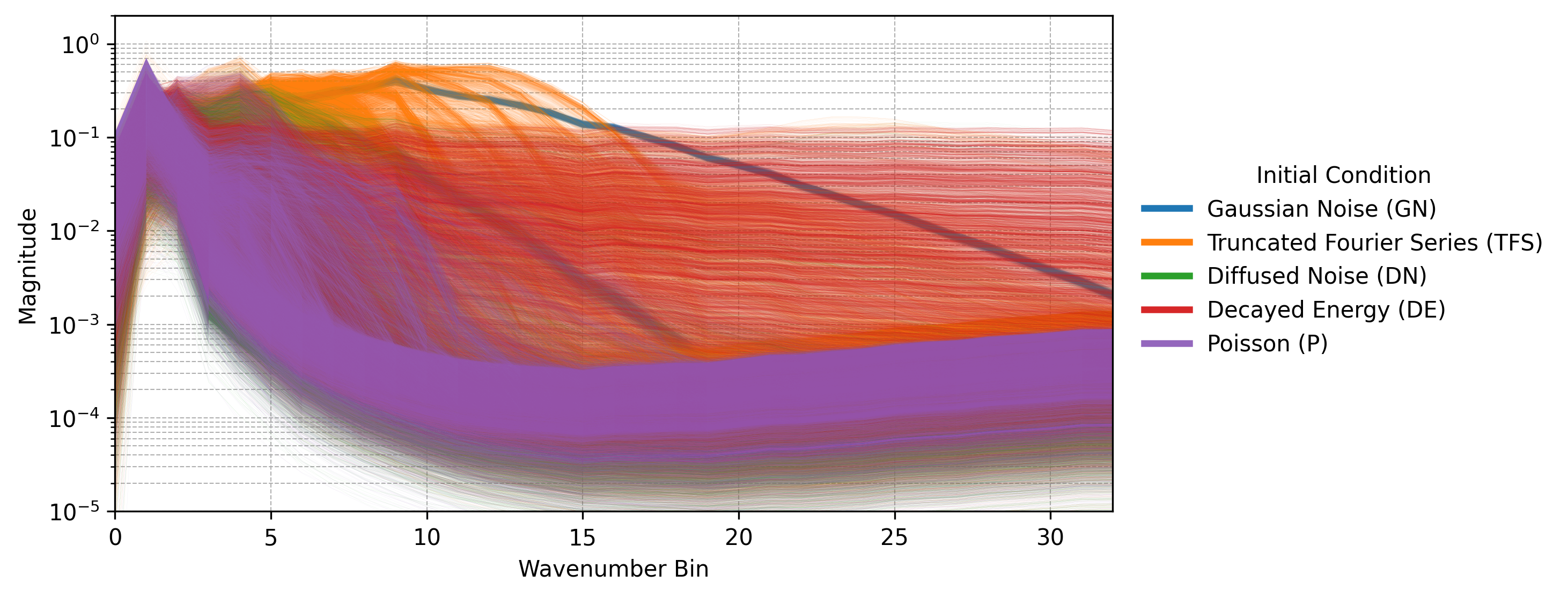}
    \caption{This collection of all \emph{shell-aggregated} spectra across 100,000 samples (about one tenth of the pre-training amount) from the simulation server highlights the diversity of states exposed to the foundational pre-training of Tadpole.}
    \label{fig:spectral_analysis_aggregated}
\end{figure}

\begin{figure}[t]
    \centering
    \includegraphics[width=0.95\textwidth]{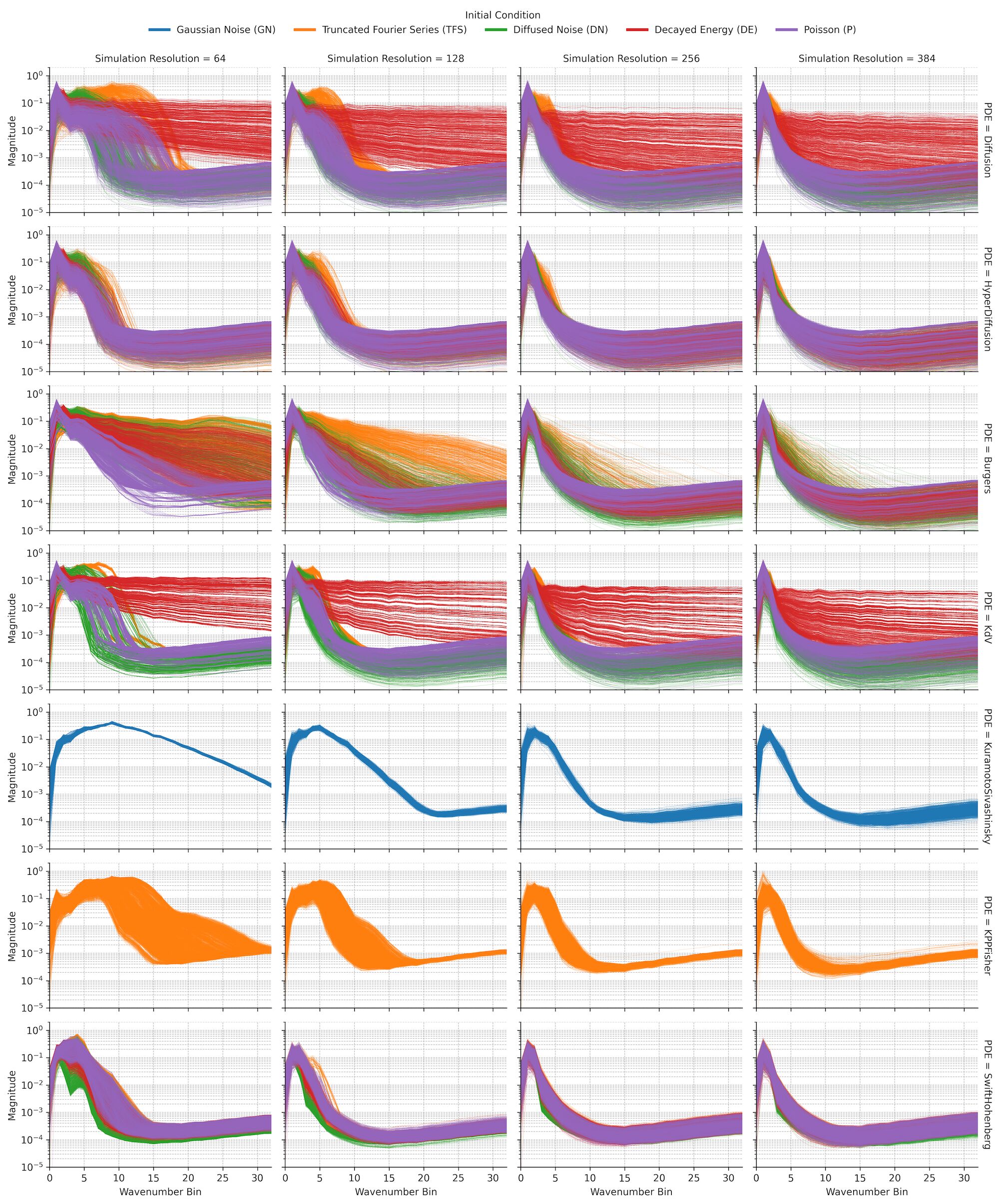}
    \caption{
        Distribution of Fourier coefficient magnitude across radially aggregated bins for a spectral analysis of the $64^3$ training crops based on 100,000 samples from the simulation server (about 10\% of the amount of data used for Tadpole B-size pre-taining). Each row represents a different PDE (according to \cref{tab:pde-overview}) and each column a different simulation resolution $\{64, 128, 256, 384\}$. Colors indicate different initial condition distributions, see \cref{tab:ic-configs} and \cref{tab:pde-and-initializers}. The states produced cover a large range of plausible physics spectra.
    }
    \label{fig:spectral_distibution}
\end{figure}

The analysis is based on the random $64^3$ crops of data used to train the model.
Since those crops are no longer ensured to be periodic, we first apply a Hann window per dimension
\begin{equation}
    w(n) = 0.5 - 0.5 \cos\left(\frac{2\pi{n}}{N-1}\right)
               \qquad 0 \leq n \leq N-1
\end{equation}
before applying the Fourier transform. While this smoothes the cropped field, it ensures alias-free spectral analysis. We radially aggregate the magnitude of the Fourier coefficients. For example, bin $3$ contains the sum of all magnitudes of Fourier coefficients such that $\|\mathbf{k}\|_2 \in [2.5, 3.5)$. The results of this, separated by PDE, resolution, and initializer, are presented in \cref{fig:spectral_distibution}. We also show an aggregated version across all PDEs and simulation resolutions in \cref{fig:spectral_analysis_aggregated}.

We see that the linear equations (Diffusion \& Hyper-Diffusion) retain the characteristic spectra of their respective initializers, but also further diffuse them. The Burgers equation develops noticeable shocks, as evidenced by a richer spectral content than in the initialization. Also, the pattern-forming KPP-Fisher and the Swift-Hohenberg equation display rich spectral content characteristic of their polynomial nonlinearities. As expected, the Kuramoto-Sivashinsky in its chaotic state attains a characteristic spectrum that depends on the domain extent $L$.

Interestingly, we see that using different simulation resolutions with consistently sized crops exposes the model to the same phenomena at different scales. This is most noticeable for the KS equation, which develops the same spectrum, independent of the resolution. However, because the crops in the high-resolution simulation occupy a small physical space, the model sees a smoother version of that same pattern. We hypothesize that this relationship of similar patterns across different resolutions enables the model to fundamentally understand and be applied at varying resolutions.

\FloatBarrier

\subsection{Online learning framework\label{sec:online_training_detail}}
\subsubsection{Sampling strategies and data pipelines}\label{sec:sampling-strategies}

Since the pre-training objective relies on reconstructing physical fields across a vast manifold of plausible physics states, we devised a procedural sampling strategy that maximizes the diversity of states $\mathbf{u}_t$ while maintaining a balanced distribution of physical phenomena. We employ a decoupled client-server architecture:
a dedicated simulation server continuously synthesizes physical trajectories and pushes individual frames to an asynchronous First-In-First-Out (FIFO) queue. The training clients consume data from this queue, ensuring that the computationally expensive simulation steps do not bottleneck the GPU training throughput. For a detailed breakdown of the communication protocol, see \cref{sec:buffer-and-communication}. The procedural generation logic is formalized in \cref{alg:data_gen} and detailed below.

\textbf{Continuous Generation Cycle}
The simulation server operates in an infinite loop, cycling through the set of available PDE systems (e.g., Burgers, Swift-Hohenberg) and the set of resolutions $\mathcal{N} = \{64, 128, 256, 384\}$ in random order. By systematically varying the simulation resolution $N$, we ensure that the training crops expose the model to spectral features at different scales, mimicking the multi-scale nature of downstream tasks.

\textbf{Throughput Standardization and Channel Batching}
To maintain consistent tensor shapes and maximize computational efficiency, we standardize the generation process around three-channel inputs. 
\begin{itemize}
    \item \textbf{Vector Fields:} Systems such as the 3D Burgers equation naturally possess three channels ($C=3$). These are integrated as a single dependent system.
    \item \textbf{Scalar Fields:} For single-component systems ($C=1$, e.g., Diffusion, Fisher-KPP), we instantiate three independent initial conditions in parallel. These are batched along the channel dimension to form a pseudo-3-channel tensor, $\mathbf{u}_t \in \mathbb{R}^{3 \times N \times N \times N}$.
\end{itemize}
Upon completion of a time-step, the channels are decoupled and pushed to the queue individually. \emph{Exception:} The Kuramoto-Sivashinsky (KS) equation is integrated as a single channel without parallel batching. Consequently, the KS equation contributes proportionally less data volume (approx. 1/3) compared to other phenomena.

\textbf{Transient Dynamics (Physics Warmup)}
For some of the PDE systems, we only produce frames within the physically most meaningful regime. In the case of the Burgers equation, this would be in the shock formation and propagation phase. For the KS equation, this is within the chaotic attractor. To ensure we produce samples in these parts of the physics state space, we implement a \emph{physics warmup} phase. For every trajectory, the simulator integrates for $W$ steps (see \cref{tab:pde-trajectory-configs}), which are strictly discarded. Data recording commences only after this period.

\textbf{Queue Pre-filling (Server Warmup)}
Distinct from the physics warmup, we employ a \emph{server warmup} strategy to ensure the initial data distribution is sufficiently diverse. We define a server round counter $r$ and a target threshold $R$. During the startup phase ($r < R$), we artificially truncate the recorded trajectory length by an early-stop ratio $\xi = \min(r/R, 1)$. This increases the turnover rate of simulators, rapidly filling the buffer with diverse physics states.

\textbf{Trajectory Balancing and Re-initialization}
Regardless of the underlying physics, each simulator runs for a specific number of repetitions (\textbf{Num Runs}, see \cref{tab:pde-trajectory-configs}) such that it contributes exactly $30$ distinct time steps to the dataset before being discarded. Once this quota is met, the simulator is torn down and a new system is instantiated with fresh constitutive parameters and initial conditions.

\textbf{Spatial Subsampling and Normalization}
To optimize bandwidth, we extract random spatial crops $\mathbf{C}$ of size $H' \times H' \times H'$ (where $H'=96$). If the native simulation resolution is $N=64$, the full frame is transmitted. Finally, to stabilize the input distribution, each crop is clamped to the precomputed value ranges defined in \cref{tab:pde-discretization-configs}.

\textbf{Fault Tolerance and Checkpointing}
To support long-running training jobs on preemptible clusters, the simulation server is designed to be fully checkpointable. We serialize the active simulator configurations. This ensures that training can be paused and resumed deterministically without altering the data distribution or repeating sequences.

\begin{algorithm}[]
\caption{Procedural Data Generation Loop}
\label{alg:data_gen}
\begin{algorithmic}[1]
\REQUIRE Set of Equations $\mathcal{E}$, Set of Resolutions $\mathcal{N} = \{64, 128, 256, 384\}$
\REQUIRE Server Warmup Rounds $R=10$, Crop size $H'=96$, Server Queue $\mathcal{Q}$
\STATE Set simulation round counter $r=0$
\WHILE{Server Running}
    \STATE $r \leftarrow r+1$
    \FOR{$\{E, N\}$ in all combinations of $\mathcal{E} \times \mathcal{N}$}
        \STATE \textbf{1. Simulator Configuration}
        \STATE Sample constitutive parameters (e.g., $\nu, \zeta$) per \cref{tab:pde-overview}
        \STATE Retrieve discretization settings ($\Delta t$, save-freq, value-range) per \cref{tab:pde-discretization-configs}
        \STATE Retrieve trajectory settings (physics-warmup $W$, length $T$, num-runs) per \cref{tab:pde-trajectory-configs}
        \STATE Instantiate ETDRK time-stepper $\mathcal{P}$
        
        \STATE
        \FOR{$1 : \text{num-runs}$}
            \STATE \textit{// Initialize States}
            \FOR{$b = 1 : 3$}
                \STATE \COMMENT{For KS equation, only $b=1$ is executed}
                \STATE Sample IC type $\mathcal{I}$ and hyper-parameters per \cref{tab:pde-and-initializers}
                \STATE Generate $\mathbf{u}_0^{(b)} \sim \mathcal{I}$ and apply IC normalization
            \ENDFOR
            \STATE Stack initial states: $\mathbf{u}_0 \leftarrow \text{Concat}(\mathbf{u}_0^{(1)}, \dots, \mathbf{u}_0^{(3)})$
            \STATE
            
            \STATE \textit{// Time Integration}
            \STATE Reset Simulator with $\mathbf{u}_0$
            \STATE Compute server early-stop ratio $\xi = \min(r/R, 1)$
            \STATE $T_{total} \leftarrow (T \cdot \text{SaveFreq} \cdot \xi) + W$
            
            \FOR{$1 : T_{total}$}
                \STATE $\mathbf{u}_{t+\Delta t} \leftarrow \mathcal{P}(\mathbf{u}_t)$ \COMMENT{Step via ETDRK (\cref{eq:etdrk,eq:etdrk_2})}
                
                \IF{$t > W$ \textbf{and} $(t-W) \pmod{\text{SaveFreq}} == 0$}
                    \FOR{$c = 1:3$}
                        \STATE \textit{// Post-processing \& Transport}
                        \STATE $\mathbf{C} \leftarrow (N > H') \; ? \; \text{RandCrop}(\mathbf{u}_{t+\Delta t}[c], H') \; : \; \mathbf{u}_{t+\Delta t}[c]$
                        \STATE Clamp $\mathbf{C}$ to limits per \cref{tab:pde-discretization-configs}
                        \STATE Push $\mathbf{C}$ to $\mathcal{Q}$
                    \ENDFOR
                \ENDIF
            \ENDFOR
        \ENDFOR
    \ENDFOR
\ENDWHILE
\end{algorithmic}
\end{algorithm}

\subsubsection{Buffer and communication strategy}\label{sec:buffer-and-communication}

To bridge the gap between high-speed numerical solvers and deep learning frameworks, we implemented a custom asynchronous data loading pipeline. This system abstracts the continuous simulation stream into a PyTorch-compatible dataset, allowing the training loop (managed via PyTorch Lightning) to interface with the procedural generators as if they were a standard static dataset.

\textbf{Architecture and Communication}
The pipeline operates on a Producer-Consumer model implemented via \texttt{torch.multiprocessing}. A dedicated subset of GPU resources is assigned to the \emph{Producer} role (simulation), while the remaining GPUs function as \emph{Consumers} (training). Communication between these processes is handled via asynchronous thread-safe queues.
Similar to the active learning framework Melissa~\cite{Melissa2017}, our approach bypasses the file system entirely. Data is passed directly through shared memory or TCP sockets (depending on the node topology), eliminating disk I/O latency.
To further reduce the communication overhead, we employ a ``transport crop'' strategy: simulation frames are cropped to an intermediate size of $H_{X,Y,Z}' = 96^3$ before transmission, as outlined in \cref{sec:sampling-strategies}. This significantly reduces payload size compared to full-resolution grids while remaining larger than the final training crop, enabling further data augmentation/cropping on the consumer side.

\textbf{Multi-Stage Buffering and Latency Hiding}
Although spectral solvers are computationally efficient, network latency and synchronization overheads can cause pipeline stalls. To mitigate this, we employ a hierarchical buffering strategy:
\begin{enumerate}
    \item \textbf{Transmission Queue (FIFO):} The simulation server pushes completed transport samples into a finite-sized First-In-First-Out buffer. If this buffer fills up, the simulator pauses, preventing memory overflows. From this queue, data is sent to all participating training GPUs in a round-robin fashion.
    \item \textbf{Local Staging Buffer (FIFO):} Each training GPU maintains an incoming ``mailbox'' queue. New frames are received here before being processed for the training cache.
    \item \textbf{Consumer Cache (MFU):} On the training side, frames are moved from the staging buffer into a larger local cache governed by a Most-Frequently-Used (MFU) replacement policy. Background threads continuously replenish this cache.
\end{enumerate}
The training loop samples batches from the Consumer Cache rather than the stream directly. This decouples the training step time from the simulation step time. Consequently, even if the simulator throughput fluctuates (e.g., due to varying solver times based on resolution or other overhead), the trainer always has immediate access to data.

\textbf{Epoch Definition in Infinite Streams}
In this procedural paradigm, the concept of an ``epoch'', traditionally seen as one full pass over a static dataset, becomes ill-defined. We redefine an epoch as a fixed number of samples seen during training, which we set to $13'200$. 

\textbf{Numerical Stability Guardrails}
Given the stochastic initialization of parameters, numerical instabilities (divergences) are rare but possible. To prevent invalid gradients from propagating into the model weights, we implement a strict fail-safe mechanism.
Before any frame enters the transmission queue, it is scanned for \texttt{NaN} or \texttt{Inf} values. If a numerical anomaly is detected:
\begin{enumerate}
    \item The corrupted trajectory is immediately discarded.
    \item The specific simulator instance responsible is reset with a new random seed and parameters.
    \item A global error counter is incremented.
\end{enumerate}
If the error counter exceeds a tolerance threshold (set to $10$ events per training run), the entire training process is halted to allow for debugging. This ensures that the model is never exposed to corrupted gradients.

\textbf{Multi-Node and Distributed Training}
Our default configuration utilizes a single node with four GPUs (1 Producer, 3 Consumers). Under PyTorch Lightning's data-parallel strategy, microbatches are distributed among the consumers, and gradients are synchronized via AllReduce.
We also experimented with multi-node configurations by replicating the topology: each node instantiates its own local Producer GPU alongside its local Consumers. This \emph{Local-Producer Strategy} offers a significant bandwidth advantage. By confining the transmission of high-dimensional simulation tensors to the intra-node PCIe bus, we ensure that inter-node interconnects (e.g., InfiniBand) are reserved exclusively for gradient synchronization. This demonstrates the practical scalability of procedural online training, effectively bypassing both disk I/O and inter-node bandwidth bottlenecks.

\subsection{Downstream datasets \label{sec:downstream_dataset}}

For the downstream tasks, we include 4 challenging datasets. Details of these datasets are summarized as follows:

\iso{} contains a direct numerical simulation of homogeneous isotropic turbulence from Johns Hopkins Turbulence Database (JHTDB, Open Data Commons Attribution License (ODC-By)) \cite{jhtdb2008}, in which the statistical properties are invariant under translations and rotations of the coordinate system. We sample 500 frames from the original dataset. In the autoencoding task, random $64^3$ crops are generated from the first 420 frames for training. And we select 3 complete $ 1024^3$-resolution frames from the remaining 80 for testing. In the dynamics learning task, random $128^3$ crops are still generated from the first 420 frames for training, and the testing crops are generated from the remaining 80 frames. Below is a brief summary of the key characteristics of the dataset:

\begin{itemize}
    \item Spatial resolution: $X=1024,Y=1024,Z=1024$
    \item Spatial size: $[0,2\pi]\times[0,2\pi]\times[0,2\pi]$
    \item Reynolds number: $Re =433$
    \item State variables: x/y/z components of velocity and pressure.
    \item Time step between stored data: 0.002
    \item Boundary conditions: periodic
\end{itemize}

\tcf{} contains 21 simulations with Reynolds numbers ranging from $Re=400$ to $Re=800$ simulated with PICT \cite{pict2026}. Each simulation contains 200 snapshots. In the autoencoding task, random $48^3$ crops are generated from the first 20 simulations for training. We select 200 full-resolution frames from the remaining 1 simulation for testing. In the generative modeling task, the latent flow matching models are trained in the latent space of the first 20 simulations.  Below is a brief summary of the key characteristics of the dataset:

\begin{itemize}
    \item Spatial resolution: $X=96,Y=96,Z=192$
    \item Spatial size: $[-1,1]\times[-1,1]\times[-\pi,\pi]$
    \item Reynolds number: $Re \in [400,800]$
    \item State variables: x/y/z components of velocity.
    \item Time step between stored data: 0.1
    \item Boundary conditions: periodic (x), wall (y,z)
\end{itemize}

\mhd{} contains 100 frames sampled from the magnetohydrodynamics turbulence simulation of the JHTDB \cite{jhtdb2008}. We generate crops of size $512^3$ from the original $1024^3$ simulations. In the autoencoding task, random $64^3$ crops are generated from the first 80 frames for training. And we select complete $ 512^3$-resolution frames from the remaining 20 for testing. Below is a brief summary of the key characteristics of the dataset:
\begin{itemize}
    \item Spatial resolution: $X=512,Y=512,Z=512$ (cropped from $1024\times 1024\times 1024$)
    \item Spatial size: $[0,\pi]\times[0,\pi]\times[0,\pi]$ (cropped from $[0,2\pi]\times[0,2\pi]\times[0,2\pi]$)
    \item Reynolds number: $Re = 186$
    \item State variables: x/y/z components of velocity, pressure, x/y/z components of magnetic field, x/y/z components of vector potential
    \item Time step between stored data: 0.025
    \item Boundary conditions: crop
\end{itemize}

\tbl{} contains 940 frames sampled from the transitional boundary layer simulations of the JHTDB \cite{jhtdb2008}. We generate crops of size $224^3$ from the original $10240\times 1536\times 2048$ simulations. In the autoencoding task, random $32^3$ crops are generated from the first 840 frames for training. And we select complete $224^3$-resolution frames from the remaining 100 for testing. In the dynamics learning task, random $128^3$ crops are still generated from the first 840 frames for training, and the testing crops are generated from the remaining 100 frames. The pressure channel is not included in the dynamics channel because the original data is not well normalized for it. Below is a brief summary of the key characteristics of the dataset:
\begin{itemize}
    \item Spatial resolution: $X=224,Y=224,Z=224$ (cropped from $10240\times 1536\times 2048$)
    \item Spatial size: $[293.2, 314.3]\times[0,3.9]\times[0,26.25]$ (cropped from $[0,969.8]\times[0,26.5]\times[0,240]$)
    \item Reynolds number: $Re = 800$
    \item State variables: x/y/z components of velocity, pressure
    \item Time step between stored data: 1.25
    \item Boundary conditions: crop
\end{itemize}

\FloatBarrier
\newpage
\section{Training details and network architectures\label{sec:training_details}}
\subsection{Network architectures\label{sec:training_details:network_details}}

\begin{landscape}
\begin{figure}[htbp]
\centering
    \centering
    \includegraphics[width=\linewidth]{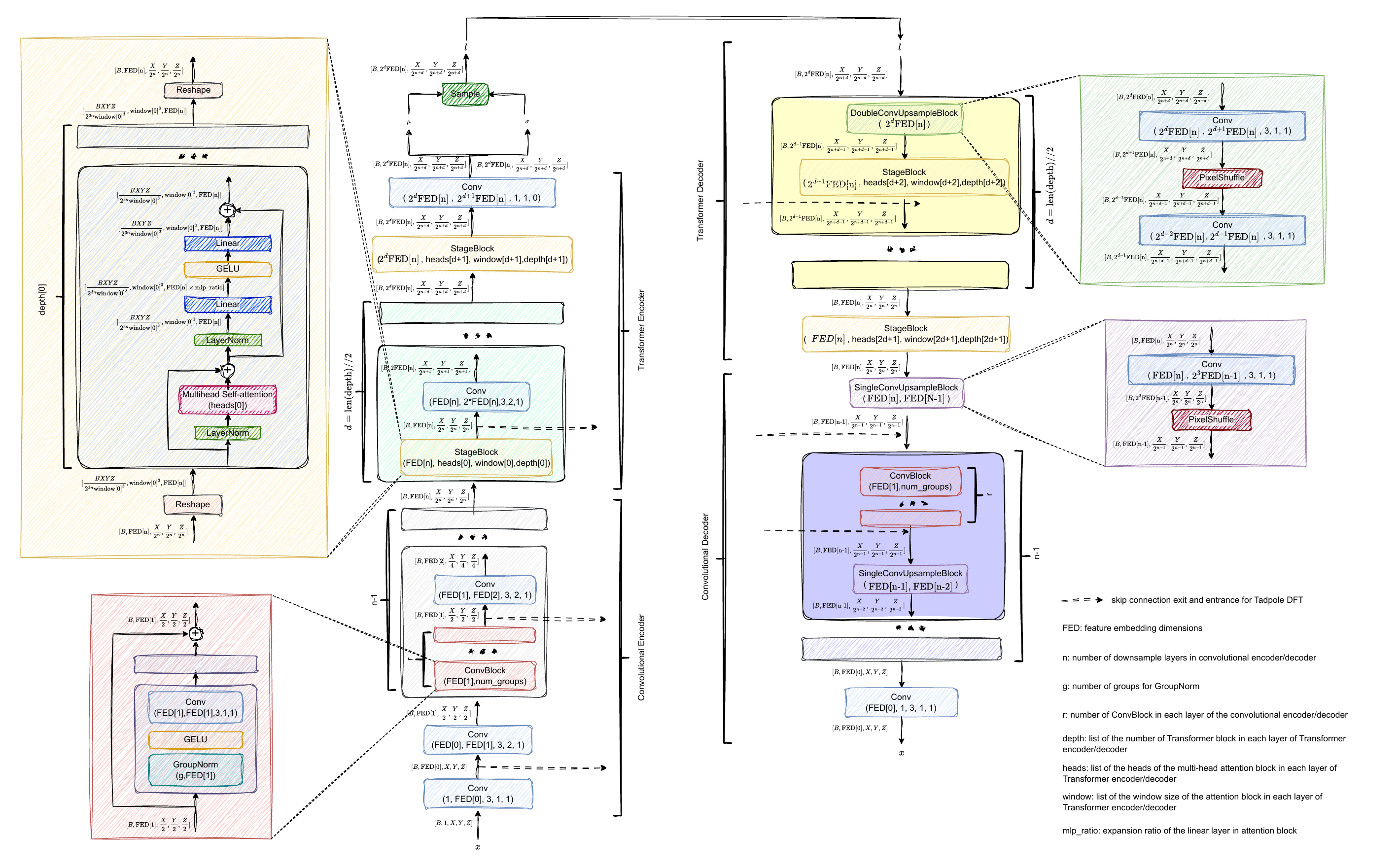}
    \caption{Network architecture of Tadpole based on P3D~\cite{p3d2025}.}
    \label{fig:network}
\end{figure}
\end{landscape}

We build the backbone of Tadpole based on P3D~\cite{p3d2025}. \cref{fig:network} illustrates the architecture of the network, and \cref{tab:network_hyperparameters} summarizes the hyperparameters used for Tadpole with different sizes. Compared to the original P3D architecture, the embedding layers for PDE parameters and skip connections are removed, and an additional convolutional layer is appended to the encoder to project its output to the mean and log-variance of the latent distribution. The discriminator network $\mathcal{A}$ adopts the same architecture as the encoder, but removes the final convolutional projection layer. This makes $\mathcal{A}$ a patch-based discriminator, and we utilize the mean of its output as the final belief.

\begin{figure}[htbp]
\centering
\includegraphics[width=0.5\linewidth]{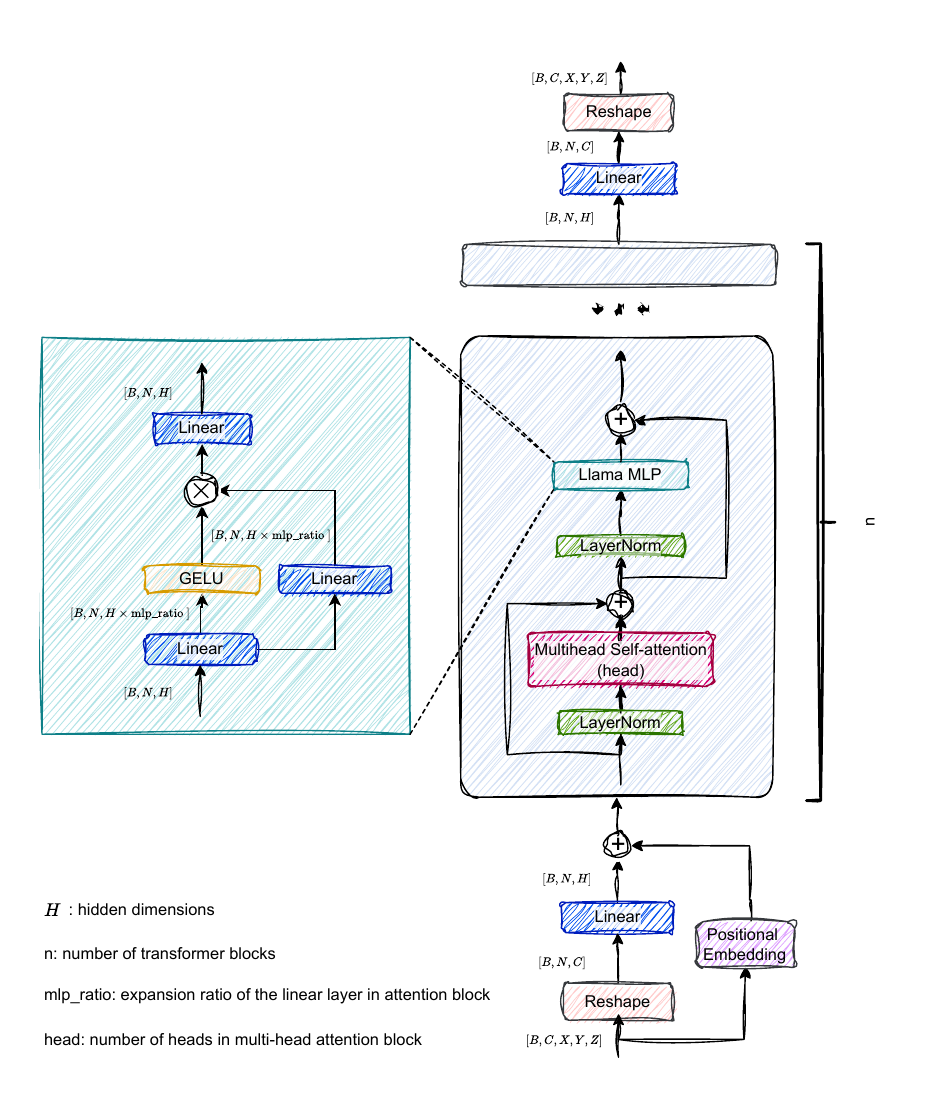}
\caption{Network architecture of the sub-network \subnet{}.}
\label{fig:sub_network}
\end{figure}

\floatingformatedtable{Architecture hyperparameters of Tadpole with different sizes. The definition of each hyperparameter can be found in \cref{fig:network}}{tab:network_hyperparameters}{

\begin{tabular}{cccc}
\toprule
          & S               & B               & L               \\ \midrule
FED       & [32,  32, 64]   & [64, 128, 128]  & [128, 256, 256] \\
n         & 2               & 2               & 2               \\
r         & 2               & 2               & 2               \\
g         & 16              & 32              & 32              \\
depth     & [2, 2, 2, 2, 2] & [2, 2, 2, 2, 2] & [2, 2, 2, 2, 2] \\
heads     & [4, 4, 4, 4, 4] & [4, 4, 4, 4, 4] & [8, 8, 8, 8, 8] \\
window    & [4, 4, 4, 4, 4] & [4, 4, 4, 4, 4] & [4, 4, 4, 4, 4] \\
mlp ratio & \multicolumn{3}{c}{4}                               \\ \bottomrule
\end{tabular}
}
For the sub-network \subnet{} used in the current experiments, we use a standard encoder-only transformer architecture. \cref{fig:sub_network} illustrates the architecture of the network, and \cref{tab:sub_network_hyperparameters} summarizes the hyperparameters with different sizes.

\floatingformatedtable{Architecture hyperparameters of \subnet{} with different sizes. The definition of each hyperparameter can be found in \cref{fig:sub_network}}{tab:sub_network_hyperparameters}{
\begin{tabular}{cccc}
\toprule
          & S     & B     & L     \\ \midrule
$H$       & 144   & 176   & 224   \\
n         & 4     & 6     & 8     \\
head      & \multicolumn{3}{c}{8} \\
mlp ratio & \multicolumn{3}{c}{4} \\ \bottomrule
\end{tabular}
}

\FloatBarrier
\subsection{Training objective\label{sec:training_objective}}
The loss function for the VAE consists of three terms: a reconstruction loss, a KL-divergence regularization term, and an adversarial loss term weighted by $\lambda_\mathcal{A}$. The discriminator loss function encourages correct classification of real and reconstructed samples. For the adversarial loss, the discriminator \discriminator{} outputs a scalar score $\mathcal{A}(\mathbf{u}_t)$ indicating the authenticity of the current state $\mathbf{u}_t$. The discriminator is trained using a hinge loss \cite{geometricgan2017,hierarchical2017,cgans2018}, and the overall objectives for the VAE and the discriminator are defined as follows:

\begin{align}
\begin{split}
\mathcal{L}_{VAE}
 = &
\mathbb{E}_{p_{\mathcal{E}}(\mathbf{z}_t|\mathbf{u}_t)}
\left[-\log p_{\mathcal{D}}(\mathbf{u}_t|\mathbf{z}_t)\right] 
\\
& +
\lambda_\text{KL}
\text{KL}(p_{\mathcal{E}}(\mathbf{z}_t|\mathbf{u}_t)||q(\mathbf{z}_t))
\\
&
-\lambda_\mathcal{A}\mathbb{E}_{p_{\mathcal{E}}(\mathbf{z}_t|\mathbf{u}_t)
}
[\mathcal{A}(\mathcal{D}(\mathbf{z}_t))]
\end{split}
\label{eq:loss_function_vae}
\end{align}
\begin{equation}
\begin{matrix}
\mathcal{L}_{Dis}
=
\mathbb{E}_{
p(\mathbf{u}_{t})
}
[
\max(0,1-
\mathcal{A}(\mathbf{u}_t))
]+
\mathbb{E}_{
p_{\mathcal{E}}(\mathbf{z}_{t}|\mathbf{u}_t)
}
[
\max(0,1+
\mathcal{A}(\mathcal{D}(\mathbf{z}_t)))
]
\end{matrix}
\label{eq:loss_function_discriminator}
\end{equation}

For the KL-divergence term in \cref{eq:loss_function_vae}, we set $\lambda_\text{KL}=10^{-6}$. For the adversarial loss term, we use a gradient-based scale strategy \cite{taming2020} for $\lambda_\text{adv}$ with a maximum scale value of $10^{-4}$. The discriminator will only be trained when the L2 reconstruction loss is below a threshold of 0.001 to stabilize training. After the start of training, the feedback from the discriminator will not be added to Tadpole's training directly until a 1000-iteration learning rate warm-up stage, followed by another 1000 iterations of warm-up for $\lambda_\text{adv}$. %

\subsection{Training hyperparameters\label{sec:training_details:training_hyperparameters}}

\floatingformatedtable{Training hyperparameters of Tadpole}{tab:training_hyperparameters}{
\begin{tabular}{ccccc}
\toprule
\multicolumn{2}{c}{Training}                          & Batch Size          & Learning Rate                                                                                             & Iterations \\ \toprule
\multirow{3}{*}{Pre-training}   & S                   & \multirow{3}{*}{48} & \multirow{2}{*}{$5\times10^{-5}$}                                                                         & 550000     \\
                                & B                   &                     &                                                                                                           & 825000     \\
                                & L                   &                     & \begin{tabular}[c]{@{}c@{}}$5\times10^{-6}$ for \encoder{}\\ $5\times10^{-5}$ for \decoder{} and \discriminator{}\end{tabular} & 1700000     \\ \midrule
\multirow{3}{*}{Downstream task} & Autoencoding        & 32                  & $5\times10^{-5}$                                                                                          & 14000      \\
                                & Dynamics Learning   & 32                  & $2\times10^{-4}$                                                                                          & 56000      \\
                                & Generative Modeling & 256                 & $1\times10^{-4}$                                                                                                 & 26400      \\ \bottomrule
\end{tabular}
}
This section summarizes the training hyperparameters for Tadpole. The primary values are presented in \cref{tab:training_hyperparameters}.

\paragraph{Pre-training:}
Pre-training is conducted in bf16-mixed precision using the AdamW optimizer \cite{adamw2019} with $\beta_1=0.9$, $\beta_2=0.999$, and a weight decay of $10^{-15}$.
A loss-adaptive learning rate scheduler reduces the learning rate by a factor of 0.5 when the training loss decreases by an order of magnitude below the previous threshold. The initial and minimal learning rates are set to $5\times 10^{-5}$ and $5\times 10^{-6}$, respectively. A linear learning rate warm-up is applied during the first 1000 iterations for both Tadpole and the discriminator. The KL-divergence term in \cref{eq:loss_function_vae} is optimized solely by the encoder and becomes increasingly unstable as network size increases. Therefore, the initial learning rate of the Tadpole-L encoder is reduced to $5\times10^{-6}$. Different-sized Tadpoles are pre-trained with varying numbers of training iterations, as larger models require more iterations to converge. The training iterations for S, B, and L-size models are $5.5\times 10 ^5$, $8.25\times 10 ^5$, and $1.7\times 10 ^6$, respectively. The batch size for pre-training is 48, with gradient accumulation employed to reduce VRAM consumption.

\paragraph{Downstream autoencoding:} The downstream autoencoding uses the same hyperparameters as pre-training, except the batch size is reduced to 32 and the number of training iterations is set to $1.4\times 10^4$.

\paragraph{Downstream dynamics:}
The same hyperparameter configuration is applied to both \iso{} and \tcf{} datasets. Training is performed in bf16-mixed precision using the AdamW optimizer with $\beta_1=0.9$, $\beta_2=0.999$, and a weight decay of $10^{-15}$.
The learning rate is fixed at $2\times10^{-4}$. The number of training iterations is $5.6\times10^{3}$. The batch size is 32, with gradient accumulation used to reduce VRAM consumption.

\paragraph{Downstream generative modeling:}
Generative modeling training is conducted in bf16-mixed precision using the AdamW optimizer with $\beta_1=0.9$, $\beta_2=0.999$, and a weight decay of $10^{-15}$.
The learning rate is fixed at $1\times10^{-4}$. The number of training iterations is $2.64\times10^{3}$. The batch size is set to 256, as the latent generative model requires less training memory. For models trained in pixel space, gradient accumulation is used to reduce memory consumption.

\subsection{Training cost\label{sec:training_cost}}
Training foundation models incurs a substantial computational cost. 
Current Tadpole training involves multiple systems with different hardware configurations.
Below, we summarize the training costs for each model in terms of GPU hours. Note that if the model is trained with multiple GPUs in parallel, the total GPU hours are estimated by multiplying the actual training hours by the number of GPUs. Meanwhile, the training costs across different downstream datasets are similar as we typically use a fixed number of training iterations. Below, we show the average training cost estimates across datasets and runs. Central cost factor for pre-training is the model size: 
\begin{itemize}
\item Tadpole S Pre-training:  372 GPU hours with L40S GPUs.
\item Tadpole B Pre-training:   620 GPU hours with A100 GPUs.
\item Tadpole L Pre-training:  2300 GPU hours with A100 GPUs.
\end{itemize}

Training via fine-tuning is substantially faster, but likewise directly scales with model size:
\begin{itemize}
\item Tadpole B fine-tuning for autoencoding task (FPFT/Scr.):  40 GPU hours with L40S GPUs
\item Tadpole B fine-tuning for autoencoding task (LoRA 32):  36 GPU hours with L40S GPUs

\item Tadpole S for dynamics task (Tadpole-DFT):  64 GPU hours with L40S GPUs
\item Tadpole B for dynamics task (Tadpole-DFT):  120 GPU hours with L40S GPUs
\item Tadpole B for dynamics task (FPFT/Scratch):  110 GPU hours with L40S GPUs
\item Tadpole L for dynamics task (Tadpole-DFT):  280 GPU hours with A100 GPUs

\item Tadpole latent  generative models:  8 GPU hours with L40S GPUs
\end{itemize}
Nonetheless, even dynamic tasks using the L-size model converged in $8\times$ fewer GPU hours than pre-training.
Above, the NVIDIA L40S GPUs were equipped with 48GB RAM, while 
the A100 GPUs had 40GB RAM.

\FloatBarrier
\newpage

\section{Evaluation metrics \label{sec:evaluation_metrics}}
\subsection{Enstrophy-based evaluation for dynamics learning\label{sec:enstrophy} }
Neural networks are known to exhibit spectral bias, favoring the learning of low-frequency components while under-resolving high-frequency structures. This limitation becomes particularly pronounced in autoregressive rollout settings, where prediction errors accumulate over time and manifest as progressive attenuation of high-frequency modes, resulting in overly smooth or blurred solutions. Conventional pixel-space metrics, such as mean squared error, are dominated by large-scale features and may therefore underestimate the degradation of fine-scale structures. To more faithfully assess model performance, especially in long-horizon predictions, we incorporate spectrum-based evaluation metrics that quantify errors across frequency bands. The spectrum-based metrics provide a scale-resolved characterization of model accuracy and explicitly capture the loss of high-frequency content, which is critical in many PDE systems. That's why we emphasize the spectral metrics in the current manuscript.

The spectrum-based evaluation for the dynamic rollout test case of the main text is performed as follows: 
The enstrophy spectrum at wavenumber $k \in \mathbb{R}_+$ is given by
\begin{equation}
    S(k) = \sum_{k < |m| \leq k+1}\frac{1}{2} \sum (|\widehat{\omega_x}(m)|^2 + |\widehat{\omega_y}(m)|^2 + |\widehat{\omega_z}(m)|^2),
\end{equation}
where $\widehat{|\boldsymbol{\omega}_{x,y,z}}(m)|$, with $m \in \mathbb{Z}^3$, denotes the Fourier coefficients of the vorticity component. To quantify discrepancies between spectra, we compute the NRMSE between the averaged reference spectrum and the averaged spectrum of generated vorticity fields,
\begin{equation}
        \text{NRMSE}^{ES} = \sqrt{\frac{\operatorname{mean}_k\bigl(
        (S_\text{pred}(k) - S_\text{ref}(k))^2\bigr)}{\operatorname{mean}_k\bigl(S_\text{ref}(k)^2\bigr)}}
\end{equation}
Since the cropped regions are not periodic, discontinuities at the domain boundaries introduce artifacts in the Fourier transform. To mitigate these effects, we apply a Hann window to smoothly attenuate $\boldsymbol{\omega}$ toward the boundaries prior to computing the Fourier coefficients.

\FloatBarrier
\FloatBarrier
\subsection{Statistical evaluation for generative modeling} \label{section:stat_tcf}

Properly assessing the quality of generative models for scientific data is an open problem. Two central difficulties are that (1) the number of samples in the reference dataset is often small and (2) the dimensionality of the data is very high. The \tcf{} dataset comprises three velocity channels at a spatial discretization of $96 \times 96 \times 192$ in 3D. When flattened, this corresponds to a ca. $5M$-dimensional vector. Taking samples from the reference simulation is futher complicated, since snapshots that are close in time are highly correlated, which can have implications for the statistical evaluation, which often assumes that samples are independent. To avoid a high auto-correlation of samples from the reference simulations, we take every 10th sample from the \tcf{} dataset for Reynolds numbers in $[400, 500, 600, 700, 800]$, which corresponds to a step size $\Delta t = 1$. This yields $100$ reference samples in total. We generate the same number of samples for each generative model.

To simplify the evaluation process, we split a single high-resolution sample into multiple low-dimensional samples. While this means that information on long-range correlation and structure is lost, we consider the distributional metrics on the set of low-dimensional derived samples as a lower bound on the distributional metrics for the high-resolution data.

There are many strategies to reduce the high-resolution samples. We choose a crop-based strategy, which partitions the full $3 \times 96 \times 96 \times 192$-sized data into chunks of size $3 \times 16 \times 16 \times 16$. This transforms a single high-resolution sample into $432$ smaller samples, which have dimensionality $12\,288$. The smaller samples are no longer independent, however, we believe that this has a negligible effect on the evaluation. In total, there are $43\,200$ samples with reduced dimensionality.

Besides the NRMSE of the mean and std., we were not able to run the computation on the full set of samples or the full dimensionality due to computation and stability constraints. We denote the maximum number of samples used for computation with $n_\mathrm{points}$ and the maximum dimensionality with $d_\mathrm{max}$, and select the maximum values acceptable for the computation budget. If $n_\mathrm{points}$ is smaller than the dataset size, we randomly sample a subset whose size matches $n_\mathrm{points}$ without replacement. If $d_\mathrm{max}$ is smaller than the dimensionality of the data, we only use the first $d_\mathrm{max}$ dimensions. \cref{tab:n_d_distribution_metrics} shows the exact values of $n_\mathrm{points}$ and $d_\mathrm{max}$ used for different distributional metrics. 

\formatedtable{$n_\mathrm{points}$ and $d_\mathrm{max}$ for different distributional metrics.}{tab:n_d_distribution_metrics}{
\begin{tabular}{cccc}
\toprule
                    & $\chi^2_\mathrm{PQM}$ & $\mathcal{W}_1$ & $\text{MMD}_{\text{RBF}}$ \\ \midrule
$n_\mathrm{points}$ & 1000                  & 50000           & 10000                     \\
$d_\mathrm{max}$    & inf                   & 10000           & 100                       \\ \bottomrule
\end{tabular}
}

\FloatBarrier

\section{Nomenclature and abbreviations \label{sec:nomenclature_abbreviations}}
\subsection{Nomenclature}

\begin{itemize}
    \item $\mathbb{P}^i$: The $i$-th PDE in the PDE family.
    \item \current: The PDE solution at time step $t$.
    \item \latentcurrent: The latent representation of \current.
    \item \encoder: The encoder network that maps the current state \current{} to a latent representation \latentcurrent.
    \item \decoder: The decoder network that reconstructs the state from the latent representation \latentcurrent.
    \item \discriminator: The adversarial network (discriminator) that distinguishes between real and reconstructed states.
    \item \subnet: The sub-network used for downstream tasks or specific applications.
    \item $\lambda_\text{KL}$: The weight for the KL-divergence term in the loss function.
    \item $B$: The batch dimension.
    \item $C$: The number of channels in the input data.
    \item $X$, $Y$, $Z$: The spatial dimensions (height, width, depth) of the 3D input data.
    \item $H_i$: The final crop size along spatial dimension $i$.
    \item $H_i'$: The pre-crop size along spatial dimension $i$.
    \item $W_0$: The pre-trained weights of the Tadpole model.
    \item $r$: The LoRA rank used in fine-tuning.
    \item $A,B$: The LoRA adaptation matrices.
    \item $D$: The dimension of the 3D PDE data, denoted as $D=C\times X \times Y \times Z$.
    \item $\gamma$: The scale factor for the skip connections.
\end{itemize}
\subsection{Abbreviations}

\begin{itemize}
    \item PDE: Partial Differential Equation.
    \item PEFT: Parameter-efficient Fine-tuning.
    \item NLP: Natural Language Processing.
    \item CV: Computer Vision.
    \item FM: Foundation Model.
    \item FPFT: Full-parameter fine-tuning.
    \item \iso{}: Isotropic turbulence test case.
    \item \tcf{}: Turbulent Channel Flow test case.
    \item \mhd{}: Magnetohydrodynamics test case.
    \item \tbl{}: Transitional Boundary Layer test case.
    \item LLM: Large Language Models.
    \item ICL: In Context Learning.
    \item FIFO: First In First Out.
    \item MFU: Most Frequently Used
\end{itemize}

\section{Changelog\label{sec:changelog}}
\textbf{Version 1}: Initial submission to arXiv on 14 May 2026.

\textbf{Version 2 (current)}: Camera-ready version for NeurIPS 2026. Compared to the initial version, the major modifications are listed below:
\begin{itemize}
    \item \cref{fig:dynamics_ablation} and \cref{fig:dynamics_ablation_nrmse} add an additional ablation examining the effect of zero initialization of the DFT weights. The results demonstrate that zero initialization further enhances the performance of Tadpole-DFT.

    \item The dynamics prediction results of Tadpole-L have been updated (\cref{fig:dynamics_bar_plot} and \cref{tab:dynamics_bl_spectrum,tab:dynamics_bl_nrmse,tab:dynamics_iso_spectrum,tab:dynamics_iso_nrmse}) following a bug fix in the training script. In our scripts, we employed activation checkpointing to reduce GPU memory consumption when training the L-size model. However, our original implementation contained an issue in how the PyTorch checkpointing function was invoked.
    
    Specifically, for each encoder and decoder layer, we used \texttt{x, res = checkpoint(encoder\_level, x)} and \texttt{x = checkpoint(decoder\_level, x)}, respectively. With the default setting \texttt{use\_reentrant=True}, PyTorch employs a reentrant checkpointing implementation that runs the forward computation under \texttt{no\_grad} and requires at least one input and output to participate in gradient computation. Because the input tensor \texttt{x} at the checkpoint boundary did not require gradients (\texttt{requires\_grad=False}), gradients were not properly propagated through the checkpointed encoder. The residual tensors were returned within Python lists, which the reentrant implementation does not recognize as participating in autograd. Consequently, these residuals were detached from the computational graph (\texttt{grad\_fn=None}), preventing gradient propagation through the skip connections to the encoder parameters.
    
    We resolved this issue by explicitly setting \texttt{use\_reentrant=False} in both checkpoint calls: \texttt{x, res = checkpoint(encoder\_level, x, use\_reentrant=False)} and \texttt{x = checkpoint(decoder\_level, x, use\_reentrant=False)}. Unlike the reentrant implementation, non-reentrant checkpointing records the autograd graph and supports tensors nested within Python data structures. After this correction, we verified that the residual tensors had valid \texttt{grad\_fn} attributes and that all parameter groups exhibited matching gradient norms with and without activation checkpointing.

    The actual performance of Tadpole-L is better than the original value shown in the previous version of the manuscript. Now, Tadpole-L also beats the Walrus model with many fewer trainable parameters in the \iso{} case for one-step predictions, again highlighting the effectiveness of Tadpole pretraining.

    \item Following the NeurIPS reviewer's request, additional experiments comparing the performance of DPOT and Morph models trained from scratch and their pre-trained counterparts are included in \cref{sec:pre_training_gain_others}, which help clarify the contributions of the backbone and pretraining to Tadpole's superior performance.

    \item Additional experiments evaluating the computation scaling of Tadpole have been included in \cref{sec:computation_scaling}. These results further demonstrate that the single-channel cropping strategy reduces the model's memory cost.

\end{itemize}

\newpage
\section*{NeurIPS Paper Checklist}



\begin{enumerate}

\item {\bf Claims}
    \item[] Question: Do the main claims made in the abstract and introduction accurately reflect the paper's contributions and scope?
    \item[] Answer: \answerYes{} 
    \item[] Justification: The main claims made in the abstract and introduction reflect the paper's contributions and scope.

\item {\bf Limitations}
    \item[] Question: Does the paper discuss the limitations of the work performed by the authors?
    \item[] Answer: \answerYes{} 
    \item[] Justification: The limitations are discussed in \cref{sec:conclustion_outlook}.

\item {\bf Theory assumptions and proofs}
    \item[] Question: For each theoretical result, does the paper provide the full set of assumptions and a complete (and correct) proof?
    \item[] Answer: \answerNA{} 
    \item[] Justification: The paper does not include theoretical results. 

    \item {\bf Experimental result reproducibility}
    \item[] Question: Does the paper fully disclose all the information needed to reproduce the main experimental results of the paper to the extent that it affects the main claims and/or conclusions of the paper (regardless of whether the code and data are provided or not)?
    \item[] Answer: \answerYes{} 
    \item[] Justification: We have provided all the technical details needed to reproduce the results in \cref{sec:dataset_details,sec:training_details,sec:training_details}.

\item {\bf Open access to data and code}
    \item[] Question: Does the paper provide open access to the data and code, with sufficient instructions to faithfully reproduce the main experimental results, as described in supplemental material?
    \item[] Answer: \answerYes{} 
    \item[] Justification: The code is available at \url{https://github.com/tum-pbs/tadpole}. The dataset is too large to share directly, but it is easy to download the original data as described in the Appendix.

\item {\bf Experimental setting/details}
    \item[] Question: Does the paper specify all the training and test details (e.g., data splits, hyperparameters, how they were chosen, type of optimizer) necessary to understand the results?
    \item[] Answer: \answerYes{}{} 
    \item[] Justification: We have provided all the technical details needed to reproduce the results in \cref{sec:dataset_details,sec:training_details,sec:training_details}.

\item {\bf Experiment statistical significance}
    \item[] Question: Does the paper report error bars suitably and correctly defined or other appropriate information about the statistical significance of the experiments?
    \item[] Answer: \answerNo{}{} 
    \item[] Justification: Error bars are not reported due to the high  computational expense of 3D experiments and foundation model pre-training runs. However, all experiments use EMA weights during training to reduce the impact of randomness and exhibited stable convergence behavior.

\item {\bf Experiments compute resources}
    \item[] Question: For each experiment, does the paper provide sufficient information on the computer resources (type of compute workers, memory, time of execution) needed to reproduce the experiments?
    \item[] Answer: \answerYes{} 
    \item[] Justification: The training cost of the current study is discussed in \cref{sec:training_cost}.
    
\item {\bf Code of ethics}
    \item[] Question: Does the research conducted in the paper conform, in every respect, with the NeurIPS Code of Ethics \url{https://neurips.cc/public/EthicsGuidelines}?
    \item[] Answer: \answerYes{}{} 
    \item[] Justification: The research reported in the paper complies with the NeurIPS Code of Ethics.

\item {\bf Broader impacts}
    \item[] Question: Does the paper discuss both potential positive societal impacts and negative societal impacts of the work performed?
    \item[] Answer: \answerNA{} 
    \item[] Justification: This paper presents work to advance the field of machine learning for physical simulations. We don’t foresee any significant societal impacts directly associated with our work to be highlighted here.
    
\item {\bf Safeguards}
    \item[] Question: Does the paper describe safeguards that have been put in place for responsible release of data or models that have a high risk for misuse (e.g., pre-trained language models, image generators, or scraped datasets)?
    \item[] Answer: \answerNA{} 
    \item[] Justification: The current paper poses no such risks.

\item {\bf Licenses for existing assets}
    \item[] Question: Are the creators or original owners of assets (e.g., code, data, models), used in the paper, properly credited and are the license and terms of use explicitly mentioned and properly respected?
    \item[] Answer: \answerYes{} 
    \item[] Justification: The current paper uses datasets from other research work, which are correctly cited in \cref{sec:downstream_dataset}, and the original licenses have been explicitly mentioned and properly respected.

\item {\bf New assets}
    \item[] Question: Are new assets introduced in the paper well documented and is the documentation provided alongside the assets?
    \item[] Answer: \answerNA{} 
    \item[] Justification: The current submission doesn't release new assets yet. The current pre-trained weights of Tadpole foundation models are available on \url{https://huggingface.co/thuerey-group/tadpole}. 

\item {\bf Crowdsourcing and research with human subjects}
    \item[] Question: For crowdsourcing experiments and research with human subjects, does the paper include the full text of instructions given to participants and screenshots, if applicable, as well as details about compensation (if any)? 
    \item[] Answer: \answerNA{} 
    \item[] Justification: The paper does not involve crowdsourcing nor research with human subjects.

\item {\bf Institutional review board (IRB) approvals or equivalent for research with human subjects}
    \item[] Question: Does the paper describe potential risks incurred by study participants, whether such risks were disclosed to the subjects, and whether Institutional Review Board (IRB) approvals (or an equivalent approval/review based on the requirements of your country or institution) were obtained?
    \item[] Answer: \answerNA{} 
    \item[] Justification: The paper does not involve crowdsourcing nor research with human subjects.

\item {\bf Declaration of LLM usage}
    \item[] Question: Does the paper describe the usage of LLMs if it is an important, original, or non-standard component of the core methods in this research? Note that if the LLM is used only for writing, editing, or formatting purposes and does \emph{not} impact the core methodology, scientific rigor, or originality of the research, declaration is not required.
    \item[] Answer: \answerNA{} 
    \item[] Justification: The core method development in this research does not involve LLMs as any important, original, or non-standard components.

\end{enumerate}
\end{document}